\documentclass[pdflatex,sn-nature]{sn-jnl}%

\usepackage{graphicx}
\usepackage{tabularx}
\usepackage{amsmath,amssymb}
\usepackage{booktabs}
\usepackage{multirow}
\usepackage{siunitx}
\usepackage{xcolor}
\usepackage{bm}
\usepackage{amsthm}
\usepackage[capitalize,nameinlink,noabbrev]{cleveref}

\crefformat{table}{#2#1#3}
\Crefformat{table}{#2#1#3}

\crefmultiformat{table}{#2#1#3}{, #2#1#3}{, #2#1#3}{, and #2#1#3}
\Crefmultiformat{table}{#2#1#3}{, #2#1#3}{, #2#1#3}{, and #2#1#3}

\crefrangeformat{table}{#3#1#4--#5#2#6}
\Crefrangeformat{table}{#3#1#4--#5#2#6}

\theoremstyle{thmstyleone}

\theoremstyle{thmstyletwo}

\theoremstyle{thmstylethree}

\begin{document}

\newgeometry{
  left=25mm,
  right=25mm,
  top=26mm,
  bottom=26mm,
  bindingoffset=0mm
}

\title[A generalizable foundation model for intraoperative understanding across surgical procedures]{A generalizable foundation model for intraoperative understanding across surgical procedures}

\author[1,2,3]{\fnm{Kanggil} \sur{Park}}
\author[2,3]{\fnm{Yongjun} \sur{Jeon}}
\author[1]{\fnm{Soyoung} \sur{Lim}}
\author[1,2]{\fnm{Seonmin} \sur{Park}}
\author[1,2]{\fnm{Jongmin} \sur{Shin}}
\author[1]{\fnm{Jung Yong} \sur{Kim}}
\author[3]{\fnm{Sehyeon} \sur{An}}
\author[1]{\fnm{Jinsoo} \sur{Rhu}}
\author[1]{\fnm{Jongman} \sur{Kim}}
\author[1]{\fnm{Gyu-Seong} \sur{Choi}}
\author*[1,2]{\fnm{Namkee} \sur{Oh}}\email{namkee.oh@samsung.com}
\author*[2,3]{\fnm{Kyu-Hwan} \sur{Jung}}\email{kyuhwanjung@gmail.com}

\affil[1]{%
  \orgdiv{Department of Surgery}, 
  \orgname{Samsung Medical Center},
  \orgaddress{\city{Seoul}, \country{South Korea}}%
}

\affil[2]{%
  \orgdiv{Clinical Robotics and Embodied AI Research Center, Smart Healthcare Research Institute, Research Institute for Future Medicine}, 
  \orgname{Samsung Medical Center},
  \orgaddress{\city{Seoul}, \country{South Korea}}%
}

\affil[3]{%
  \orgdiv{Department of Medical Device Management and Research, 
    Samsung Advanced Institute for Health Sciences \& Technology (SAIHST)}, 
  \orgname{Sungkyunkwan University},
  \orgaddress{\city{Seoul}, \country{South Korea}}%
}

\abstract{
In minimally invasive surgery, clinical decisions depend on real-time visual interpretation, yet intraoperative perception varies substantially across surgeons and procedures. This variability limits consistent assessment, training, and the development of reliable artificial intelligence systems, as most surgical AI models are designed for narrowly defined tasks and do not generalize across procedures or institutions. Here we introduce ZEN, a generalizable foundation model for intraoperative surgical video understanding trained on more than 4 million frames from over 21 procedures using a self-supervised multi-teacher distillation framework. We curated a large and diverse dataset and systematically evaluated multiple representation learning strategies within a unified benchmark. Across 20 downstream tasks and full fine-tuning, frozen-backbone, few-shot and zero-shot settings, ZEN consistently outperforms existing surgical foundation models and demonstrates robust cross-procedure generalization. These results suggest a step toward unified representations for surgical scene understanding and support future applications in intraoperative assistance and surgical training assessment.
}

\keywords{Foundation Model, Minimally Invasive Surgery, Self-Supervised Learning, Surgical Video Analysis}

\maketitle

\begin{figure}[p]
  \centering
  \includegraphics[width=0.9\linewidth]{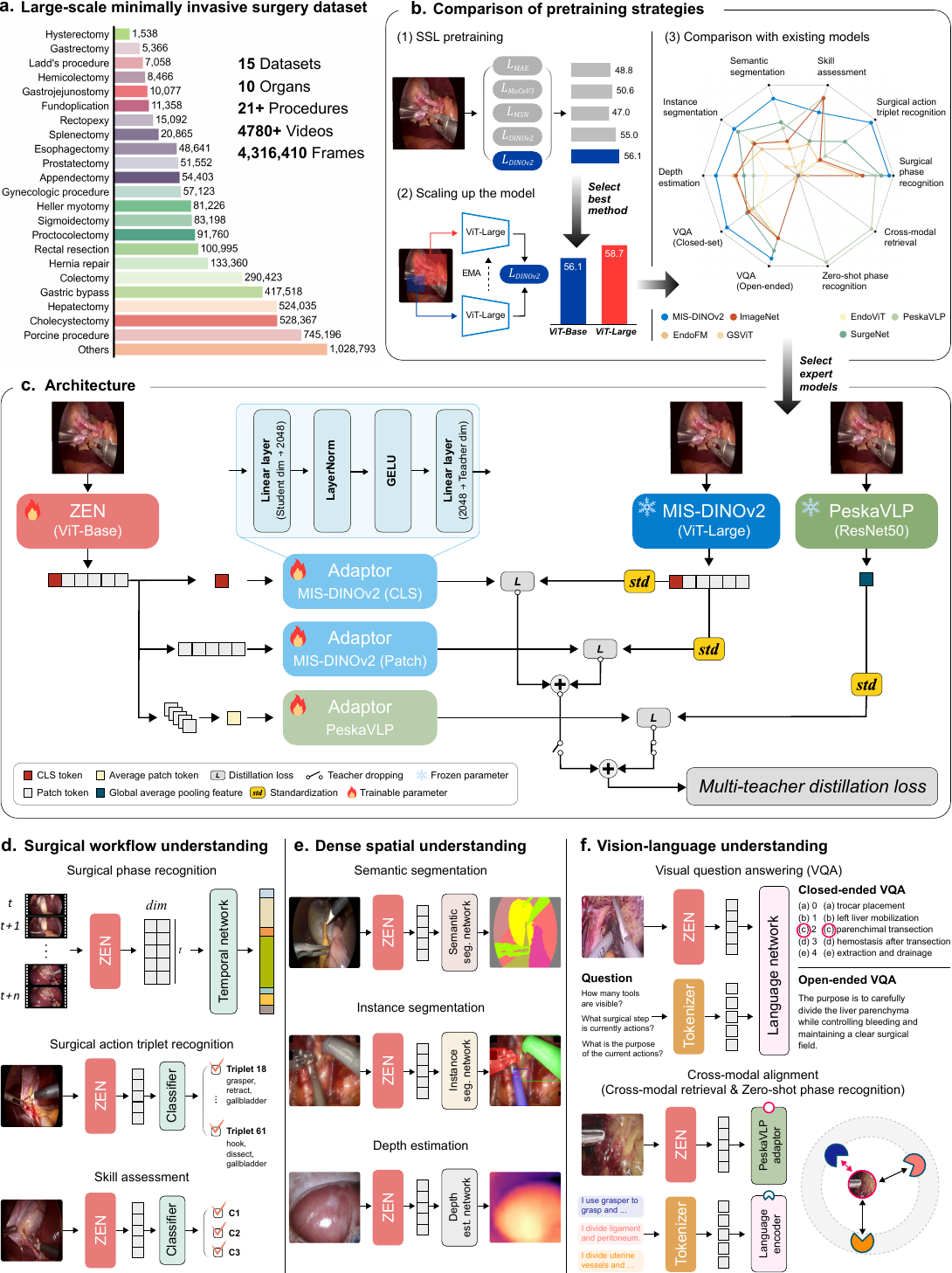}
  \caption{\textbf{Overview of this study.}
  \textbf{a,} Large-scale pretraining dataset comprising over 4.3 million frames from over 4,780 minimally invasive surgery videos, spanning over 21 procedures across 10 organs.
\textbf{b,} Comparison of pretraining strategies. Various self-supervised methods were evaluated across surgical downstream tasks, after which the best-performing strategy was scaled up and compared against existing pretrained models.
\textbf{c,} Architecture of ZEN and the multi-teacher distillation framework. ZEN is trained via feature-level distillation from multiple frozen expert teachers, including MIS-DINOv2 (ViT-Large) and PeskaVLP.
\textbf{d--f,} Downstream surgical evaluation tasks.
\textbf{d,} Surgical workflow understanding: surgical phase recognition, surgical action triplet recognition, and skill assessment.
\textbf{e,} Dense spatial understanding: semantic segmentation, instance segmentation, and monocular depth estimation.
\textbf{f,} Vision–language understanding: closed-ended and open-ended visual question answering, cross-modal retrieval, and zero-shot phase recognition.
}
  \label{fig:fig1}
\end{figure}

\section*{Introduction}\label{sec:introduction}
Minimally invasive surgery (MIS) is now the standard of care for many procedures, providing reduced postoperative pain, shorter hospital stays, and faster recovery compared to open surgery~\cite{MIS1,MIS2,MIS3}. Despite these advantages, MIS introduces greater procedural complexity, as surgeons must operate with limited visual and tactile feedback~\cite{spatialcog,visualcue}. Consequently, intraoperative decision-making in MIS remains highly dependent on individual surgeon experience and local practice patterns~\cite{decision}. Variability in anatomical recognition, workflow interpretation, and situational awareness contributes to preventable complications, technical errors, and inconsistent training outcomes~\cite{error}. Since MIS relies on real-time endoscopic video, computational video analysis is essential to improve intraoperative decision-making and patient safety~\cite{MISAI1,MISAI2}. Recent advances in artificial intelligence (AI) have shown strong potential for surgical video analysis in tasks such as surgical workflow analysis~\cite{phase1,phase2,phase3,phase4,phase5}. Despite this progress, clinically deployable systems remain limited. A major barrier is poor generalizability. Most existing models are developed for narrowly defined tasks within single procedures or institutions and fail to transfer across variations in anatomy, surgical technique, equipment, and operative context~\cite{general}. In addition, reliance on large, manually annotated datasets—particularly for video—poses substantial scalability challenges~\cite{lessismore}, limiting clinical translation.

Foundation models (FM) have emerged as a powerful paradigm to overcome these limitations. Pretrained on large datasets, FMs show broad generalization across diverse tasks and perform well even when labeled data are limited~\cite{generalfoundation1}. Building on these advances, medical AI is shifting towards domain-specific FMs~\cite{medicalfoundation1,medicalfoundation2,medicalfoundation3,medicalfoundation4,medicalfoundation5}. As open-access surgical video datasets continue to grow, this trend is expanding into the surgical field, and initial efforts to develop surgical FMs have begun to emerge~\cite{surgfm1,surgfm2,surgfm3,surgfm4,surgfm5,surgfm6}. However, several key questions remain unresolved. Previous work has relied on varying pretraining datasets and strategies, making it unclear which approaches are most effective for the surgical domain. Moreover, most existing models focus on workflow analysis, with only a few extending to tasks such as tool or anatomy segmentation (Fig.\ref{fig:fig2}a). Consequently, it remains uncertain whether current approaches can support the diverse and complex tasks required for real-world clinical use. Therefore, a comprehensive evaluation across a broader range of downstream tasks is essential for developing FMs that generalize reliably within the surgical domain.

To address this gap, we curated a large-scale pretraining dataset comprising 4,316,410 frames from over 4,780 videos spanning 10 organs and over 21 procedures (Fig.\ref{fig:fig1}a). Additionally, we established 20 clinical benchmarks covering surgical workflow understanding (phase recognition, action triplet recognition, and skill assessment) (Fig.\ref{fig:fig1}d), dense spatial understanding (semantic and instance segmentation, and depth estimation) (Fig.\ref{fig:fig1}e), and vision–language understanding, including VQA and cross-modal alignment tasks (cross-modal retrieval and zero-shot phase recognition) (Fig.\ref{fig:fig1}f). Using this unified resource, we conducted the first systematic comparison of various self-supervised learning (SSL) methods and evaluated them alongside existing surgical FMs (Fig.\ref{fig:fig1}b and Extended Fig.\ref{fig:Extended_Data_Fig1}).  To ensure a rigorous comparison, all SSL methods were initially trained on the same pretraining dataset using a Vision Transformer (ViT)~\cite{vit} Base architecture and were evaluated under identical downstream conditions. Our evaluation revealed that SimDINOv2~\cite{simdino} exhibited the strongest overall performance. Motivated by this result, we scaled SimDINOv2 to a ViT-Large encoder to enhance model capacity, producing ``MIS-DINOv2''. When benchmarked against existing surgical FMs, MIS-DINOv2 demonstrated superior performance across most downstream tasks. Notably, PeskaVLP~\cite{surgfm4} retained its advantage in cross-modal alignment tasks (cross-modal retrieval and zero-shot phase recognition), reflecting its training with vision–language contrastive learning~\cite{clip}, whereas other vision-only models lack this capability. These findings show that individual pretraining strategies offer distinct strengths across downstream tasks.

To unify these complementary strengths within a single framework, we introduce ZEN, a generalizable FM for MIS video analysis. Specifically, we introduce a self-supervised multi-teacher distillation strategy~\cite{amradio,unic,radiov2,dune} to synergize MIS-DINOv2 and PeskaVLP, enabling ZEN to inherit both robust spatial representations and precise vision–language alignment (Fig.\ref{fig:fig1}c). Across our comprehensive benchmark evaluation, ZEN not only demonstrates advanced cross-modal alignment capabilities but also consistently outperforms existing surgical FMs (Fig.\ref{fig:fig2}b). Moreover, ZEN demonstrates this superior performance across diverse evaluation settings, including limited labeled data and frozen-backbone settings, underscoring its robust generalization as a FM for the surgical domain.

\section*{Results}
\subsection*{Overall performance}
We evaluated ZEN and other surgical FMs using a clinical benchmark comprising 20 tasks across nine surgical procedures. These tasks cover three core domains: surgical workflow understanding (three phase recognition, two action triplet recognition, and one skill assessment task); dense spatial understanding (three semantic segmentation, one instance segmentation, and two depth estimation tasks); and vision–language understanding, comprising VQA (two closed- and one open-ended VQA) and vision–language alignment (text-to-image and image-to-text retrieval, and three zero-shot phase recognition tasks). The five vision–language alignment tasks were evaluated in a zero-shot setting, while the remaining 15 tasks used supervised evaluation. For supervised tasks, we evaluated the representation quality using a frozen-backbone setting and model adaptability using full fine-tuning. Overall performance was summarized using the average of each task’s representative metric, complemented by task-level rankings to account for heterogeneity in evaluation metrics across tasks.

\begin{figure}[!t]
  \centering
  \includegraphics[width=0.85\linewidth]{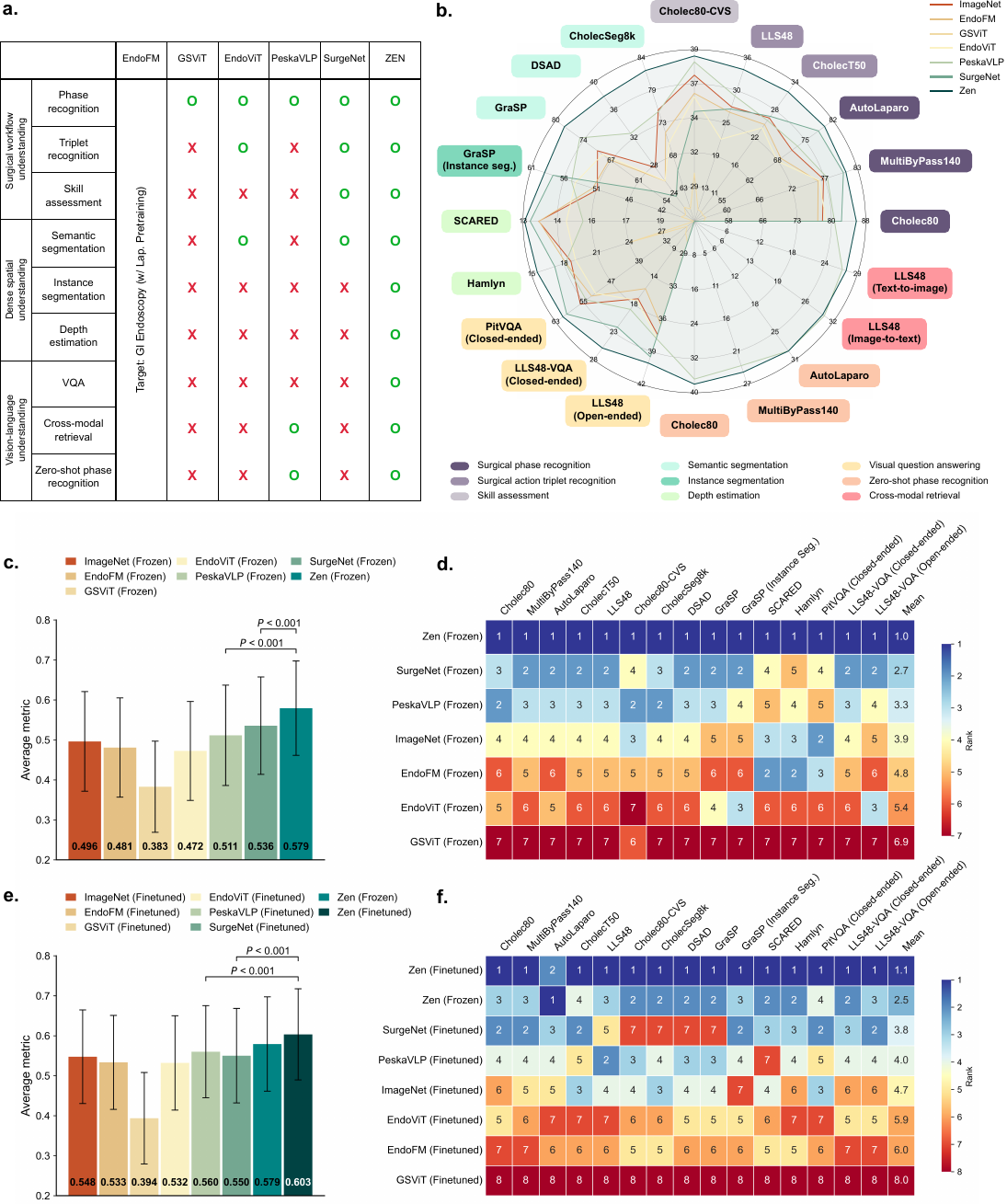}
  \caption{\textbf{Generalization performance of surgical foundation models on comprehensive clinical benchmarks.}
\textbf{a,} Comparison of downstream tasks targeted by existing surgical foundation models. Note that although EndoFM~\cite{surgfm2} utilizes MIS videos for pretraining, its target tasks primarily focused on gastrointestinal endoscopy.
\textbf{b,} ZEN outperforms other pretrained models across 20 clinical tasks in surgical video.
\textbf{c,} Average performance in the frozen-backbone setting across 15 supervised tasks, computed using task-specific representative metrics: video-level macro F1 score and accuracy for phase recognition; triplet (IVT) mean average precision (mAP) for action triplet recognition; mAP for skill assessment; Dice score for semantic segmentation; the average of detection and segmentation mAP for instance segmentation; 1 $-$ absolute relative error for depth estimation; the average of macro F1 score and balanced accuracy for closed-ended VQA; and the average of BLEU, ROUGE-L, and METEOR for open-ended VQA.
\textbf{d,} Ranking heatmap across the supervised tasks in the frozen-backbone setting, based on representative task-level metrics.
\textbf{e,} Average performance in the fine-tuned backbone setting across the supervised tasks, computed using representative metrics.
\textbf{f,} Ranking heatmap across the supervised tasks in the fine-tuned backbone setting.
Error bars in \textbf{c} and \textbf{e} indicate 95\% confidence intervals. $P$ values were calculated using a two-sided Wilcoxon signed-rank test.}
  \label{fig:fig2}
\end{figure}

In the frozen-backbone setting, ZEN achieved the highest average score (0.579) and the best mean rank of 1.0, outperforming the next-best model, SurgeNet (score 0.536; rank 2.7) (Fig.\ref{fig:fig2}c,d). In the full fine-tuning setting, ZEN again achieved the top performance with an average score of 0.603 and a mean rank of 1.1. Notably, the frozen ZEN model (score 0.579; rank 2.5) exceeded the performance of all other fully fine-tuned FMs, including SurgeNet (score 0.550; rank 3.8) (Fig.\ref{fig:fig2}e,f). When compared with the second- and third-best performing models, ZEN achieved significantly higher average performance ($P < 0.001$) in both the frozen and fine-tuning settings. Together, these results demonstrate ZEN’s strong generalization capabilities across diverse tasks in surgical domain.

\begin{figure}[!t]
  \centering
  \includegraphics[width=0.85\linewidth]{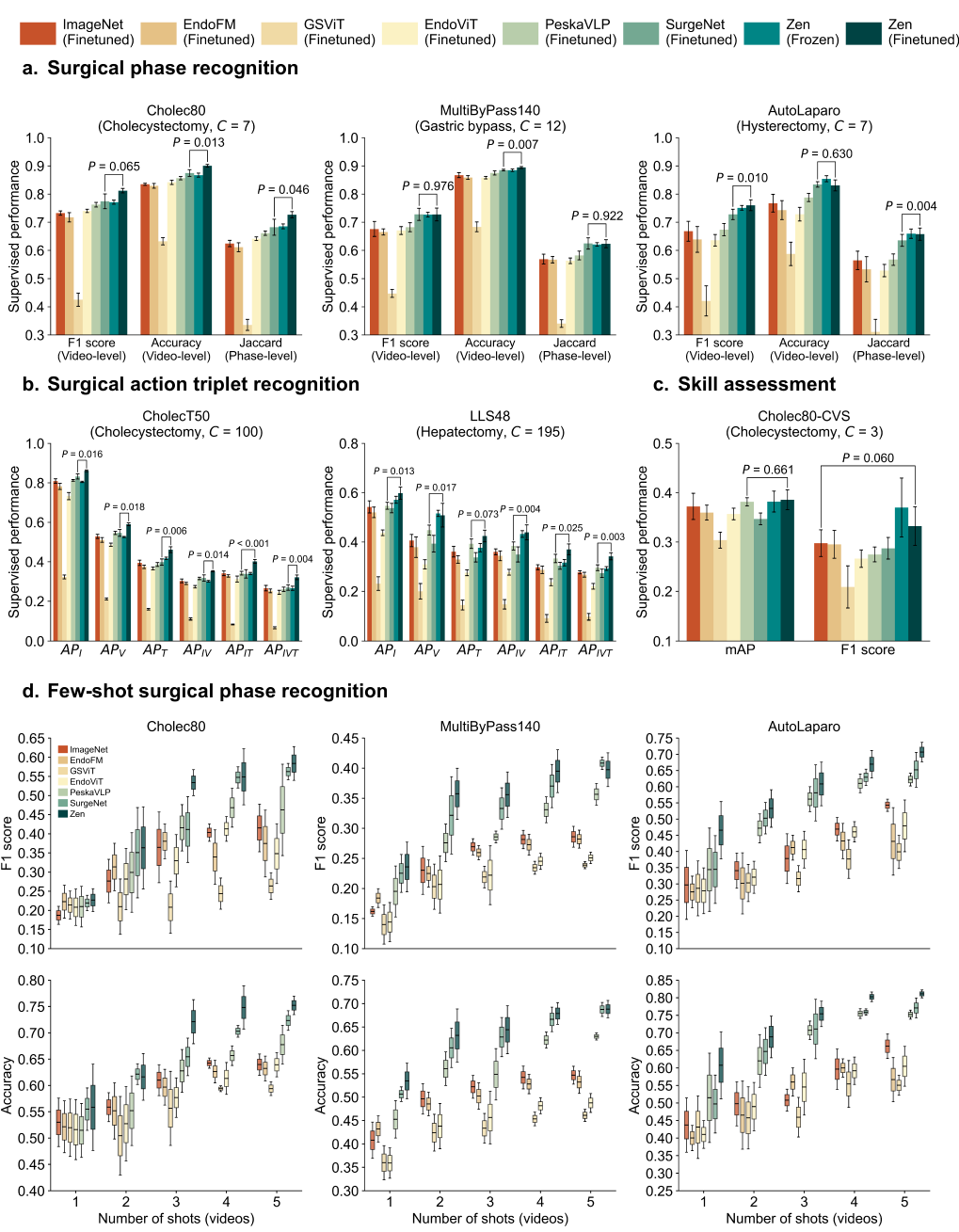}
  \caption{\textbf{Performance comparison for surgical workflow understanding.}
  \textbf{a,} Performance of ZEN and other pretrained models on surgical phase recognition across three datasets in the fine-tuned backbone setting. Metrics include video-level macro F1 score, accuracy, and phase-level Jaccard index. $C$ denotes the number of phases.
  \textbf{b,} Surgical action triplet recognition performance across two datasets in the fine-tuned backbone setting. Performance is evaluated using mean average precision (mAP) for instrument (I), verb (V), and target (T) components, as well as their combinations. $C$ denotes the number of triplet (IVT) classes.
  \textbf{c,} Skill assessment performance in the fine-tuned backbone setting, evaluated using mAP and macro F1 score. $C$ denotes the number of safety criteria.
  \textbf{d,} Few-shot surgical phase recognition performance using 1--5 training videos across three datasets. Results represent five independent runs ($n=5$) for each model and shot condition. The center of each box indicates the mean, box bounds denote the standard error, and whiskers indicate the lower and upper bounds of the 95\% confidence interval (CI). For \textbf{a--c}, error bars represent 95\% CIs over five independent runs ($n=5$). $P$ values were calculated using two-sided paired $t$-test.}
  \label{fig:fig3}
\end{figure}

\subsection*{Surgical workflow understanding}
Surgical workflows follow a hierarchical structure in which combinations of instruments, verbs, and targets define individual actions, and sequences of these actions form surgical phases. Skill assessment complements this hierarchy by evaluating whether key safety criteria are satisfied throughout the procedure. To assess ZEN’s understanding of surgical workflows at multiple levels of granularity, we benchmarked the model on three representative tasks: surgical phase recognition~\cite{phase1,phase2,phase3,phase4,phase5}, which captures the temporal progression of procedure phases; surgical action triplet recognition~\cite{rdv,selfd,terl,curconmix}, which models fine-grained instrument–verb–target (I–V–T) combinations; and skill assessment~\cite{deepcvs}, which evaluates whether predefined surgical safety criteria are satisfied during the procedure.

\paragraph{Surgical phase recognition}
Surgical phase recognition was evaluated on three public laparoscopic datasets: Cholec80~\cite{phase1} (cholecystectomy), MultiBypass140~\cite{multibypass} (Roux-en-Y gastric bypass surgery) and AutoLaparo~\cite{autolaparo} (hysterectomy). ZEN consistently outperformed other FMs on all three surgical phase recognition datasets under full fine-tuning, across multiple evaluation metrics including video-level F1 score, accuracy, and phase-level Jaccard index (Fig.\ref{fig:fig3}a and Supplementary Tables~\cref{tab:cholec80-finetune,tab:multibypass140-finetune,tab:autolaparo-finetune}). In the frozen-backbone setting, ZEN also achieved the strongest overall performance across the three datasets (Extended Fig.\ref{fig:Extended_Data_Fig2}a and Supplementary Tables~\cref{tab:cholec80-frozen,tab:multibypass140-freeze,tab:autolaparo-freeze}). Notably, the frozen ZEN model remained competitive with fully fine-tuned models, underscoring the strength of its pretrained representations (Fig.\ref{fig:fig3}a). We further evaluated ZEN under few-shot settings (limited task-specific data conditions), using 1--5 training videos. Across all datasets, ZEN demonstrated competitive performance relative to other models (Fig.\ref{fig:fig3}d and Supplementary Tables~\cref{tab:cholec80-fewshot-1,tab:cholec80-fewshot-2,tab:cholec80-fewshot-3,tab:cholec80-fewshot-4,tab:cholec80-fewshot-5,tab:mbp-fewshot-1,tab:mbp-fewshot-2,tab:mbp-fewshot-3,tab:mbp-fewshot-4,tab:mbp-fewshot-5,tab:fewshot-auto-1,tab:fewshot-auto-2,tab:fewshot-auto-3,tab:fewshot-auto-4,tab:fewshot-auto-5}). Additional results for phase-level precision and recall are included in the aforementioned Supplementary Tables.

\paragraph{Surgical action triplet recognition}
Surgical action triplet recognition was evaluated on two procedures using the open-access CholecT50~\cite{cholect50} dataset (cholecystectomy) and an in-house LLS48~\cite{lls48} dataset (hepatectomy). Both datasets provide frame-level multi-label annotations of surgical actions, with each action represented as a combination of instrument (I), verb (V), and target (T). ZEN demonstrated superior performance across the two datasets in both full fine-tuning and frozen-backbone settings. Under the full fine-tuning setting, ZEN attained the highest full triplet (IVT) mean average precision (mAP), outperforming the second-best model, SurgeNet~\cite{surgfm5}, by +5.2\% on CholecT50 ($P = 0.004$) and +6.9\% on LLS48 ($P = 0.003$) (Fig.\ref{fig:fig3}b and Supplementary Tables~\cref{tab:cholecT50-finetune,tab:lls48-finetune}). ZEN also achieved the highest performance on individual and paired sub-components (I, V, T, IV, and IT). In the frozen-backbone setting, ZEN achieved the top performance for all component (all $P \leq 0.002$) (Extended Fig.\ref{fig:Extended_Data_Fig2}b and Supplementary Tables~\cref{tab:cholecT50-freeze,tab:lls48-freeze}). Together, these results demonstrate ZEN’s robust capability for fine-grained surgical workflow understanding at the action level.

\paragraph{Skill assessment}
We evaluated ZEN's skill assessment performance using the critical view of safety (CVS), a widely adopted safety concept for laparoscopic cholecystectomy proposed by Strasberg~\cite{strasberg}. CVS is defined by three criteria: (1) only two structures are clearly identified entering the gallbladder; (2) the lower third of the gallbladder is dissected off the liver bed; and (3) the hepatocystic triangle is completely cleared to expose all cystic structures. This task was evaluated on the Cholec80-CVS~\cite{cholec80cvs} dataset with expert annotations of CVS achievement. The three CVS criteria were formulated as a multi-label format, with each criterion predicted as satisfied or not. Overall, ZEN achieved strong performance in both the full fine-tuning and frozen-backbone settings. In the full fine-tuning setting, ZEN reached an mAP of 0.386 and a macro F1 score of 0.332, showing performance comparable to PeskaVLP (mAP 0.382; $P = 0.661$) and higher performance than the ImageNet-pretrained model~\cite{vit} (macro F1 0.298; $P = 0.060$) (Fig.\ref{fig:fig3}c). In the frozen-backbone setting, ZEN achieved an mAP of 0.382 and a macro F1 of 0.370, comparable to PeskaVLP (mAP 0.369; $P = 0.517$) and SurgeNet (macro F1 0.327; $P = 0.342$) (Extended Data Fig. \ref{fig:Extended_Data_Fig2}c). Notably, even without task-specific fine-tuning, the frozen ZEN model obtained higher mAP and macro F1 than any other fully fine-tuned model, highlighting the strength of its pretrained representations (Supplementary Table \cref{tab:cholec80-cvs}).

\begin{figure}[!t]
  \centering
  \includegraphics[width=0.9\linewidth]{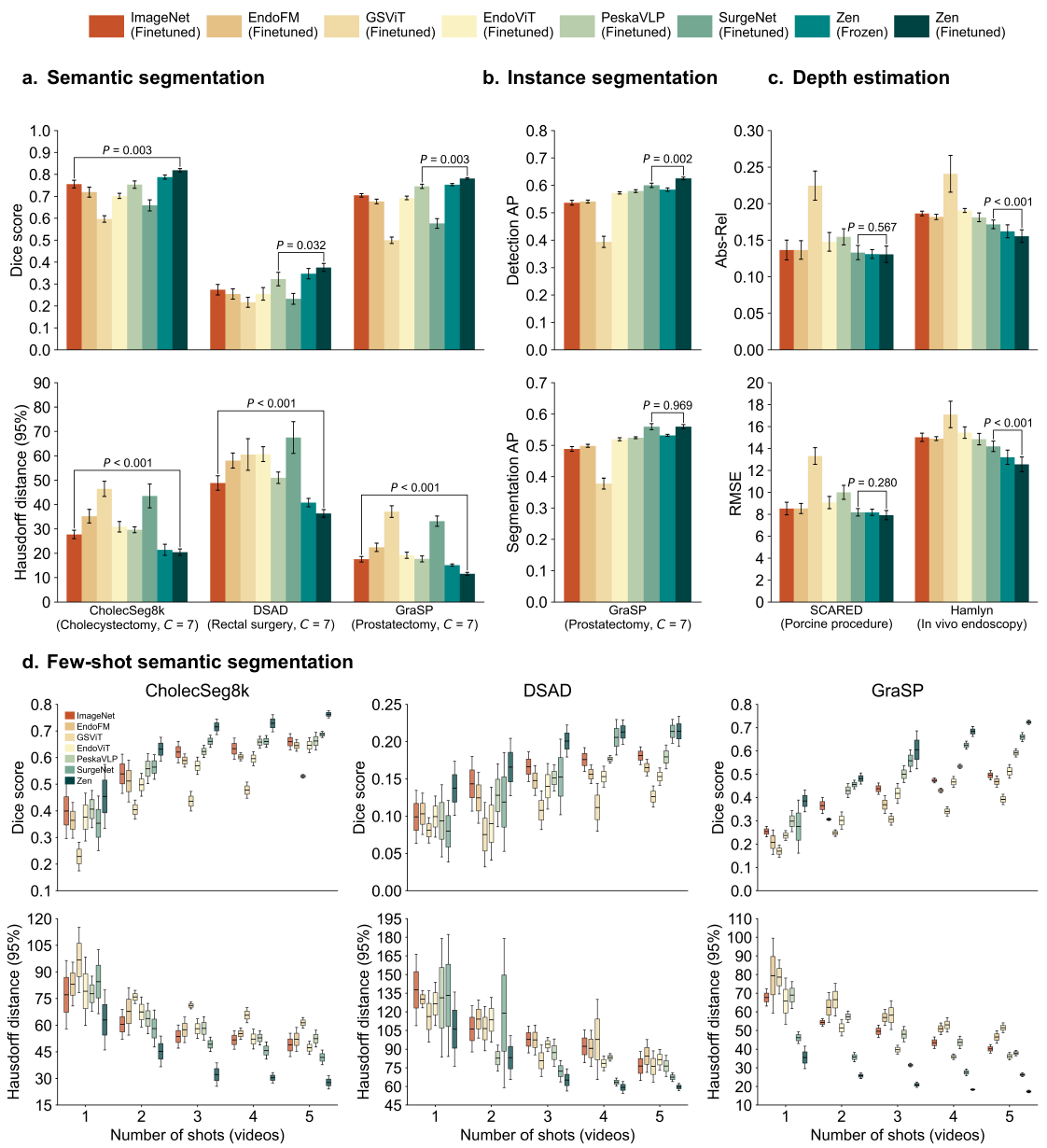}
  \caption{\textbf{Performance comparison for dense spatial understanding.}
\textbf{a,} Semantic segmentation performance across three datasets in the fine-tuned backbone setting. Performance is measured using Dice score and 95\% Hausdorff distance. $C$ denotes the number of classes.
\textbf{b,} Instance segmentation performance in the fine-tuned backbone setting. Performance is evaluated using mean Average Precision (mAP), averaged over Intersection over Union thresholds from 0.5 to 0.95. Results are reported for bounding box detection (Detection AP) and segmentation masks (Segmentation AP). $C$ denotes the number of classes.
\textbf{c,} Depth estimation performance across two datasets in the fine-tuned backbone setting. Metrics include Absolute Relative Error (Abs-Rel) and Root Mean Squared Error (RMSE).
\textbf{d,} Few-shot semantic segmentation performance (label efficiency) using 1--5 training videos across three datasets. Data represent five independent runs ($n=5$) for each model and shot condition. The center of the box represents the mean, the bounds of the box indicate standard error, and the whiskers denote the lower and upper bounds of the 95\% confidence interval (CI). For \textbf{a--c}, error bars represent 95\% CIs over five independent runs ($n=5$). $P$ values were calculated using two-sided paired $t$-tests.}
  \label{fig:fig4}
\end{figure}

\subsection*{Dense spatial understanding}
Unlike open surgery, MIS relies exclusively on video data for intraoperative guidance and postoperative analysis. Consequently, the accurate extraction of spatial and physical information from surgical videos—such as anatomical structure localization~\cite{anatomyseg1,anatomyseg2}, tool trajectories~\cite{toolseg,cholectrack} and depth estimation~\cite{depth1,depth2,depth3}—is critical for advanced clinical applications, including surgical navigation and automated skill assessment. To assess pixel-, instance-level, and geometric understanding, we evaluated ZEN on three dense prediction tasks: semantic segmentation of tools and anatomy, instance segmentation of surgical instruments, and monocular depth estimation.

\paragraph{Semantic segmentation}
To evaluate pixel-level semantic understanding, we used three public datasets: CholecSeg8k~\cite{cholecseg8k} (cholecystectomy; instruments and anatomy), DSAD~\cite{dsad} (rectal surgery; abdominal organs), and GraSP~\cite{grasp} (prostatectomy; instruments). Under the full fine-tuning protocol, ZEN outperformed all other models across all datasets. ZEN achieved higher Dice scores than the second-best model by 6.3\% on CholecSeg8k ($P = 0.003$), 5.3\% on DSAD ($P = 0.032$), and 3.6\% on GraSP ($P = 0.003$). In parallel, ZEN achieved lower 95\% Hausdorff distances by 7.26, 12.50, and 6.02, respectively (all $P < 0.001$; Fig.\ref{fig:fig4}a and Supplementary Table~\ref{tab:semseg-finetune}). Under the frozen-backbone setting, ZEN also achieved higher performance than the second-best model across all datasets ($P < 0.05$) (Extended Fig.~\ref{fig:Extended_Data_Fig3}a and Supplementary Table~\ref{tab:semseg-frozen}). Notably, frozen ZEN surpassed all other fully fine-tuned other models, indicating that ZEN learns transferable spatial semantics during pretraining. Furthermore, in few-shot settings (1--5 videos), ZEN consistently achieved the best performance across all three datasets (Fig.\ref{fig:fig4}d and Supplementary Table~\cref{tab:seg_1shot,tab:seg_2shot,tab:seg_3shot,tab:seg_4shot,tab:seg_5shot}). Qualitative segmentation examples are provided in Extended Fig.~\ref{fig:Extended_Data_Fig4}.

\paragraph{Instance segmentation}
We evaluated instance-level instrument localization on the GraSP~\cite{grasp} dataset using mAP computed over intersection-over-union thresholds from 0.5 to 0.95 for both bounding box (bbox) detection and mask segmentation. In the full fine-tuning setting, ZEN achieved the best performance for both bbox detection and mask segmentation. Specifically, ZEN surpassed the second-best model, SurgeNet, by 2.5\% in bbox detection ($P = 0.002$) and by 0.16\% in segmentation ($P = 0.969$) (Fig.\ref{fig:fig4}b). This trend continued in the frozen-backbone setting, where ZEN outperformed SurgeNet by 4.4\% in bbox detection ($P = 0.001$), with comparable segmentation performance (+0.2\%; $P = 0.716$) (Extended Fig.\ref{fig:Extended_Data_Fig3}b). Results for both full fine-tuning and frozen-backbone settings are summarized in Supplementary Table~\ref{tab:grasp-instance}. Qualitative examples are provided in Extended Fig.\ref{fig:Extended_Data_Fig4}.

\paragraph{Depth estimation}
We also assessed the model’s capacity to capture 3D geometric structures via monocular depth estimation using the SCARED~\cite{scared} (porcine procedure) and Hamlyn~\cite{hamlyn} (in vivo endoscopy video) datasets. In the full fine-tuning setting, ZEN achieved the best performance in terms of Absolute Relative error (Abs-Rel) and Root Mean Square Error (RMSE). On the SCARED dataset, ZEN demonstrated performance comparable to the second-best model, SurgeNet, recording an Abs-Rel of 0.131 and RMSE of 7.92 (SurgeNet: 0.133 and 8.18; $P > 0.05$). On the Hamlyn dataset, ZEN achieved significantly lower errors than SurgeNet, with an Abs-Rel of 0.155 and RMSE of 12.55 (SurgeNet: 0.172 and 14.20; both $P < 0.001$; Fig.\ref{fig:fig4}c and Supplementary Tables~\cref{tab:depth-scared-finetune,tab:depth-hamlyn-finetune}). In the frozen-backbone setting, ZEN consistently achieved the lowest errors across both datasets. On the SCARED dataset, ZEN outperformed the second-best model, EndoFM~\cite{surgfm2}, achieving an Abs-Rel of 0.131 and RMSE of 8.18 (EndoFM: 0.154 and 9.52; $P < 0.05$). On the Hamlyn dataset, ZEN surpassed EndoFM with an Abs-Rel of 0.155 and RMSE of 12.55 (EndoFM: 0.172 and 14.20; $P < 0.05$; Extended Data Fig.\ref{fig:Extended_Data_Fig3}c and Supplementary Tables~\cref{tab:depth-scared-frozen,tab:depth-hamlyn-frozen}). Additional metrics, including square relative error, log-scale RMSE, and accuracy under threshold, are detailed in Supplementary Tables, with qualitative examples provided in Extended Data Fig.\ref{fig:Extended_Data_Fig5}.

\begin{figure}[p]
  \centering
  \includegraphics[width=0.85\linewidth]{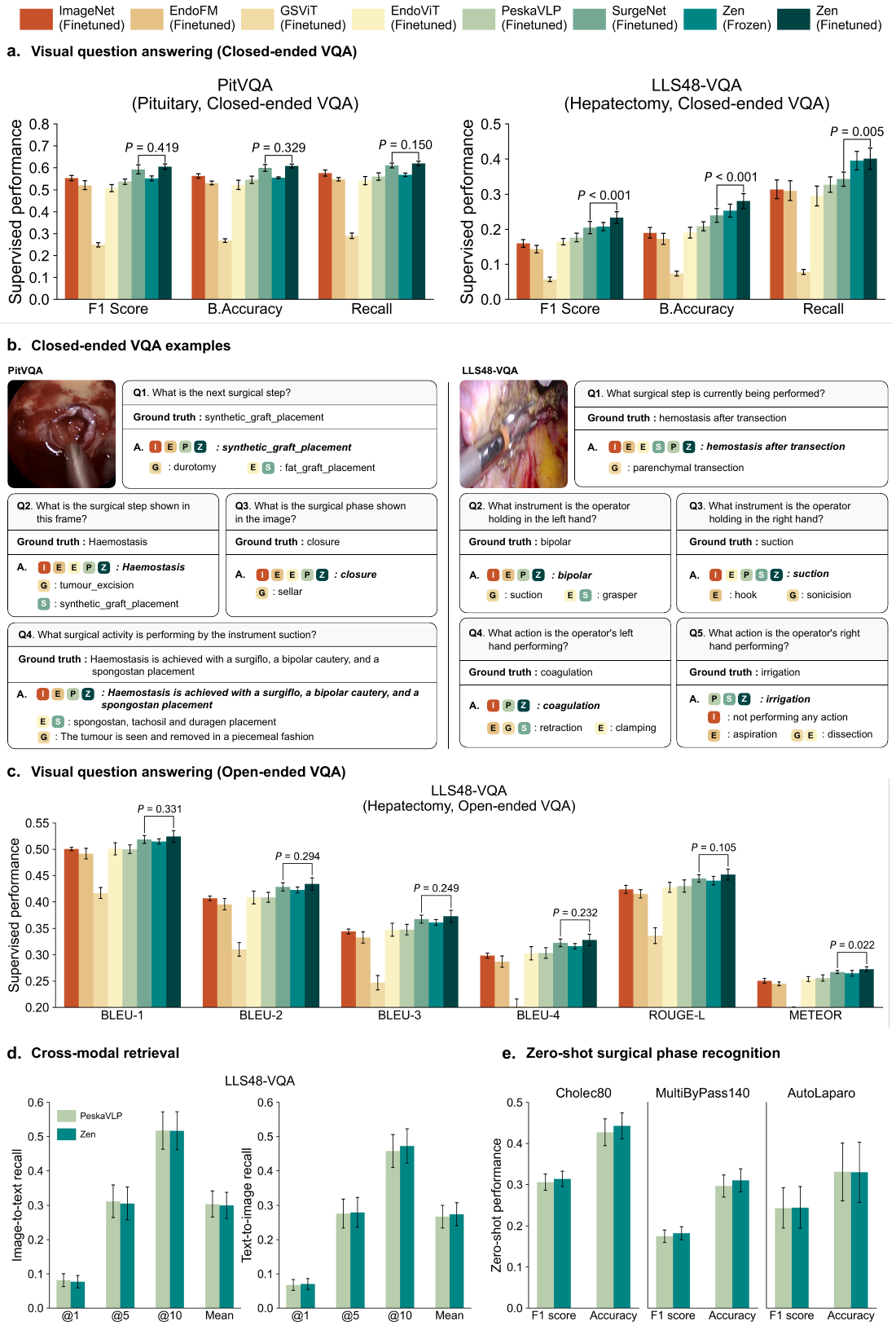}
  \caption{\textbf{Performance comparison for vision–language understanding.}
\textbf{a,} Closed-ended VQA performance in the fine-tuned backbone setting. Evaluation metrics include macro F1 score, balanced accuracy (B.Accuracy), and recall, measured by exact match criteria.
\textbf{b,} Representative closed-ended VQA examples from two datasets.
\textbf{c,} Open-ended VQA performance in the fine-tuned backbone setting, evaluated using BLEU-1--4, ROUGE-L, and METEOR.
\textbf{d,} Cross-modal retrieval performance of ZEN compared with PeskaVLP. Metrics include Recall@$K$ ($K=\{1, 5, 10\}$) and mean recall for both image-to-text and text-to-image retrieval tasks.
\textbf{e,} Zero-shot surgical phase recognition performance on three datasets. Performance is measured by video-level F1 score and accuracy. For \textbf{a} and \textbf{c}, error bars represent 95\% confidence intervals (CIs) over five independent runs ($n=5$). $P$ values were computed using two-sided paired $t$-tests. For \textbf{d} and \textbf{e}, error bars represent 95\% CIs across test videos.}
  \label{fig:fig5}
\end{figure}

\subsection*{Vision–Language understanding}
Interpreting complex surgical dynamics requires specialized expertise; however, limited availability of such expertise constrains access for medical trainees and non-expert clinicians~\cite{surgicaledu}. This bottleneck highlights a critical unmet need for automated systems capable of bridging visual data with natural language to provide accessible explanations and precise information retrieval~\cite{multimodalmedical,biomedclip}. To assess ZEN’s potential to address these clinical challenges, we evaluated its multimodal capabilities in two distinct domains. First, we assessed vision–language reasoning via Visual Question Answering (VQA)~\cite{surgicalvqa,pitvqa,surgen}. This task evaluates the model’s ability to interpret surgical scenes in response to clinical language queries. Second, we evaluated vision–language alignment through zero-shot cross-modal retrieval and zero-shot phase recognition~\cite{surgvlp,hecvlp,surgfm4}. These tasks measure how well the model aligns surgical video representations with natural language in a shared embedding space, enabling generalization without task-specific training.

\paragraph{Visual question answering}
We assessed ZEN’s VQA performance using two datasets: PitVQA~\cite{pitvqa}, which contains closed-ended VQA for endoscopic pituitary surgery, and LLS48-VQA, which includes both closed- and open-ended VQA for laparoscopic hepatectomy. Closed-ended VQA addressed categorical, clinically grounded questions—such as tool counts, instrument types, and surgical phases—and was evaluated using exact-match metrics (macro F1 score, balanced accuracy, and recall). Open-ended VQA, defined only on LLS48-VQA, targeted non-categorical questions related to surgical scene description, procedural intent, and planning. Performance was evaluated using standard text-generation metrics (BLEU-1 to BLEU-4, ROUGE-L, and METEOR). Under full fine-tuning setting, ZEN achieved the best performance on closed-ended VQA across both datasets (Fig.~\ref{fig:fig5}a and Supplementary Tables~\cref{tab:vqa-pitvqa-classification,tab:vqa-lls48-classification}). On PitVQA, ZEN achieved the highest scores across all metrics, although differences were not statistically significant. For closed-ended VQA on LLS48-VQA, ZEN achieved significantly higher macro F1 and balanced accuracy than the second-best model, with performance gains of 2.9\% ($P < 0.001$) and 4.1\% ($P < 0.001$), respectively, and also showed higher recall with an improvement of 5.8\% ($P = 0.005$). In the frozen-backbone setting, ZEN significantly outperformed the second-best model on both datasets ($P < 0.05$) (Extended Data Fig.~\ref{fig:Extended_Data_Fig6}a and Supplementary Tables~\cref{tab:vqa-pitvqa-classification,tab:vqa-lls48-classification}). For open-ended VQA on LLS48-VQA, ZEN achieved the highest scores across all text-generation metrics under both full fine-tuning (Fig.~\ref{fig:fig5}c and Supplementary Tables~\cref{tab:vqa-lls48-generation-finetune}) and frozen-backbone settings (Extended Data Fig.~\ref{fig:Extended_Data_Fig6}b and Supplementary Tables~\cref{tab:vqa-lls48-generation-freeze}). Representative qualitative examples for both closed- and open-ended VQA, including input questions, corresponding images, and model predictions, are shown in Fig.\ref{fig:fig5}b and Extended Data Fig.~\ref{fig:Extended_Data_Fig7}, respectively.

\paragraph{Vision–language alignment}
We evaluated ZEN’s vision–language alignment using two zero-shot tasks: cross-modal retrieval and phase recognition. Zero-shot cross-modal retrieval was conducted on the LLS48-VQA dataset and included both text-to-image and image-to-text retrieval, using natural-language descriptions as queries (Fig.~\ref{fig:fig5}d and Supplementary Table~\ref{tab:retrieval_results}). Zero-shot phase recognition was evaluated on three surgical phase recognition datasets (Cholec80, MultiBypass140, and AutoLaparo) using textual phase descriptions~\cite{surgfm4} (Fig.~\ref{fig:fig5}e and Supplementary Table~\ref{tab:zero-shot-phase}). Because these tasks require a shared image–text embedding space, image-only FMs could not be directly evaluated. Accordingly, we benchmarked ZEN against PeskaVLP, a vision–language foundation model that served as the teacher during multi-teacher distillation. Cross-modal retrieval performance was assessed using Recall@K (K = 1, 5, 10). In image-to-text retrieval, ZEN achieved slightly lower recall than PeskaVLP, with a mean recall of 0.300 compared with 0.304, whereas in text-to-image retrieval, ZEN achieved comparable or higher recall, reaching a mean recall of 0.274 relative to 0.267 for PeskaVLP. For zero-shot phase recognition, evaluated using video-level F1 score and accuracy, ZEN outperformed PeskaVLP on two of the three datasets. Together, these results demonstrate that the proposed multi-teacher distillation framework effectively transfers vision–language alignment from PeskaVLP to ZEN, enabling cross-modal alignment between surgical images and textual descriptions.

\subsection*{Effectiveness of multi-teacher distillation}
To assess the effectiveness of our proposed multi-teacher distillation framework, we compared ZEN against its two specialized teacher models: MIS-DINOv2 (ViT-Large) and PeskaVLP (ResNet50) (Extended Data Fig.~\ref{fig:Extended_Data_Fig8}). Performance was evaluated across 15 supervised downstream tasks. Although ZEN uses a lighter ViT-Base architecture than MIS-DINOv2 (ViT-Large), it achieved competitive performance across tasks. In the frozen-backbone setting, ZEN showed marginally lower average performance than MIS-DINOv2; however, after backbone fine-tuning, ZEN outperformed MIS-DINOv2, highlighting the effectiveness of knowledge distillation in transferring rich representations from a larger teacher to a more compact student model. This efficiency is particularly important in surgical video understanding, where models must process large volumes of video frames. Compared with PeskaVLP, ZEN outperformed the teacher across most tasks and metrics under both frozen and fine-tuned settings. For tasks involving vision–language alignment, ZEN exhibited performance comparable to PeskaVLP, despite the absence of direct vision–language supervision during training (Fig.~\ref{fig:fig5}d,e). Overall, these results show that multi-teacher distillation enables ZEN to inherit both rich visual representations and vision–language alignment from specialized teacher models, allowing a compact architecture to achieve strong performance across diverse surgical tasks.


\section*{Discussion}
In this study, we introduce ZEN, a generalizable FM for MIS video analysis, trained using a self-supervised multi-teacher distillation. We curated a large-scale surgical video dataset comprising over 4 million frames covering more than 21 surgical procedures for pretraining. We then evaluated its performance against existing surgical FMs across 20 downstream tasks covering surgical workflow, dense spatial, and vision–language understanding under full fine-tuning, frozen-backbone, few-shot and zero-shot settings. Across supervised tasks, ZEN achieved the strongest overall performance under both full fine-tuning and frozen-backbone settings. Notably, ZEN maintained competitive or superior performance even when the backbone was frozen, outperforming fully fine-tuned models on several tasks. ZEN also exhibited strong few-shot performance, indicating effective generalization with minimal labeled data. On zero-shot cross-modal alignment tasks, ZEN achieved strong performance without direct vision–language supervision, approaching that of its vision–language teacher, PeskaVLP~\cite{surgfm4}. Together, these results show ZEN not only performs strongly across a wide range of surgical tasks but also learns highly transferable representations that remain effective when labeled data or computational resources are constrained.

Despite recent progress, robust surgical video analysis remains particularly challenging. Surgical videos exhibit substantial visual variability due to differences in anatomy, technique, image conditions, and surgeon expertise~\cite{datachallenges}. In addition, procedures are inherently long and complex, with temporally evolving scenes and subtle visual cues that are often difficult to annotate consistently, even for experts. As a result, large-scale labeled datasets are expensive to obtain. In such data-constrained settings, task-specific supervised models are prone to overfitting to narrow data distributions, which often limits their ability to generalize across institutions, surgeons, and procedures. In this context, the strong performance of ZEN under frozen-backbone and few-shot settings is particularly meaningful. Frozen-backbone evaluations provide a stringent test of representation quality by assessing whether pretrained features remain general and transferable when the backbone is fixed. ZEN’s ability to outperform or match fully fine-tuned models in this setting suggests that it learns semantically rich and stable representations that capture fundamental aspects of surgical scenes, rather than overfitting to narrow task-specific cues. Similarly, strong few-shot performance indicates that ZEN can adapt to new tasks with minimal labeled data, which is critical in surgical domains with limited annotation. Importantly, ZEN demonstrated robust generalization to surgical procedures not seen during pretraining. Although the pretraining dataset was predominantly composed of abdominal laparoscopic surgeries, ZEN achieved strong performance not only on external abdominal surgical datasets but also on pituitary surgery videos, where it outperformed existing surgical FMs despite the substantial anatomical and procedural differences. Such robust generalization is essential for real-world deployment, where surgical AI systems are frequently applied to new procedures, institutions, or devices. These results underscore the potential of ZEN as a practical FM for surgical video analysis, capable of maintaining strong performance under data scarcity, domain shift.

Existing surgical FMs~\cite{surgfm1,surgfm2,surgfm3,surgfm4,surgfm5,surgfm6} have typically relied on heterogeneous combinations of open-access datasets and institution-specific private data, trained using a single pretraining strategy. As a result, such heterogeneity has obscured which approaches are most effective for surgical video understanding. Moreover, most previous work has focused primarily on workflow analysis, with comparatively less attention paid to dense spatial and geometric understanding or vision–language reasoning. Consequently, it remains unclear whether these models generalize across the wide range of complex surgical tasks encountered in real clinical practice. To address these limitations, we pretrained and evaluated multiple self-supervised learning strategies using a unified pretraining dataset and benchmark, and compared them with existing surgical FMs across a broad range of downstream tasks. This analysis revealed that different pretraining strategies offer complementary strengths. Vision-only self-supervised approaches, exemplified by MIS-DINOv2 pretrained on our dataset using SimDINOv2 with a ViT-Large backbone, showed strong performance across a broad range of downstream tasks. In contrast, models pretrained with vision–language contrastive objectives, such as PeskaVLP, demonstrated clear advantages in tasks requiring explicit zero-shot cross-modal alignment. Motivated by these observations, we integrated both forms of supervision via multi-teacher distillation, resulting in ZEN, which achieved strong and balanced performance across diverse surgical tasks. This approach highlights the potential of multi-teacher distillation as an effective strategy for building versatile medical FMs in settings where large-scale data sharing is constrained by privacy considerations, yet pretrained model weights can be shared.

Unlike many other medical domains, surgery involves substantial human intervention and limited experimental control. This makes rigorous clinical validation of individual AI methods particularly challenging and requires a gradual and exploratory approach to assessing clinical benefit~\cite{surgexpect}. In this context, recent research has begun to introduce surgical AI systems into the operating room to evaluate their real-world clinical impact. Most current deployments focus on perceptual assistance, such as real-time anatomical segmentation to support intraoperative awareness. Although these systems are starting to enter clinical practice, their deployment remains limited, and their impact on patient outcomes is still under active investigation~\cite{rctsurg}. In parallel, recent research has increasingly explored surgical AI systems that engage more directly with the physical and cognitive processes of surgery. These efforts include multimodal and interactive interfaces~\cite{endochat,surgvlm}, as well as physically grounded AI systems~\cite{robonurse,srth} that integrate perception, spatial reasoning, and surgeon interaction. A central bottleneck in this line of work lies in accurately modeling the surgeon's intraoperative interactions with instruments and anatomy in a manner that generalizes across procedures and settings. Addressing this challenge requires coordinated automated capabilities spanning surgical workflow understanding, dense spatial and semantic perception of anatomy and instruments, and language-based interaction with the surgeon. Across these approaches, a robust and generalizable visual representation is a critical prerequisite. Together, these requirements highlight the need for generalizable foundation models that perform reliably across diverse perceptual tasks. By demonstrating strong and balanced performance across surgical workflow, dense spatial, and vision–language understanding, ZEN may provide a useful foundation for future surgical AI systems, supporting both rigorous clinical impact analysis and the development of more interactive and treatment-relevant applications.

Several limitations of this study should be acknowledged. First, although our pretraining dataset spans a wide range of procedures, it is predominantly composed of abdominal laparoscopic surgeries and may not fully capture the variability of robotic-assisted or non-abdominal endoscopic interventions. While our results show encouraging generalization to anatomically and procedurally distinct settings, incorporating a broader range of non-abdominal and robotic surgical videos during pretraining will be important to further improve coverage and robustness. Second, while our benchmark is broader than those used in previous studies, the 20 downstream tasks are derived from datasets covering nine surgical procedures, which still represent a relatively limited procedural scope. As a result, certain surgical specialties and rarer interventions are underrepresented. Third, the tasks considered in this study focus primarily on surgical video analysis and therefore provide limited insight into direct clinical impact, such as effects on patient outcomes or therapeutic decision-making. Future work could address these limitations by expanding data collection through collaboration with multiple institutions and surgical specialties. Beyond video analysis, an important next step is to link visual understanding models with clinically meaningful endpoints, including patient outcomes and intraoperative decision support. Integrating such models into intervention-oriented systems, including robotic-assisted surgery and emerging physical AI frameworks, may help translate advances in surgical video understanding into direct benefits for patient care.


\section*{Methods}

\subsection*{Pretraining dataset}
Developing a generalizable surgical FM requires a large-scale and diverse pretraining dataset. We curated a dataset comprising over 4,780 videos spanning more than 21 surgical procedures across 10 organ systems, including porcine procedures used for surgical training. The dataset consists of 100 in-house hepatectomy cases and 14 publicly available open-access surgical video datasets, with detailed statistics on the number of videos and frames provided in Supplementary Table~\ref{tab:upstream-datasets}. The in-house hepatectomy videos were collected under institutional review board approval (IRB No. SMC2024-08-027). The raw videos were processed using a standardized preprocessing pipeline consisting of frame sampling and data filtering. Video frames were uniformly sampled at 1 frame per second using FFmpeg. To ensure data quality and remove irrelevant content common in raw surgical videos, such as out-of-body views or blank frames, all sampled frames underwent a two-stage filtering process. First, a deep learning–based detector was applied to automatically remove frames in which the surgical field was not visible~\cite{oob}. This automated filtering was followed by manual verification to exclude any remaining irrelevant frames, including blank frames or segments depicting non–minimally invasive procedures. The resulting pretraining dataset comprised a total of 4,316,410 frames.

\subsection*{Comparative benchmarking of self-supervised learning methods}
To systematically compare different pretraining strategies for surgical video representation learning, we conducted a comparative study of five self-supervised learning methods spanning diverse paradigms: masked autoencoders (MAE~\cite{mae}), contrastive learning (MoCoV3~\cite{mocov3}), masked siamese networks (MSN~\cite{msn}), and two distillation approaches (SimDINO and SimDINOv2~\cite{simdino}). All methods were evaluated using a consistent ViT-Base backbone (13 transformer blocks, 768-dimensional embeddings) with a \(224 \times 224\) input resolution and initialized with publicly available ImageNet-pretrained weights. Implementations were based on official code, with MAE adapted from the EndoViT~\cite{surgfm3} repository. For MAE, checkpoint selection was based on validation performance on Cholec80 (8 videos) and MultiByPass140 (10 videos), with the checkpoint achieving the lowest reconstruction error selected. These validation videos were excluded from the upstream pretraining corpus and downstream test sets, but included in the downstream training set. For the other SSL methods, default configurations were used, with checkpoint selection guided by validation performance on two downstream tasks: semantic segmentation (CholecSeg8k) and surgical action triplet recognition (CholecT50). Among all evaluated methods, SimDINOv2 achieved the strongest performance in the frozen-backbone setting across most downstream tasks (Extended Data Fig.~\ref{fig:Extended_Data_Fig1}a,b). Motivated by this result, we selected SimDINOv2 as the pretraining strategy and applied it to a ViT-Large architecture (24 transformer blocks, 1,024-dimensional embeddings, \(224 \times 224\) input resolution), resulting in the higher-capacity model, MIS-DINOv2. All models were trained using three NVIDIA H100 GPUs (80 GB each). Batch size and other hardware-dependent parameters were adjusted to fit the available resources, while all remaining hyperparameters followed the original implementations; links to the code are provided in the Computing hardware and software setting section.

\subsection*{Pretraining with the multi-teacher distillation}
ZEN was trained using a multi-teacher distillation strategy that integrates the unique properties of different expert models into a single model. Specifically, we distilled knowledge from two surgical-domain expert teachers into a ViT-Base student model with an input resolution of \(224 \times 224\) pixels. The teachers were: (1) our MIS-DINOv2 (ViT-Large) model, serving as the vision expert for high-fidelity spatial representations, and (2) PeskaVLP, a ResNet50-based vision–language model pretrained with CLIP-style contrastive learning on approximately 26,000 surgical video–narration pairs, providing robust zero-shot recognition and cross-modal retrieval capabilities. To transfer knowledge from the teachers to the student, we adopted a feature-level distillation scheme that aligns the final-stage representations of each teacher with those of the student. For the vision expert teacher (MIS-DINOv2, ViT-Large), distillation was performed by matching both \([\text{CLS}]\) token and the patch token embeddings with those of the student. For the vision–language expert teacher (PeskaVLP, ResNet50), we distilled knowledge by aligning the mean of the student’s patch tokens with the teacher’s final feature representation obtained through global average pooling (GAP). 

To facilitate this alignment, we introduced three separate adaptors, each ensuring compatibility between the student and teacher feature spaces. Each adaptor consisted of two linear layers with a LayerNorm~\cite{layernorm} and GELU~\cite{gelu} activation in between, following the adaptor design used in AM-RADIO~\cite{amradio}. The adaptor input dimension matched the student’s embedding dimension. The intermediate dimension was set to 2{,}048, and the output dimension matched that of the corresponding teacher. The distillation loss was defined at the feature level and then aggregated across feature types and teachers. For each student feature \(z_{s,f}\) distilled from teacher \(t\), the loss was formulated as:
\begin{equation}
L_{t,f}(x) = \alpha \, L^{\text{cos}}\bigl(h_t(z_{s,f}), y_{t,f}\bigr) 
           + \beta \, L^{\text{smooth-L1}}\bigl(h_t(z_{s,f}), y_{t,f}\bigr),
\end{equation}
where \(f\) denotes the feature type (\([\text{CLS}]\) token or patch tokens), \(h_t(z_{s,f})\) is the student feature after passing through the adaptor for teacher \(t\), and \(y_{t,f}\) is the corresponding teacher feature. We set \(\alpha = 0.9\) and \(\beta = 0.1\), assigning greater weight to cosine similarity while retaining the contribution of the smooth-L1 term. This loss formulation follows the feature-level distillation objective used in AM-RADIO. To mitigate inconsistencies in feature statistics between teachers that could degrade performance, all teacher features were standardized to zero mean and unit variance before loss computation. These statistics were estimated on-the-fly during training using an exponential moving average, following the feature standardization strategy introduced in UNIC~\cite{unic}.

For the vision teacher (MIS-DINOv2, ViT-Large), the final teacher loss was obtained by averaging the \([\text{CLS}]\) and patch token losses:
\begin{equation}
L_{\text{DINOv2}}(x) = \frac{1}{2}\Bigl(L_{\text{DINOv2,CLS}}(x) + L_{\text{DINOv2,patch}}(x)\Bigr).
\end{equation}
For the vision–language teacher (PeskaVLP, ResNet50), the loss was defined by aligning the mean of the student’s patch tokens with the teacher’s global average pooling feature representation:
\begin{equation}
L_{\text{PeskaVLP}}(x) = L_{\text{PeskaVLP,pooled}}(x).
\end{equation}
The final distillation loss was defined as the sum of the individual teacher losses, with stochastic teacher dropping applied during training. Formally,
\begin{equation}
L_{\text{distill}}(x) = \sum_{t \in \{\text{DINOv2}, \text{PeskaVLP}\}} m_t \, L_t(x),
\end{equation}
where \(m_t \in \{0,1\}\) is a stochastic mask indicating whether teacher \(t\) contributes to the loss. To prevent the student from becoming biased toward a single teacher, the teacher with the smaller loss magnitude was dropped with probability \(p_{\text{drop}} = 0.25\). This probabilistic masking ensures that both teachers provide balanced supervision during training, following the stochastic teacher dropping strategy introduced in UNIC. During training, the parameters of both teacher models were kept frozen, and only the student network and adaptors were updated. Multi-teacher distillation pretraining was conducted for 200 epochs on three NVIDIA H100 (80~GB each) GPUs, with a per-GPU batch size of 512 and an input resolution of \(224 \times 224\) pixels. All experimental results reported in this study are based on the checkpoint obtained in the final pretraining epoch.

\subsection*{Baselines}
To comprehensively benchmark the performance of ZEN, we compared it against baseline models, including publicly available surgical FMs with vision-only architectures (GSViT~\cite{surgfm1}, EndoFM~\cite{surgfm2}, EndoViT~\cite{surgfm3}, and SurgeNet~\cite{surgfm5}) and a vision–language model (PeskaVLP~\cite{surgfm4}). We additionally included an ImageNet-pretrained ViT-Base model as a general-purpose visual baseline. All models were evaluated using a \(224 \times 224\) input resolution across all downstream tasks. Detailed model configurations and training details for all evaluated models are provided in Supplementary Table~\cref{tab:surg_foundation_summary}.

\subsection*{Statistical analysis}
For all supervised downstream tasks, except visual question answering (VQA), experiments were repeated over five independent runs using identical training protocols and hyperparameter settings. The test set was fixed, whereas the training and validation sets were independently sampled for each run. Performance is reported as the mean across runs. Variability was quantified using the standard deviation and the standard error, computed as the standard deviation divided by the square root of the number of runs. The 95\% confidence interval was computed as 1.96 times the standard error. For VQA tasks, no separate validation set was used. Instead, experiments were conducted over five independent runs with independently sampled train–test splits. Performance metrics were averaged across runs using the same statistical reporting procedure described above. Statistical significance for task-specific comparisons across repeated runs was assessed using two-sided paired t-tests comparing ZEN with each corresponding baseline model. To evaluate overall performance differences aggregated across tasks, we performed two-sided Wilcoxon signed-rank tests on task-level average metrics (Fig.\ref{fig:fig2}c,e). 

For cross-modal retrieval and zero-shot phase recognition, which do not involve task-specific training, statistical analysis was performed at the video level. We computed the mean and standard deviation of performance across videos, calculated the standard error by dividing the standard deviation by the square root of the number of videos, and computed the 95\% confidence interval as 1.96 times the standard error.

\subsection*{Downstream evaluation details}
In this section, we describe the experimental settings for evaluating model performance across downstream tasks. Detailed descriptions of the training protocols and evaluation metrics for each task are provided in the corresponding subsections below.

\paragraph{Surgical phase recognition}
We followed the two-stage evaluation protocol introduced in TeCNO~\cite{phase2}, consisting of frame-level feature extraction followed by temporal aggregation using a multi-stage temporal convolutional network~\cite{mstcn}. Models were evaluated under three settings: (1) frozen-backbone: The pretrained encoder was kept fixed, and frame-level features were directly extracted from videos. These features were then used to train the temporal aggregation network; (2) fine-tuning: The pretrained encoder was first fine-tuned alongside a linear classifier using frame-wise supervision. After fine-tuning, the encoder was used to extract frame-level features, which were then used to train the temporal aggregation network for temporal modeling; (3) Few-shot learning (1–5 videos; limited training data): Similar to the frozen setting, the encoder remained fixed. The temporal aggregation network was trained using features extracted from only 1 to 5 training videos to evaluate performance under limited data conditions. For fine-tuning, the encoder and linear classifier were trained for 15 epochs using cross-entropy loss, the AdamW~\cite{adamw} optimizer (learning rate: 1$\times10^{-5}$, weight decay: 0.01), and a batch size of 128, with data augmentations including flipping, shift--scale--rotate, HSV adjustments, and Gaussian noise. The temporal aggregation network was trained for 100 epochs using the Adam~\cite{adam} optimizer (initial learning rate: 7$\times10^{-4}$, weight decay: 1$\times10^{-5}$), with a step scheduler that halved the learning rate every 30 epochs. Performance was evaluated using video-level F1 score and accuracy, as well as phase-level recall, precision, and Jaccard index~\cite{phasemetric}.

\paragraph{Surgical action triplet recognition}
For this task, we appended a linear classifier to the pretrained encoder and followed the Self-Distillation~\cite{selfd}, formulating the problem as multi-label classification. The model was trained to jointly predict the composite triplet class (instrument–verb–target, IVT) as well as the individual instrument (I), verb (V), and target (T) components. We evaluated performance under two settings: (1) frozen-backbone, where the encoder was fixed and only the classifier was trained; and (2) fine-tuning, where both the encoder and classifier were trained. All experiments used the AdamW optimizer (weight decay: 0.01), binary cross-entropy loss, and a cosine annealing scheduler with a 3-epoch warm-up. Input images were resized to \(224\times224\) pixels and augmented with horizontal flipping, shift--scale--rotate, hue--saturation--value adjustments, and Gaussian noise. Training configurations varied by dataset and setting: On CholecT50, models were trained for 15 epochs (frozen) and 20 epochs (fine-tuning). The frozen setting used a base learning rate of \(2\times10^{-4}\) and minimum learning rate of \(1\times10^{-5}\), while the fine-tuning setting used \(1\times10^{-4}\) and \(1\times10^{-6}\). On LLS48, models were trained for 20 epochs (frozen) and 30 epochs (fine-tuning), with base and minimum learning rates of \(5\times10^{-4} / 1\times10^{-5}\) (frozen) and \(2\times10^{-4} / 1\times10^{-6}\) (fine-tuning), respectively. Performance was evaluated using mAP, computed for individual components (I, V, T), component pairs (IV, IT), and the full triplet (IVT).

\paragraph{Skill assessment}
Skill assessment was performed based on Strasberg’s criteria for achieving the critical view of safety (CVS) during laparoscopic cholecystectomy~\cite{strasberg}. Following COLENET~\cite{cholec80cvs}, we appended a linear classifier to the pretrained encoder and trained the model to predict whether each CVS criterion was satisfied in a multi-label setting. We evaluated performance under two settings: (1) a frozen-backbone setting, in which the encoder was kept fixed and only the classifier was trained; and (2) a fine-tuning setting, in which both the encoder and classifier were jointly optimized. All models were trained for 10 epochs using the AdamW optimizer (learning rate: 1$\times10^{-5}$, weight decay: 0.01) with binary cross-entropy loss. Input images were resized to 224$\times$224 pixels and augmented using horizontal flipping, shift--scale--rotate transformations, hue--saturation--value adjustments, and Gaussian noise. Performance was evaluated using mAP and macro F1 score.

\paragraph{Semantic segmentation}
We employed the UperNet~\cite{upernet} architecture from the mmSegmentation toolbox to perform semantic segmentation of surgical instruments and anatomy. Models were evaluated under three settings: (1) frozen-backbone, where only UperNet was trained to assess the quality of the pretrained features; (2) fine-tuning, where all parameters, including the encoder, were updated end-to-end; (3) Few-shot learning (1--5 videos; limited training data), which followed the same training strategy as the frozen setting but used a small training set. We applied a strict video-wise split to ensure that training, validation, and test sets did not share any overlapping frames from the same video. All models were trained for 20 epochs using the AdamW optimizer (weight decay: 0.01), a batch size of 64, and a cosine annealing learning rate scheduler with a 1-epoch linear warm-up. Input images were resized to \(224 \times 224\) pixels and augmented with horizontal and vertical flipping. For the fine-tuning setting, the base and minimum learning rates were set to 5$\times10^{-4}$ and 1$\times10^{-6}$, respectively. For the frozen and few-shot settings, they were set to 5$\times10^{-3}$ and 1$\times10^{-5}$. Performance was evaluated using the Dice score and 95\% Hausdorff distance.

\paragraph{Instance segmentation}
For instance segmentation of surgical instruments, we used the Mask R-CNN~\cite{maskrcnn} architecture from the Torchvision library. To isolate the effect of different pretrained backbones, all other components were kept identical across experiments, including the feature pyramid network~\cite{fpn}, box head, and mask head. Models were evaluated under two settings: (1) frozen-backbone, where only the non-backbone components (FPN and heads) were trained to assess the quality of the pretrained features; (2) Full fine-tuning, where all network parameters, including the encoder backbone, were updated. All models were trained for 100 epochs using the AdamW optimizer, a batch size of 16, and a cosine annealing learning rate scheduler with a 5-epoch linear warm-up. Input images were resized to \(224 \times 224\) pixels and augmented with horizontal and vertical flipping. In the frozen setting, the base and minimum learning rates were set to 5$\times10^{-3}$ and 1$\times10^{-5}$, respectively. In the fine-tuning setting, they were set to 3$\times10^{-4}$ and 1$\times10^{-6}$. Standard Mask R-CNN loss functions were used. Performance was evaluated using mAP, computed by averaging precision across intersection-over-union thresholds ranging from 0.5 to 0.95, for both bounding box detection and segmentation masks. All models were evaluated on the GraSP dataset using the same train, validation, and test splits as those employed for semantic segmentation.

\paragraph{Depth estimation}
We adopted the supervised evaluation protocol from Surgical-DINO~\cite{surgdino} for monocular depth estimation. This protocol uses a lightweight decoder composed of a single linear layer, which takes as input the concatenated features from four intermediate stages of the encoder. For ViT backbones, we concatenated the \([\text{CLS}]\) token with the patch tokens; for other architectures, intermediate feature maps were used directly. Models were evaluated under two settings: (1) frozen-backbone, where only the decoder was trained; (2) fine-tuning, where both the encoder and decoder were trained. During evaluation, predicted depth maps were rescaled using a median scaling method. All models were trained using scale-invariant log loss~\cite{depthloss}, the AdamW optimizer (learning rate: 1$\times10^{-5}$, weight decay: 1$\times10^{-4}$), and a batch size of 8. Input images were resized to \(224 \times 224\) pixels, and data augmentation included color jittering and random horizontal flipping. Following prior works~\cite{depth3,surgdino}, performance was evaluated using five standard depth estimation metrics: absolute relative error (Abs-Rel), square relative error (Sq-Rel), root mean square error (RMSE), log-scale RMSE (RMSE log), and accuracy under threshold ($\delta$).

\paragraph{Visual question answering (VQA)}
We adapted the nanoVLM-450M~\cite{nanovlm} framework for visual question answering by replacing its default vision encoder with each pretrained models. Input images were resized to \(224 \times 224\) pixels before being fed into the vision encoder. We evaluated two training settings: a frozen-backbone setting and a fine-tuning setting. In the frozen-backbone setting, the vision encoder was kept fixed, and only two components were updated during training: (1) a linear projection layer that mapped the final spatial features from the vision encoder to the language model input space, and (2) Low-rank adaptation~\cite{lora} (LORA) modules applied to the query and value projections of the language model. The LORA modules were configured with rank 8, alpha 8, and a dropout rate of 0.1. Models were trained for 5 epochs using the AdamW optimizer with a batch size of 16 and a cosine learning rate schedule with a linear warm-up over the first 3\% of training steps. Learning rates were set to 1$\times10^{-3}$ for the projection layer and 1$\times10^{-4}$ for the LORA modules. In the fine-tuning setting, the vision encoder was jointly optimized together with the projection layer and LORA modules, enabling end-to-end adaptation of the visual representations. In this setting, the vision encoder was trained with a learning rate of 1$\times10^{-5}$, while the projection layer and LORA modules used the same learning rates as in the frozen-backbone setting. Models were trained for 3 epochs using the same optimization strategy. For both settings, evaluation was performed on the test set using the checkpoint from the final training epoch. We evaluated both closed-ended and open-ended VQA tasks. Closed-ended VQA required selecting an answer from a predefined set of labels and was evaluated using exact match-based metrics, including macro F1 score, balanced accuracy, and recall. Open-ended VQA required generating free-form natural language responses and was evaluated using text-generation metrics, including BLEU-1--BLEU-4, ROUGE-L, and METEOR.

\paragraph{Cross-modal retrieval}
We evaluated cross-modal retrieval performance under a zero-shot inference setting, without any additional task-specific training. This evaluation was restricted to vision–language aligned models (ZEN and PeskaVLP), as image-only FMs do not provide a shared image-text embedding space. All experiments were conducted on the LLS48 dataset, which is organized at the clip level, with each clip consisting of five consecutive frames. During inference, all frames were resized to \(224 \times 224\) pixels. Following the PeskaVLP protocol, visual features were first extracted at the frame level and then aggregated into a clip-level representation via average pooling. Text embeddings were obtained using the ClinicalBERT~\cite{clinicalbert} language encoder of PeskaVLP. Cross-modal similarity was computed as the dot product between image and text embeddings. For PeskaVLP, image embeddings were obtained directly from the visual encoder. For ZEN, visual features were passed through the adaptor introduced during distillation from the PeskaVLP teacher to ensure alignment with the text embedding space. We evaluated two retrieval tasks: (1) image-to-text retrieval, which retrieves the correct textual description given an image clip; and (2) text-to-image retrieval, which retrieves the corresponding image clip given a textual prompt. In both tasks, the textual descriptions were the open-ended answers to the question ``Describe the current surgical scene'' from the LLS48-VQA dataset. Performance was measured using recall@$k$ ($k = 1, 5, 10$), defined as the proportion of queries for which the correct match appears among the top-$k$ retrieved results. We additionally report mean recall, computed as the average of recall@1, recall@5, and recall@10.

\paragraph{Zero-shot phase recognition}
We evaluated zero-shot retrieval performance following the protocol and text prompts provided in the PeskaVLP. This evaluation was restricted to vision–language aligned models (ZEN and PeskaVLP), as image-only FMs do not provide a shared image--text embedding space. Experiments were conducted on the hold-out test sets of the Cholec80, MultiByPass140, and AutoLaparo datasets, using the same splits described in the surgical phase recognition experiments. During inference, all images were resized to \(224 \times 224\) pixels. Text embeddings corresponding to each surgical phase were generated using the language encoder provided in the PeskaVLP framework. Zero-shot phase prediction was performed by computing the dot product similarity between image embeddings and all phase text embeddings, and assigning the phase with the highest similarity score. For PeskaVLP, image embeddings were extracted directly from the visual encoder. For ZEN, visual features were passed through the adaptor module introduced during distillation from the PeskaVLP teacher to ensure alignment with the text embedding space. Performance was evaluated using video-level macro F1 score and accuracy.

\subsection*{Downstream datasets}
This section details the datasets used for downstream tasks, including the surgical procedures and dataset statistics (number of videos and frames), as well as the dataset splits for each training protocol. An overview of the datasets and tasks is provided in Supplementary Table~\ref{tab:downstream-dataset}.

\paragraph{Cholec80 (7 classes)}
The Cholec80~\cite{phase1} dataset contains 80 laparoscopic cholecystectomy videos, totaling 184,578 frames. Each frame is labeled with one of 7 surgical phases. To prevent data leakage, 32 videos were held out as a test set and excluded from all pretraining stages. The remaining 48 videos were used for training and validation. For frozen and fine-tuning experiments, we split the 48 videos into 40 for training and 8 for validation (5:1 ratio). In the few-shot setting, the test set (32 videos) and validation set (8 videos) were fixed, while \(k\) videos (1 to 5) were sampled from the 40 training videos for each of five runs.

\paragraph{MultiByPass140 (12 classes)}
The MultiByPass140~\cite{multibypass} dataset contains 140 videos of laparoscopic Roux-en-Y gastric bypass surgeries, totaling 770,617 frames. Each frame is labeled with one of 12 surgical phases. We held out 40 videos as a test set, strictly excluding them from all pretraining stages. The remaining 100 videos were used for training and validation. For frozen and fine-tuning experiments, the 100 videos were split into 80 for training and 20 for validation (4:1 ratio). In the few-shot setting, the test set (40 videos) and validation set (20 videos) were fixed, while \(k\) videos (1 to 5) were sampled from the 80 training videos across five runs.

\paragraph{AutoLaparo (7 classes)}
The AutoLaparo~\cite{autolaparo} contains 21 videos of laparoscopic hysterectomy, with a total of 83,243 frames annotated across 7 surgical phases. It was not included in the pretraining corpus and was used as an external test set. Among the 21 videos, 6 were held out as a test set, and the remaining 15 were used for training and validation. For frozen and fine-tuning experiments, the 15 videos were split into 10 for training and 5 for validation (2:1 ratio). In the few-shot setting, the test set (6 videos) and validation set (5 videos) were fixed, while \(k\) videos (1 to 5) were sampled from the 10 training videos over five runs.

\paragraph{CholecT50 (100 classes)}
This CholecT50~\cite{cholect50} dataset was developed for surgical action triplet recognition and includes 50 videos, totaling 100,863 frames. Of these, 45 videos are sourced from Cholec80, with 5 additional cases included to complete the set. CholecT50 features 100 unique triplet classes, derived from 6 instruments, 10 verbs, and 15 targets. To avoid data leakage, the 40 videos overlapping with the pretraining data were used for training and validation, while the 10 non-overlapping videos served as a hold-out test set. For evaluation, the training split consisted of 35 training and 5 validation videos (7:1 ratio).

\paragraph{LLS48 (195 classes)}
The LLS48~\cite{lls48} dataset is an in-house collection of 48 laparoscopic hepatectomy videos, specifically laparoscopic left lateral sectionectomy procedures, acquired at Samsung Medical Center, totaling 9,560 frames. The videos were segmented into 1,912 five-second clips, sampled every 90 seconds across full procedures. All annotations were performed by experienced clinicians, following SAGES consensus guidelines~\cite{sages}. Each clip was labeled with 17 instruments, 16 verbs, and 38 targets, resulting in 195 unique triplet classes. This dataset was excluded from the pretraining corpus. We reserved 8 videos as a hold-out test set and used the remaining 40 for training and validation. For evaluation, the 40 videos were divided into 35 training and 5 validation videos (7:1 ratio).

\paragraph{Cholec80-CVS (3 classes)}
The Cholec80-CVS~\cite{cholec80cvs} dataset comprises 62,760 frames annotated with the critical view of safety (CVS) criteria for laparoscopic cholecystectomy. Each frame is labeled with three CVS criteria, scored on a scale from 0 to 2. Following the COLENET, we binarized these annotations by converting the scores into binary labels indicating whether each criterion was satisfied or not. The Cholec80-CVS dataset is derived from the same videos as the Cholec80 dataset. Accordingly, the dataset was split using the same video-level partitions as those used for surgical phase recognition on Cholec80, ensuring identical training, validation, and test splits.

\paragraph{CholecSeg8k (7 classes)}
This CholecSeg8k~\cite{cholecseg8k} dataset consists of 8,080 frames with pixel-level annotations for surgical instruments and anatomical structures in laparoscopic cholecystectomy. The dataset was constructed from 17 video clips, each corresponding to a different patient in the Cholec80 dataset. Following prior work, the original 13 classes were merged into 7 classes: abdominal wall, liver, gastrointestinal tract, fat, gallbladder, miscellaneous, and instrument. To prevent data leakage, we used 4 clips that do not overlap with our pretraining data as a hold-out test set. Due to this strict split, the `miscellaneous' class was absent in the test set, though present in training. The remaining 13 clips, which include overlaps with the pretraining corpus, were used for training and validation. For frozen and fine-tuning settings, they were split into 11 for training and 2 for validation (6:1 ratio). In the few-shot setting, the 4 video test set and 2 video validation set were fixed, while $k$ videos (1--5) were sampled from the 11 training clips across five runs.

\paragraph{DSAD (7 classes)}
The DSAD~\cite{dsad} dataset contains videos of robot-assisted anterior rectal resections and rectal extirpations with pixel-level semantic segmentation masks for surgical anatomy. It includes both binary and multi-class annotations; we used the multi-class subset to better reflect real-world surgical scenes. This subset consists of 23 videos with 1,430 annotated frames across 8 classes: abdominal wall, colon, liver, pancreas, small intestine, spleen, and stomach. As the dataset was not part of the pretraining corpus, it was used as an external test set. We used 5 videos as a hold-out test set, and the remaining 18 videos were used for training and validation. For frozen and fine-tuning settings, these 18 videos were divided into 14 training and 4 validation videos (7:2 ratio). In the few-shot setting, the 5 video test set and 4 video validation set were fixed, while $k$ videos (1--5) were sampled from the 14 training videos over five runs.

\paragraph{GraSP (7 classes)}
The GraSP~\cite{grasp} dataset contains 3,449 annotated frames from 13 videos of robotic-assisted radical prostatectomy. It provides labels for 7 instrument classes: bipolar forceps, prograsp forceps, large needle driver, monopolar curved scissors, suction instrument, clip applier, and laparoscopic instrument (which includes laparoscopic retraction forceps, laparoscopic suture scissors, and laparoscopic needle holder). The dataset supports both semantic and instance segmentation tasks and was used for external evaluation, as it was not included in the pretraining data. We used 3 videos as a fixed hold-out test set and the remaining 10 videos for training and validation, applying the same split across both segmentation tasks. For frozen and fine-tuning settings, these were divided into 8 training and 2 validation videos (4:1 ratio). In the few-shot setting, the 3 video test set and 2 video validation set were fixed, while $k$ videos (1--5) were sampled from the 8 training videos over five runs.

\paragraph{SCARED}
The SCARED~\cite{scared} dataset contains 35 endoscopic video clips (22,950 frames) collected from porcine cadaver surgeries. The SCARED dataset was not included in our pretraining corpus and was used as an external test set. We used 7 clips as a hold-out test set, with the remaining 28 for training and validation. For five runs, we created five distinct random splits of the 28 clips at approximately an 11:3 ratio. The same split was applied to both the frozen and fine-tuning settings.

\paragraph{Hamlyn}
The Hamlyn~\cite{hamlyn} dataset consists of 21 video clips (91,866 frames) capturing in vivo scenes from laparoscopic and endoscopic surgeries. We used 6 video clips as a fixed hold-out test set, and the remaining 15 were used for training and validation. For five independent runs, we randomly partitioned these 15 clips into 12 for training and 3 for validation (4:1 ratio). The same split was applied to both the frozen and fine-tuning settings.

\paragraph{PitVQA}
The PitVQA~\cite{pitvqa} dataset contains 25 videos of endoscopic pituitary surgeries, comprising 109,173 frames and 884,242 question--answer pairs. It includes only closed-ended questions, each requiring a selection from predefined answer classes. Annotations span 59 distinct classes, including 4 surgical phases, 15 procedural steps, 18 instruments, 3 instrument presence categories, 5 instrument positions, and 14 types of operation notes. We evaluated closed-ended VQA performance across all 59 classes. The 25 videos were split into 20 training and 5 test videos, and experiments were repeated over five runs.

\paragraph{LLS48-VQA}
The LLS48-VQA dataset comprises 22 question–answer pairs per video clip and was derived from the in-house LLS48 action triplet recognition dataset. It includes two types of questions: closed-ended and open-ended. Closed-ended questions cover tool count (5 classes), instrument classification (18), action recognition (16), target identification (38), triplet (IVT) classification (196), anatomical structure recognition (115), and surgical phase classification (5), resulting in a total of 393 classes. Open-ended questions consist of free-form queries related to scene description, current task objectives, and subsequent surgical actions. Answer generation protocols differed by question type. Closed-ended answers were manually constructed using consistent sentence templates based on action triplet recognition annotations. For open-ended VQA, answers were generated using structured prompting with GPT-4.1~\cite{chatgpt} under predefined guidelines. An experienced surgeon first provided clip-level descriptions detailing the visible anatomy, ongoing surgical actions, and expected subsequent steps, informed by the underlying action triplet annotations. Based on these expert descriptions, GPT-4.1 was prompted to generate responses to the questions ``Describe the current surgical scene'', ``What is the purpose of the current actions?'', and ``What should be done next?''. Closed-ended VQA performance was evaluated across all 393 classes, while open-ended VQA was evaluated on the three open-ended question types. The 48 videos were split into 40 for training and 8 for testing (5:1 ratio) across five runs.

\subsection*{Computing hardware and software}
Video-to-frame sampling during data preprocessing was performed using FFmpeg (v6.1.1). Self-supervised pretraining was conducted on a single node equipped with 3$\times$80~GB NVIDIA H100 GPUs, using multi-GPU training implemented via PyTorch DistributedDataParallel (DDP). Pretraining experiments were implemented in Python (v3.9.13) with PyTorch (v2.2.1, CUDA 11.8) and Torchvision (v0.17.1). Our foundation model was developed using the UNIC codebase, which is publicly available at \url{https://github.com/naver/unic}. Self-supervised learning methods were trained using their official implementations. Only the batch size was adjusted to accommodate the available hardware. Specifically, we used MAE (EndoViT; \url{https://github.com/DominikBatic/EndoViT}), MoCoV3 (\url{https://github.com/facebookresearch/moco-v3}), MSN (\url{https://github.com/facebookresearch/msn}), and SimDINO/SimDINOv2 (\url{https://github.com/RobinWu218/SimDINO}). All downstream task evaluations were conducted on 4$\times$24~GB NVIDIA RTX4090 GPUs. These experiments were implemented in Python (v3.9.19) using PyTorch (v2.2.2, CUDA 11.8) and Torchvision (v0.17.2). Task-specific implementations were adapted from publicly available repositories: surgical phase recognition (\url{https://github.com/tobiascz/TeCNO}), action triplet recognition (\url{https://github.com/IMSY-DKFZ/self-distilled-swin}), action triplet recognition metric (\url{https://github.com/CAMMA-public/ivtmetrics}), skill assessment 
(\url{https://github.com/ManuelRios18/CHOLEC80-CVS-PUBLIC}), semantic segmentation (\url{https://github.com/open-mmlab/mmsegmentation}), instance segmentation (\url{https://github.com/pytorch/vision/tree/main/torchvision/models/detection}), depth estimation (\url{https://github.com/BeileiCui/SurgicalDINO}), and visual question answering (\url{https://github.com/huggingface/nanoVLM}). Implementations of other pretrained baseline models followed the official configurations provided in their respective repositories, including EndoFM (\url{https://github.com/med-air/Endo-FM}), GSViT (\url{https://github.com/SamuelSchmidgall/GSViT}), EndoViT (\url{https://github.com/DominikBatic/EndoViT}), PeskaVLP (\url{https://github.com/CAMMA-public/SurgVLP}), and SurgeNet (\url{https://github.com/TimJaspers0801/SurgeNet}). The ImageNet-pretrained ViT-Base model was obtained from Hugging Face (\url{https://huggingface.co/timm/vit_base_patch16_224.augreg_in1k}). Statistical analyses were performed using SciPy (v1.15.3).

\bmhead{Data availability}
Most of datasets used in this study are publicly available: Cholec80 (\url{https://camma.unistra.fr/datasets/}), CholecT50 (\url{https://github.com/CAMMA-public/cholect50}), CholecSeg8k (\url{https://www.kaggle.com/datasets/newslab/cholecseg8k}), Cholec80-CVS (\url{https://github.com/ManuelRios18/CHOLEC80-CVS-PUBLIC}), hSBD-instrument (\url{https://hsdb-instrument.github.io/}), Sisvse (\url{https://sisvse.github.io/}), HeiChole (\url{https://www.synapse.org/Synapse:syn18824884/wiki/591922}), Endoscapes (\url{https://github.com/CAMMA-public/Endoscapes}), ART-Net (\url{https://github.com/kamruleee51/ART-Net}), ESAD (\url{https://saras-esad.grand-challenge.org/}), AutoLaparo (\url{https://autolaparo.github.io/}), MultiByPass140 (\url{https://github.com/CAMMA-public/MultiBypass140}), LapGyn4 (\url{https://ftp.itec.aau.at/datasets/LapGyn4}), SurgicalAction160 (\url{https://ftp.itec.aau.at/datasets/SurgicalActions160}), GLENDA (\url{https://ftp.itec.aau.at/datasets/GLENDA}), HeiCo (\url{https://www.synapse.org/Synapse:syn21903917/wiki/601992}), SurgToolLoc2022 (\url{https://surgtoolloc23.grand-challenge.org/surgtoolloc-2022-resources/}), DSAD (\url{https://www.kaggle.com/datasets/anindyamajumder/the-dresden-surgical-anatomy-dataset}), GraSP (\url{https://github.com/BCV-Uniandes/GraSP}), SCARED (\url{https://endovissub2019-scared.grand-challenge.org/}), Hamlyn (\url{https://hamlyn.doc.ic.ac.uk/vision/}), PitVQA (\url{https://github.com/mobarakol/PitVQA}) and SurgicalYoutube (\url{https://github.com/TimJaspers0801/SurgeNet}). Links to publicly available datasets are also provided in Supplementary Table~\cref{tab:public_datasets}. Access to in-house datasets is restricted due to patient privacy and data protection regulations. Access requests may be directed to the corresponding author and will be evaluated subject to institutional approval and applicable regulations.

\bmhead{Code availability}
The code and pretrained weights for ZEN will be made publicly available upon acceptance of this manuscript on GitHub.

\bmhead{Acknowledgements}
This work was supported by a grant of the Korean ARPA-H Project through the Korea Health Industry Development Institute (KHIDI), funded by the Ministry of Health \& Welfare, Republic of Korea (RS-2025-25424639); by the National Research Foundation of Korea (NRF) grant funded by the Korean government (Ministry of Science and ICT) (RS-2024-00392495); by the Future Medicine 2030 Project of Samsung Medical Center (SMX1230771); and by a grant from Samsung Medical Center (SMO1250271).

\bmhead{Author contributions}
K.P., N.O. and K.H.J. conceived the study and designed the experiments. K.P., S.P., J.S. and J.Y.K. collected the data for self-supervised learning. K.P. developed the models and conducted downstream task evaluation. K.P., Y.J. and S.L. conducted result summarization and visualization. K.P., S.P. and J.S. organized the datasets for VQA tasks. K.P., Y.J., J.S. and S.A. carried out analysis of surgical action triplet recognition and semantic segmentation. J.R., J.K., G.S.C. and N.O. provided in-house surgical videos and domain expertise. N.O. and K.H.J. supervised the study. All authors contributed to the drafting and revising of the paper.

\bmhead{Competing interests}
The authors declare no competing interests.


\bibliography{sn-bibliography}
\clearpage
\setcounter{figure}{0}
\renewcommand{\figurename}{Extended Data Fig.}
\renewcommand{\thefigure}{\arabic{figure}}

\makeatletter
\renewcommand{\theHfigure}{ED.\arabic{figure}}
\makeatother

\begin{figure}[p]
  \centering
  \includegraphics[width=0.9\linewidth]{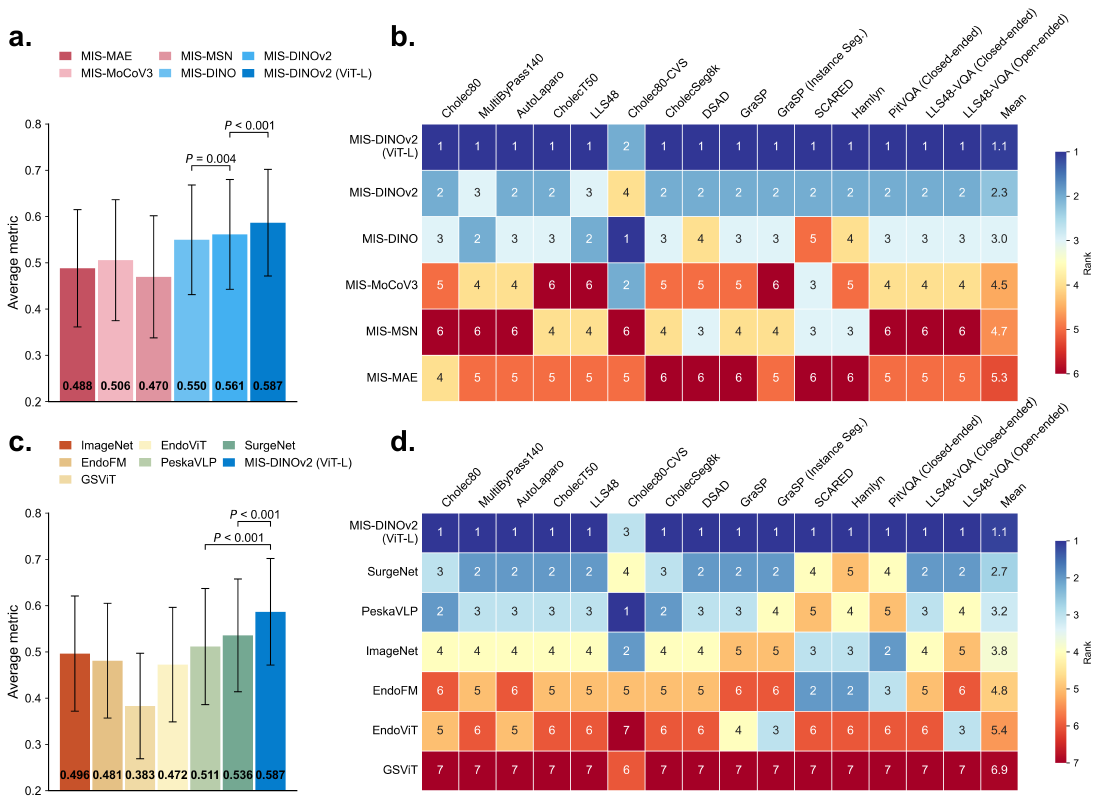}
  \caption{\textbf{Comprehensive comparison of self-supervised and existing pretrained models.}
\textbf{a,} Average performance in the frozen-backbone setting across 15 supervised downstream tasks for self-supervised pretrained models trained on minimally invasive surgical videos. Task-specific representative metrics are used, including the average of video-level macro F1 score and accuracy for surgical phase recognition; mean average precision (mAP) for surgical action triplet recognition; mAP for skill assessment; Dice score for semantic segmentation; the average of detection and segmentation mAP for instance segmentation; 1 $-$ absolute relative error for depth estimation; the average of macro F1 score and balanced accuracy for closed-ended visual question answering (VQA); and the average of BLEU, ROUGE-L, and METEOR scores for open-ended VQA.
\textbf{b,} Ranking heatmap of self-supervised pretrained models across the same 15 supervised tasks in the frozen-backbone setting, based on the corresponding representative task-level metrics.
\textbf{c,} Average performance in the frozen-backbone setting across the 15 supervised tasks for MIS-DINOv2 (ViT-L) and existing pretrained models, computed using the representative metrics as in \textbf{a}.
\textbf{d,} Ranking heatmap comparing MIS-DINOv2 (ViT-L) and existing pretrained models across the 15 supervised tasks in the frozen-backbone setting.
For \textbf{a} and \textbf{c}, error bars indicate 95\% confidence intervals. $P$ values were calculated using a two-sided Wilcoxon signed-rank test.}
  \label{fig:Extended_Data_Fig1}
\end{figure}
\clearpage

\begin{figure}[p]
  \centering
  \includegraphics[width=0.9\linewidth]{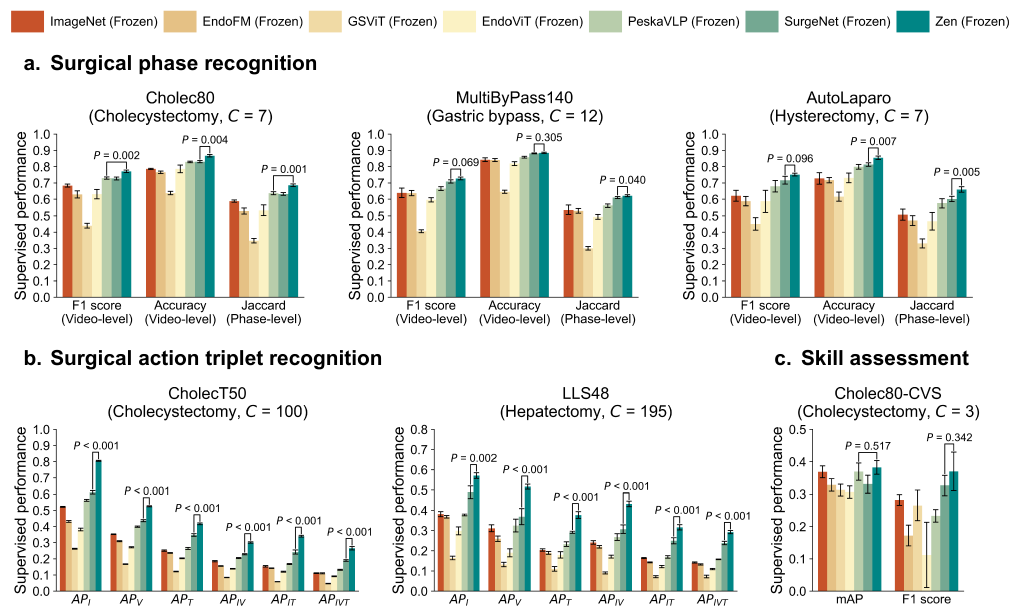}
  \caption{\textbf{Performance comparison for surgical workflow understanding.}
  \textbf{a,} Performance of ZEN and other pretrained models on surgical phase recognition across three datasets in the frozen-backbone setting. Metrics include video-level macro F1 score, accuracy, and phase-level Jaccard index. $C$ denotes the number of phases.
  \textbf{b,} Surgical action triplet recognition performance across two datasets in the frozen-backbone setting. Performance is evaluated using mean Average Precision (mAP) for instrument (I), verb (V), and target (T) components, as well as their combinations. $C$ denotes the number of triplet (IVT) classes.
  \textbf{c,} Skill assessment performance using mAP and macro F1 score in the frozen-backbone setting. $C$ denotes the number of safety criteria.
    Error bars represent 95\% confidence intervals over five independent runs ($n=5$). $P$ values were calculated using two-sided paired $t$-test.}
  \label{fig:Extended_Data_Fig2}
\end{figure}
\clearpage

\begin{figure}[p]
  \centering
  \includegraphics[width=0.9\linewidth]{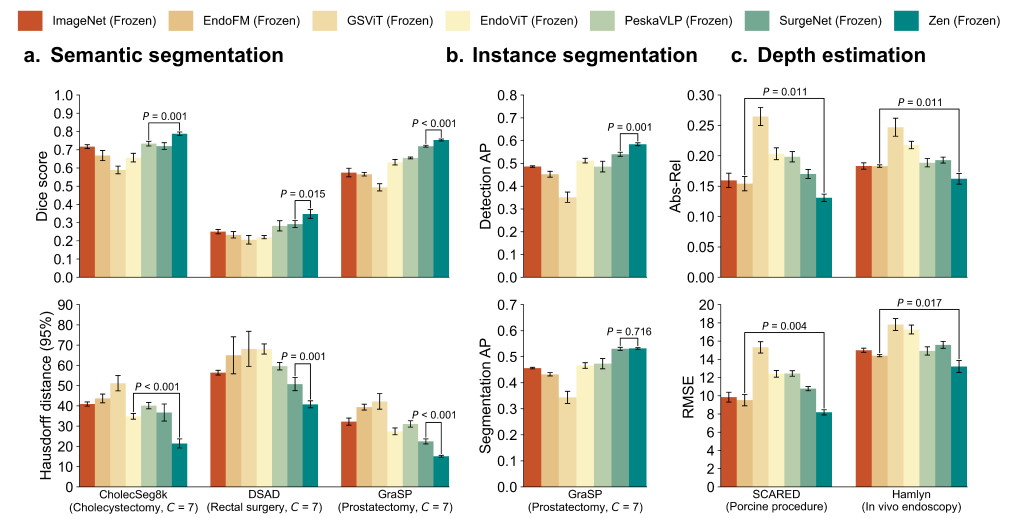}
  \caption{\textbf{Performance comparison for dense spatial understanding.}
\textbf{a,} Semantic segmentation performance across three datasets in the frozen-backbone setting. Performance is measured using Dice score and 95\% Hausdorff distance. $C$ denotes the number of classes.
\textbf{b,} Instance segmentation performance in the frozen-backbone setting. Performance is evaluated using mean Average Precision (mAP), averaged over Intersection over Union thresholds from 0.5 to 0.95. Results are reported for bounding box detection (Detection AP) and segmentation masks (Segmentation AP). $C$ denotes the number of classes.
\textbf{c,} Depth estimation performance across two datasets in the frozen-backbone setting. Metrics include Absolute Relative Error (Abs-Rel) and Root Mean Squared Error (RMSE). Error bars represent 95\% confidence intervals over five independent runs ($n=5$). $P$ values were calculated using two-sided paired $t$-tests.}
  \label{fig:Extended_Data_Fig3}
\end{figure}
\clearpage

\begin{figure}[p]
  \centering
  \includegraphics[width=0.9\linewidth]{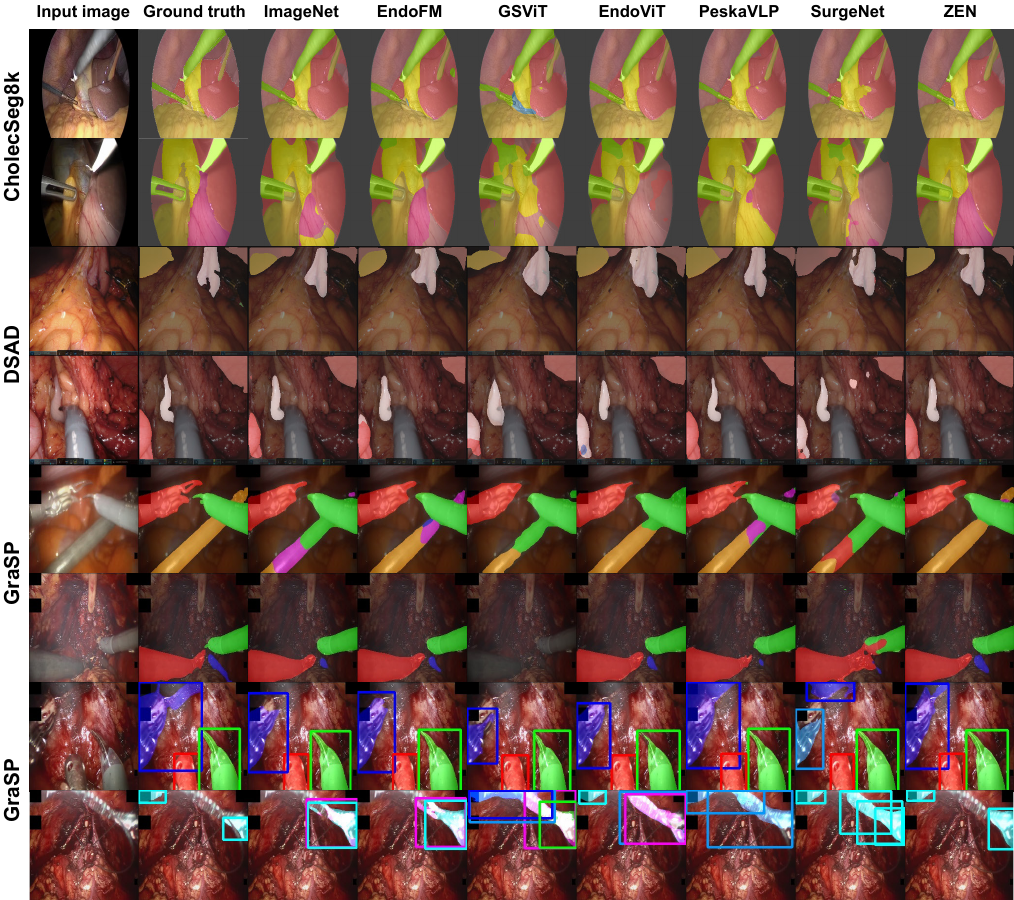}
  \caption{\textbf{Qualitative comparison of semantic and instance segmentation.}
  Visual examples comparing ZEN with other pretrained models across multiple surgical datasets.}
  \label{fig:Extended_Data_Fig4}
\end{figure}
\clearpage

\begin{figure}[p]
  \centering
  \includegraphics[width=0.9\linewidth]{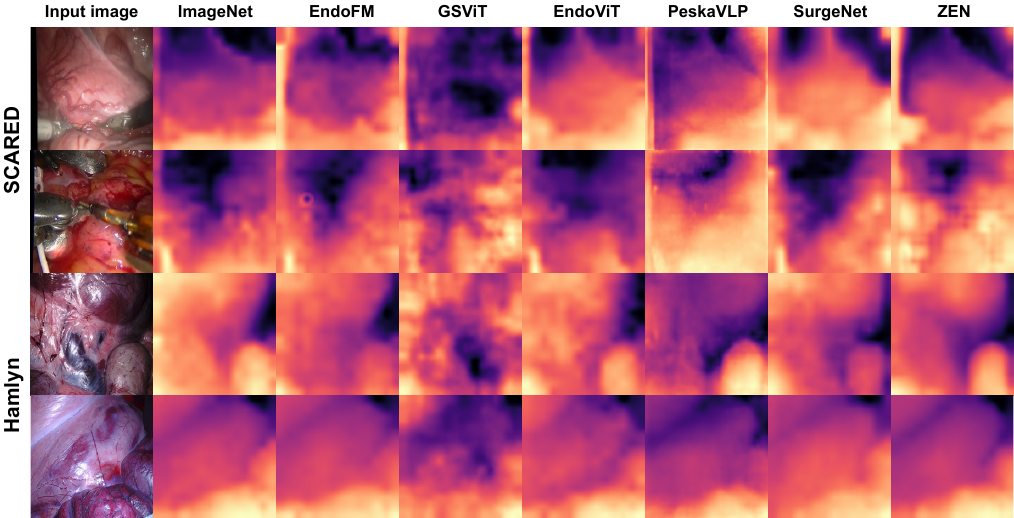}
  \caption{\textbf{Qualitative comparison of depth estimation.} 
  Visual examples comparing ZEN with other pretrained models across two surgical datasets.}
  \label{fig:Extended_Data_Fig5}
\end{figure}
\clearpage

\begin{figure}[p]
  \centering
  \includegraphics[width=0.9\linewidth]{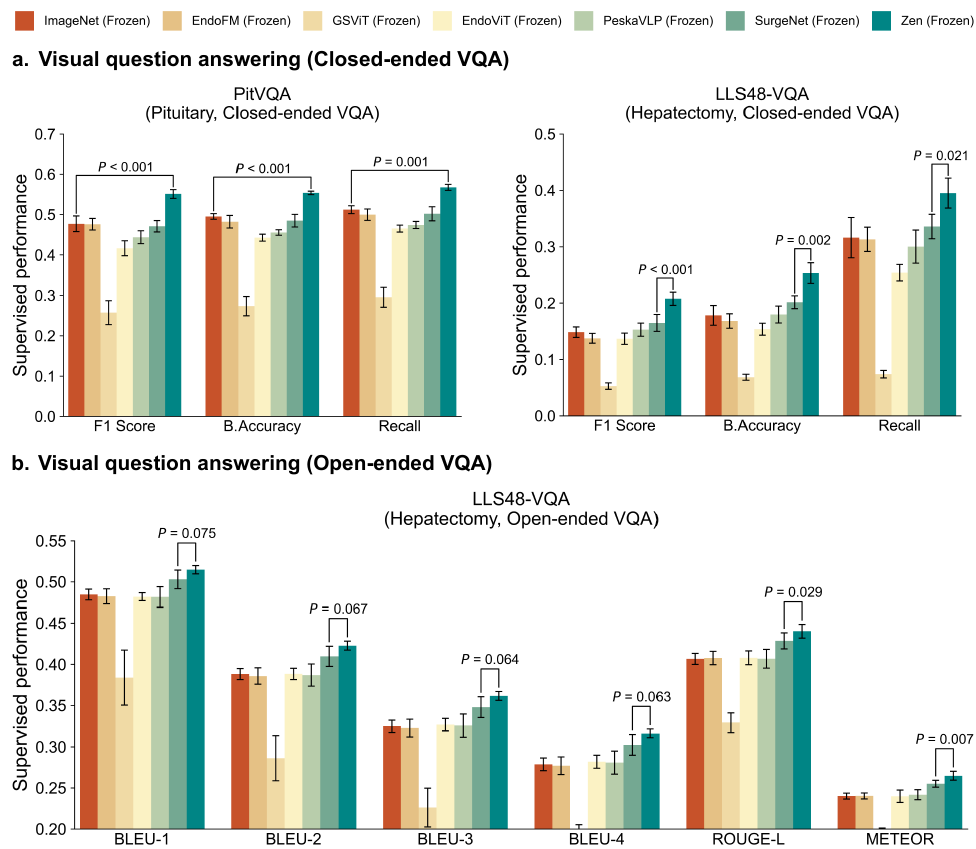}
  \caption{\textbf{Performance comparison for visual question answering.}
  \textbf{a,} Closed-ended VQA performance in the frozen-backbone setting. Evaluation metrics include macro F1 score, balanced accuracy (B.Accuracy), and recall, measured by exact match.
\textbf{b,} Open-ended VQA performance in the frozen-backbone setting, evaluated using BLEU-1--4, ROUGE-L, and METEOR. Error bars represent 95\% confidence intervals over five independent runs ($n=5$). $P$ values were calculated using two-sided paired $t$-tests.}
  \label{fig:Extended_Data_Fig6}
\end{figure}
\clearpage

\begin{figure}[p]
  \centering
  \includegraphics[width=0.9\linewidth]{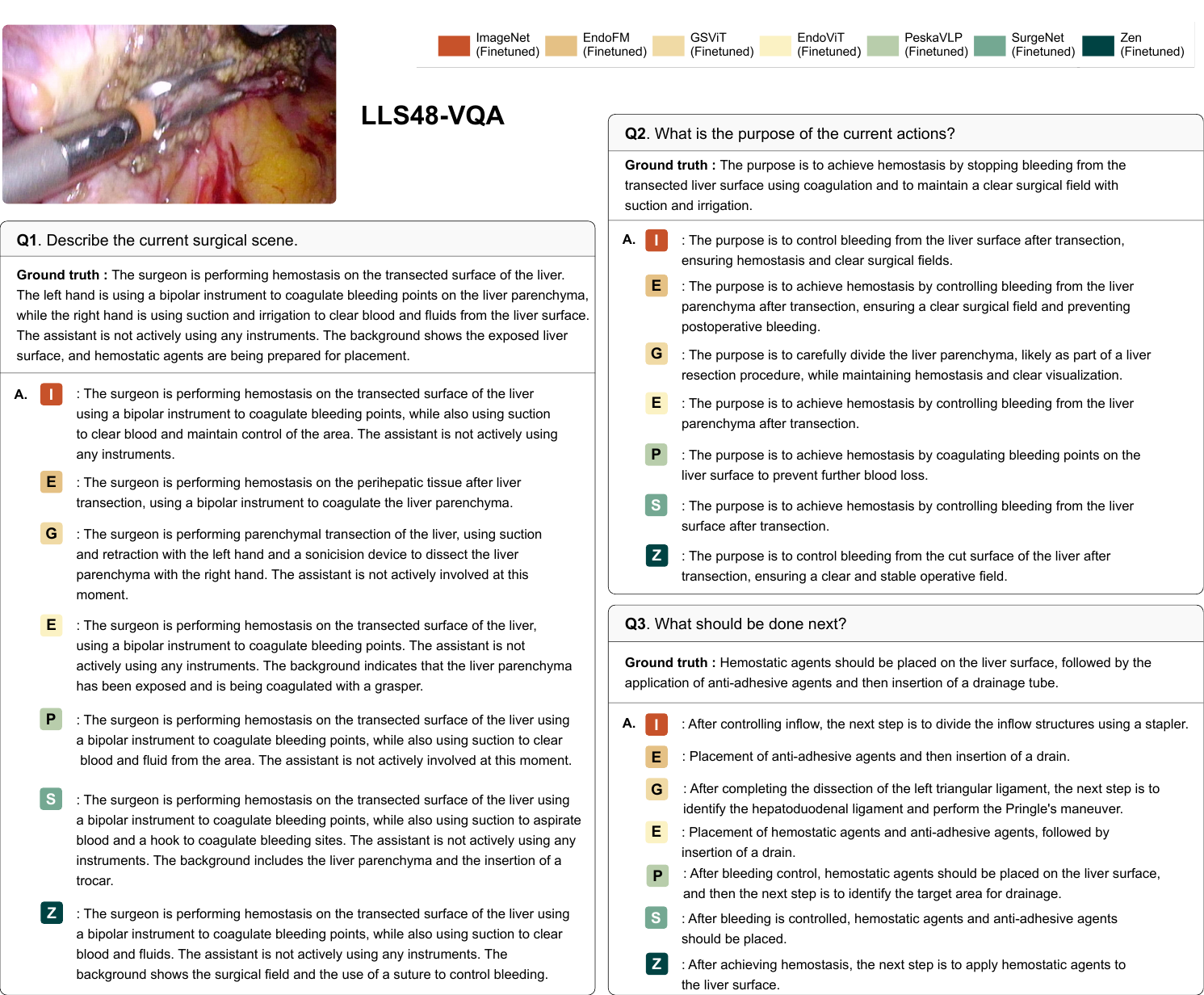}
  \caption{\textbf{Examples of Open-ended VQA.}}
  \label{fig:Extended_Data_Fig7}
\end{figure}
\clearpage

\begin{figure}[p]
  \centering
  \includegraphics[width=0.9\linewidth]{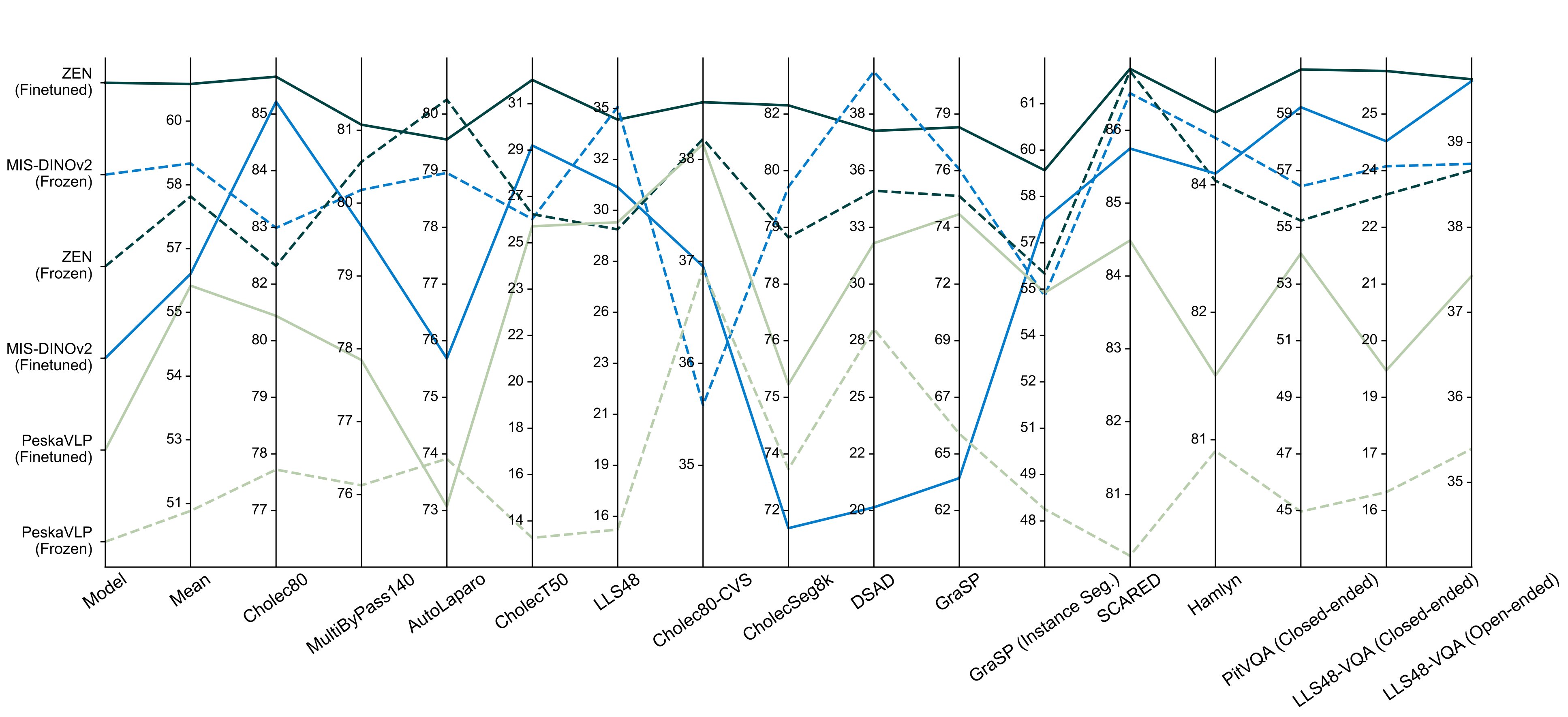}
  \caption{\textbf{Downstream performance comparison of ZEN and its teacher models.}
Performance of ZEN and its teacher models, MIS-DINOv2 and PeskaVLP, are shown across 15 supervised downstream tasks, under frozen and fine-tuned backbone settings. Each axis represents a dataset, and values correspond to task-specific representative evaluation metrics (higher is better).}
  \label{fig:Extended_Data_Fig8}
\end{figure}
\clearpage

\begin{table}[t]
\centering
\caption{\textbf{Surgical phase recognition performance comparison on the Cholec80 dataset using fine-tuned backbones.} The upper block reports results for self-supervised learning pretrained models, and the lower block reports results for existing pretrained models. Best-performing models for each metric are highlighted in bold. Values are reported as mean $\pm$ standard deviation over five independent runs.}
\label{tab:cholec80-finetune}
\setlength{\tabcolsep}{1.5pt} 
\begin{tabular}{lcccccc}
\toprule
\multirow{2}{*}{Method} & \multirow{2}{*}{Backbone} & \multicolumn{2}{c}{Video-level} & \multicolumn{3}{c}{Phase-level} \\ \cmidrule(lr){3-4} \cmidrule(lr){5-7}
                        &                           & F1 score           & Accuracy           & Precision          & Recall             & Jaccard            \\ \midrule
MIS-MAE                 & ViT-B                     & $0.7732\pm0.0246$  & $0.8680\pm0.0163$  & $0.7924\pm0.0215$  & $0.8249\pm0.0212$  & $0.6802\pm0.0285$  \\
MIS-MoCoV3              & ViT-B                     & $0.7398\pm0.0138$  & $0.8509\pm0.0104$  & $0.7727\pm0.0119$  & $0.7891\pm0.0118$  & $0.6435\pm0.0151$  \\
MIS-MSN                 & ViT-B                     & $0.7455\pm0.0189$  & $0.8566\pm0.0077$  & $0.7745\pm0.0156$  & $0.8058\pm0.0246$  & $0.6603\pm0.0183$  \\
MIS-DINO                & ViT-B                     & $0.7939\pm0.0156$  & $0.8756\pm0.0102$  & $0.7921\pm0.0208$  & $\mathbf{0.8647\pm0.0115}$ & $0.7040\pm0.0172$  \\
MIS-DINOv2              & ViT-B                     & $0.7901\pm0.0146$  & $0.8774\pm0.0111$  & $0.7924\pm0.0159$  & $0.8526\pm0.0099$  & $0.7005\pm0.0160$  \\
MIS-DINOv2              & ViT-L                     & $\mathbf{0.8065\pm0.0119}$ & $\mathbf{0.8961\pm0.0072}$ & $\mathbf{0.8146\pm0.0101}$ & $0.8622\pm0.0106$  & $\mathbf{0.7169\pm0.0144}$ \\ \midrule
ImageNet                & ViT-B                     & $0.7324\pm0.0086$  & $0.8346\pm0.0043$  & $0.7526\pm0.0065$  & $0.7941\pm0.0153$  & $0.6244\pm0.0128$  \\
EndoFM                  & ViT-B                     & $0.7175\pm0.0190$  & $0.8293\pm0.0094$  & $0.7519\pm0.0109$  & $0.7747\pm0.0217$  & $0.6113\pm0.0181$  \\
GSViT                   & EfficientViT              & $0.4248\pm0.0268$  & $0.6322\pm0.0146$  & $0.5454\pm0.0149$  & $0.4704\pm0.0327$  & $0.3356\pm0.0224$  \\
EndoViT                 & ViT-B                     & $0.7401\pm0.0067$  & $0.8416\pm0.0086$  & $0.7690\pm0.0086$  & $0.7959\pm0.0167$  & $0.6414\pm0.0074$  \\
PeskaVLP                & ResNet50                  & $0.7622\pm0.0090$  & $0.8564\pm0.0069$  & $0.7852\pm0.0072$  & $0.8195\pm0.0163$  & $0.6611\pm0.0088$  \\
SurgeNet                & CAFormer18                & $0.7742\pm0.0298$  & $0.8746\pm0.0135$  & $0.7922\pm0.0130$  & $0.8353\pm0.0372$  & $0.6822\pm0.0324$  \\
ZEN                     & ViT-B                     & $\mathbf{0.8120\pm0.0089}$ & $\mathbf{0.9004\pm0.0055}$ & $\mathbf{0.8203\pm0.0086}$ & $\mathbf{0.8690\pm0.0129}$ & $\mathbf{0.7267\pm0.0124}$ \\ \bottomrule
\end{tabular}
\end{table}

\begin{table}[t]
\centering
\caption{\textbf{Surgical phase recognition performance comparison on the MultiByPass140 dataset using fine-tuned backbones.} The upper block reports results for self-supervised learning pretrained models, and the lower block reports results for existing pretrained models. Best-performing models for each metric are highlighted in bold. Values are reported as mean $\pm$ standard deviation over five independent runs.}
\label{tab:multibypass140-finetune}
\setlength{\tabcolsep}{1.5pt} 
\begin{tabular}{lcccccc}
\toprule
\multirow{2}{*}{Method} & \multirow{2}{*}{Backbone} & \multicolumn{2}{c}{Video-level} & \multicolumn{3}{c}{Phase-level} \\ \cmidrule(lr){3-4} \cmidrule(lr){5-7}
                        &                           & F1 score           & Accuracy           & Precision          & Recall             & Jaccard            \\ \midrule
MIS-MAE                 & ViT-B                     & $0.6901\pm0.0292$  & $0.8707\pm0.0088$  & $0.6991\pm0.0299$  & $0.7473\pm0.0114$  & $0.5829\pm0.0247$  \\
MIS-MoCoV3              & ViT-B                     & $0.6889\pm0.0227$  & $0.8718\pm0.0074$  & $0.7010\pm0.0240$  & $0.7346\pm0.0131$  & $0.5820\pm0.0154$  \\
MIS-MSN                 & ViT-B                     & $0.7048\pm0.0083$  & $0.8833\pm0.0017$  & $0.7099\pm0.0157$  & $0.7737\pm0.0220$  & $0.6047\pm0.0058$  \\
MIS-DINO                & ViT-B                     & $\mathbf{0.7195\pm0.0193}$ & $0.8846\pm0.0036$  & $\mathbf{0.7418\pm0.0309}$ & $0.7705\pm0.0192$  & $\mathbf{0.6158\pm0.0085}$ \\
MIS-DINOv2              & ViT-B                     & $0.6845\pm0.0142$  & $0.8744\pm0.0079$  & $0.6958\pm0.0211$  & $0.7551\pm0.0234$  & $0.5805\pm0.0127$  \\
MIS-DINOv2              & ViT-L                     & $0.7038\pm0.0179$  & $\mathbf{0.8898\pm0.0050}$ & $0.7029\pm0.0192$  & $\mathbf{0.7831\pm0.0134}$ & $0.6068\pm0.0202$  \\ \midrule
ImageNet                & ViT-B                     & $0.6760\pm0.0303$  & $0.8672\pm0.0097$  & $0.6999\pm0.0274$  & $0.7230\pm0.0252$  & $0.5690\pm0.0197$  \\
EndoFM                  & ViT-B                     & $0.6655\pm0.0124$  & $0.8587\pm0.0076$  & $0.6891\pm0.0142$  & $0.7385\pm0.0108$  & $0.5670\pm0.0132$  \\
GSViT                   & EfficientViT              & $0.4470\pm0.0160$  & $0.6835\pm0.0196$  & $0.5015\pm0.0356$  & $0.4799\pm0.0229$  & $0.3404\pm0.0156$  \\
EndoViT                 & ViT-B                     & $0.6705\pm0.0160$  & $0.8581\pm0.0052$  & $0.6887\pm0.0218$  & $0.7262\pm0.0116$  & $0.5631\pm0.0111$  \\
PeskaVLP                & ResNet50                  & $0.6821\pm0.0183$  & $0.8748\pm0.0084$  & $0.6955\pm0.0114$  & $0.7521\pm0.0213$  & $0.5821\pm0.0175$  \\
SurgeNet                & CAFormer18                & $\mathbf{0.7278\pm0.0245}$ & $0.8851\pm0.0035$  & $0.7347\pm0.0196$  & $0.7714\pm0.0210$  & $\mathbf{0.6246\pm0.0227}$ \\
ZEN                     & ViT-B                     & $0.7273\pm0.0261$  & $\mathbf{0.8942\pm0.0042}$ & $\mathbf{0.7381\pm0.0319}$ & $\mathbf{0.7737\pm0.0116}$ & $0.6236\pm0.0163$  \\ \bottomrule
\end{tabular}
\end{table}

\begin{table}[t]
\centering
\caption{\textbf{Surgical phase recognition performance comparison on the AutoLaparo dataset using fine-tuned backbones.} The upper block reports results for self-supervised learning pretrained models, and the lower block reports results for existing pretrained models. Best-performing models for each metric are highlighted in bold. Values are reported as mean $\pm$ standard deviation over five independent runs.}
\label{tab:autolaparo-finetune}
\setlength{\tabcolsep}{1.5pt} 
\begin{tabular}{lcccccc}
\toprule
\multirow{2}{*}{Method} & \multirow{2}{*}{Backbone} & \multicolumn{2}{c}{Video-level} & \multicolumn{3}{c}{Phase-level} \\ \cmidrule(lr){3-4} \cmidrule(lr){5-7}
                        &                           & F1 score           & Accuracy           & Precision          & Recall             & Jaccard            \\ \midrule
MIS-MoCoV3              & ViT-B                     & $0.6878\pm0.0205$  & $0.7889\pm0.0122$  & $0.7279\pm0.0245$  & $0.7093\pm0.0259$  & $0.5783\pm0.0298$  \\
MIS-MSN                 & ViT-B                     & $0.7232\pm0.0191$  & $0.8325\pm0.0137$  & $0.7533\pm0.0213$  & $0.7474\pm0.0338$  & $0.6300\pm0.0229$  \\
MIS-DINO                & ViT-B                     & $\mathbf{0.7417\pm0.0157}$ & $\mathbf{0.8362\pm0.0116}$ & $\mathbf{0.7911\pm0.0181}$ & $\mathbf{0.7605\pm0.0041}$ & $\mathbf{0.6488\pm0.0156}$ \\
MIS-DINOv2              & ViT-B                     & $0.6982\pm0.0259$  & $0.8009\pm0.0258$  & $0.7659\pm0.0171$  & $0.7294\pm0.0286$  & $0.6085\pm0.0288$  \\
MIS-DINOv2              & ViT-L                     & $0.7058\pm0.0441$  & $0.8079\pm0.0258$  & $0.7636\pm0.0164$  & $0.7367\pm0.0433$  & $0.6101\pm0.0471$  \\ \midrule
ImageNet                & ViT-B                     & $0.6679\pm0.0401$  & $0.7673\pm0.0360$  & $0.7118\pm0.0315$  & $0.6907\pm0.0304$  & $0.5646\pm0.0373$  \\
EndoFM                  & ViT-B                     & $0.6391\pm0.0523$  & $0.7429\pm0.0383$  & $0.7095\pm0.0138$  & $0.6794\pm0.0460$  & $0.5330\pm0.0512$  \\
GSViT                   & EfficientViT              & $0.4206\pm0.0609$  & $0.5876\pm0.0474$  & $0.6707\pm0.0426$  & $0.4403\pm0.0607$  & $0.3109\pm0.0506$  \\
EndoViT                 & ViT-B                     & $0.6358\pm0.0232$  & $0.7285\pm0.0277$  & $0.6953\pm0.0136$  & $0.6651\pm0.0169$  & $0.5284\pm0.0248$  \\
PeskaVLP                & ResNet50                  & $0.6737\pm0.0248$  & $0.7878\pm0.0166$  & $0.7308\pm0.0175$  & $0.6876\pm0.0270$  & $0.5672\pm0.0231$  \\
SurgeNet                & CAFormer18                & $0.7277\pm0.0209$  & $\mathbf{0.8341\pm0.0107}$ & $0.7702\pm0.0172$  & $0.7362\pm0.0267$  & $0.6356\pm0.0242$  \\
ZEN                     & ViT-B                     & $\mathbf{0.7603\pm0.0218}$ & $0.8307\pm0.0222$  & $\mathbf{0.7826\pm0.0136}$ & $\mathbf{0.7597\pm0.0220}$ & $\mathbf{0.6576\pm0.0248}$ \\ \bottomrule
\end{tabular}
\end{table}

\begin{table}[t]
\centering
\caption{\textbf{Surgical phase recognition performance comparison on the Cholec80 dataset using frozen backbones.} The upper block reports results for self-supervised learning pretrained models, and the lower block reports results for existing pretrained models. Best-performing models for each metric are highlighted in bold. Values are reported as mean $\pm$ standard deviation over five independent runs.}
\label{tab:cholec80-frozen}
\setlength{\tabcolsep}{1.5pt} 
\begin{tabular}{lcccccc}
\toprule
\multirow{2}{*}{Method} & \multirow{2}{*}{Backbone} & \multicolumn{2}{c}{Video-level} & \multicolumn{3}{c}{Phase-level} \\ \cmidrule(lr){3-4} \cmidrule(lr){5-7}
                        &                           & F1 score           & Accuracy           & Precision          & Recall             & Jaccard            \\ \midrule
MIS-MAE                 & ViT-B                     & $0.7015\pm0.0196$  & $0.8293\pm0.0107$  & $0.7543\pm0.0130$  & $0.7468\pm0.0324$  & $0.6093\pm0.0216$  \\
MIS-MoCoV3              & ViT-B                     & $0.6790\pm0.0152$  & $0.8077\pm0.0108$  & $0.7329\pm0.0126$  & $0.7342\pm0.0220$  & $0.5833\pm0.0169$  \\
MIS-MSN                 & ViT-B                     & $0.5714\pm0.0354$  & $0.7573\pm0.0091$  & $0.6833\pm0.0121$  & $0.6199\pm0.0460$  & $0.4790\pm0.0328$  \\
MIS-DINO                & ViT-B                     & $0.7261\pm0.0137$  & $0.8306\pm0.0134$  & $0.7483\pm0.0209$  & $0.7935\pm0.0121$  & $0.6297\pm0.0166$  \\
MIS-DINOv2              & ViT-B                     & $0.7448\pm0.0080$  & $0.8421\pm0.0057$  & $0.7788\pm0.0147$  & $0.7957\pm0.0164$  & $0.6508\pm0.0073$  \\
MIS-DINOv2              & ViT-L                     & $\mathbf{0.7796\pm0.0120}$ & $\mathbf{0.8735\pm0.0059}$ & $\mathbf{0.7814\pm0.0136}$ & $\mathbf{0.8499\pm0.0105}$ & $\mathbf{0.6921\pm0.0127}$ \\ \midrule
ImageNet                & ViT-B                     & $0.6832\pm0.0102$  & $0.7855\pm0.0039$  & $0.7396\pm0.0094$  & $0.7278\pm0.0149$  & $0.5878\pm0.0081$  \\
EndoFM                  & ViT-B                     & $0.6291\pm0.0250$  & $0.7647\pm0.0090$  & $0.7210\pm0.0069$  & $0.6579\pm0.0287$  & $0.5300\pm0.0214$  \\
GSViT                   & EfficientViT              & $0.4396\pm0.0167$  & $0.6374\pm0.0120$  & $0.5466\pm0.0291$  & $0.4936\pm0.0162$  & $0.3470\pm0.0159$  \\
EndoViT                 & ViT-B                     & $0.6308\pm0.0325$  & $0.7860\pm0.0267$  & $0.7377\pm0.0183$  & $0.6671\pm0.0296$  & $0.5336\pm0.0357$  \\
PeskaVLP                & ResNet50                  & $0.7296\pm0.0080$  & $0.8287\pm0.0054$  & $0.7526\pm0.0109$  & $0.7923\pm0.0044$  & $0.6364\pm0.0107$  \\
SurgeNet                & CAFormer18                & $0.7270\pm0.0102$  & $0.8304\pm0.0071$  & $0.7645\pm0.0146$  & $0.7807\pm0.0106$  & $0.6319\pm0.0101$  \\
ZEN                     & ViT-B                     & $\mathbf{0.7710\pm0.0083}$ & $\mathbf{0.8672\pm0.0082}$ & $\mathbf{0.7870\pm0.0140}$ & $\mathbf{0.8339\pm0.0171}$ & $\mathbf{0.6853\pm0.0098}$ \\ \bottomrule
\end{tabular}
\end{table}

\begin{table}[t]
\centering
\caption{\textbf{Surgical phase recognition performance comparison on the MultiByPass140 dataset using frozen backbones.} The upper block reports results for self-supervised learning pretrained models, and the lower block reports results for existing pretrained models. Best-performing models for each metric are highlighted in bold. Values are reported as mean $\pm$ standard deviation over five independent runs.}
\label{tab:multibypass140-freeze}
\setlength{\tabcolsep}{1.5pt} 
\begin{tabular}{lcccccc}
\toprule
\multirow{2}{*}{Method} & \multirow{2}{*}{Backbone} & \multicolumn{2}{c}{Video-level} & \multicolumn{3}{c}{Phase-level} \\ \cmidrule(lr){3-4} \cmidrule(lr){5-7}
                        &                           & F1 score           & Accuracy           & Precision          & Recall             & Jaccard            \\ \midrule
MIS-MAE                 & ViT-B                     & $0.6614\pm0.0076$  & $0.8537\pm0.0076$  & $0.6746\pm0.0095$  & $0.7133\pm0.0181$  & $0.5629\pm0.0058$  \\
MIS-MoCoV3              & ViT-B                     & $0.6610\pm0.0100$  & $0.8598\pm0.0060$  & $0.6779\pm0.0081$  & $0.7351\pm0.0224$  & $0.5642\pm0.0110$  \\
MIS-MSN                 & ViT-B                     & $0.6183\pm0.0119$  & $0.8316\pm0.0036$  & $0.6333\pm0.0077$  & $0.6627\pm0.0113$  & $0.5134\pm0.0071$  \\
MIS-DINO                & ViT-B                     & $0.7076\pm0.0166$  & $0.8781\pm0.0057$  & $0.7097\pm0.0122$  & $\mathbf{0.7592\pm0.0268}$ & $0.5985\pm0.0159$  \\
MIS-DINOv2              & ViT-B                     & $0.7043\pm0.0155$  & $0.8740\pm0.0059$  & $0.7143\pm0.0184$  & $0.7508\pm0.0078$  & $0.5973\pm0.0147$  \\
MIS-DINOv2              & ViT-L                     & $\mathbf{0.7221\pm0.0181}$ & $\mathbf{0.8815\pm0.0085}$ & $\mathbf{0.7256\pm0.0171}$ & $0.7574\pm0.0209$  & $\mathbf{0.6156\pm0.0171}$ \\ \midrule
ImageNet                & ViT-B                     & $0.6387\pm0.0333$  & $0.8425\pm0.0133$  & $0.6531\pm0.0168$  & $0.6940\pm0.0257$  & $0.5365\pm0.0314$  \\
EndoFM                  & ViT-B                     & $0.6373\pm0.0183$  & $0.8403\pm0.0109$  & $0.6561\pm0.0089$  & $0.6835\pm0.0244$  & $0.5310\pm0.0165$  \\
GSViT                   & EfficientViT              & $0.4053\pm0.0103$  & $0.6451\pm0.0106$  & $0.4612\pm0.0256$  & $0.4296\pm0.0210$  & $0.3005\pm0.0121$  \\
EndoViT                 & ViT-B                     & $0.5959\pm0.0150$  & $0.8185\pm0.0133$  & $0.6160\pm0.0202$  & $0.6481\pm0.0145$  & $0.4940\pm0.0168$  \\
PeskaVLP                & ResNet50                  & $0.6648\pm0.0132$  & $0.8577\pm0.0061$  & $0.6758\pm0.0084$  & $0.7020\pm0.0194$  & $0.5615\pm0.0108$  \\
SurgeNet                & CAFormer18                & $0.7082\pm0.0115$  & $0.8812\pm0.0028$  & $0.7175\pm0.0094$  & $0.7619\pm0.0128$  & $0.6096\pm0.0069$  \\
ZEN                     & ViT-B                     & $\mathbf{0.7268\pm0.0096}$ & $\mathbf{0.8845\pm0.0049}$ & $\mathbf{0.7260\pm0.0116}$ & $\mathbf{0.7697\pm0.0148}$ & $\mathbf{0.6212\pm0.0071}$ \\ \bottomrule
\end{tabular}
\end{table}

\begin{table}[t]
\centering
\caption{\textbf{Surgical phase recognition performance comparison on the AutoLaparo dataset using frozen backbones.} The upper block reports results for self-supervised learning pretrained models, and the lower block reports results for existing pretrained models. Best-performing models for each metric are highlighted in bold. Values are reported as mean $\pm$ standard deviation over five independent runs.}
\label{tab:autolaparo-freeze}
\setlength{\tabcolsep}{1.5pt} 
\begin{tabular}{lcccccc}
\toprule
\multirow{2}{*}{Method} & \multirow{2}{*}{Backbone} & \multicolumn{2}{c}{Video-level} & \multicolumn{3}{c}{Phase-level} \\ \cmidrule(lr){3-4} \cmidrule(lr){5-7}
                        &                           & F1 score           & Accuracy           & Precision          & Recall             & Jaccard            \\ \midrule
MIS-MAE                 & ViT-B                     & $0.6483\pm0.0305$  & $0.7452\pm0.0220$  & $0.7280\pm0.0358$  & $0.6572\pm0.0398$  & $0.5290\pm0.0387$  \\
MIS-MoCoV3              & ViT-B                     & $0.6755\pm0.0233$  & $0.7928\pm0.0224$  & $0.7167\pm0.0346$  & $0.6973\pm0.0204$  & $0.5752\pm0.0195$  \\
MIS-MSN                 & ViT-B                     & $0.5196\pm0.0480$  & $0.6716\pm0.0374$  & $0.6965\pm0.0454$  & $0.5241\pm0.0406$  & $0.3934\pm0.0405$  \\
MIS-DINO                & ViT-B                     & $0.7023\pm0.0431$  & $0.8088\pm0.0430$  & $0.7534\pm0.0244$  & $0.7328\pm0.0440$  & $0.6158\pm0.0453$  \\
MIS-DINOv2              & ViT-B                     & $0.7148\pm0.0511$  & $0.8230\pm0.0173$  & $\mathbf{0.7863\pm0.0280}$ & $0.7275\pm0.0608$  & $0.6157\pm0.0547$  \\
MIS-DINOv2              & ViT-L                     & $\mathbf{0.7410\pm0.0179}$ & $\mathbf{0.8382\pm0.0143}$ & $0.7845\pm0.0361$  & $\mathbf{0.7429\pm0.0176}$ & $\mathbf{0.6376\pm0.0205}$ \\ \midrule
ImageNet                & ViT-B                     & $0.6208\pm0.0382$  & $0.7273\pm0.0408$  & $0.6747\pm0.0528$  & $0.6402\pm0.0304$  & $0.5075\pm0.0391$  \\
EndoFM                  & ViT-B                     & $0.5876\pm0.0325$  & $0.7167\pm0.0182$  & $0.6501\pm0.0166$  & $0.6040\pm0.0385$  & $0.4710\pm0.0350$  \\
GSViT                   & EfficientViT              & $0.4503\pm0.0440$  & $0.6155\pm0.0319$  & $0.6636\pm0.0437$  & $0.4639\pm0.0333$  & $0.3306\pm0.0318$  \\
EndoViT                 & ViT-B                     & $0.5875\pm0.0761$  & $0.7307\pm0.0335$  & $0.7043\pm0.0753$  & $0.5991\pm0.0649$  & $0.4668\pm0.0615$  \\
PeskaVLP                & ResNet50                  & $0.6792\pm0.0405$  & $0.7991\pm0.0164$  & $0.7293\pm0.0463$  & $0.6844\pm0.0299$  & $0.5758\pm0.0309$  \\
SurgeNet                & CAFormer18                & $0.7160\pm0.0271$  & $0.8124\pm0.0123$  & $0.7734\pm0.0525$  & $0.7169\pm0.0250$  & $0.6008\pm0.0186$  \\
ZEN                     & ViT-B                     & $\mathbf{0.7509\pm0.0096}$ & $\mathbf{0.8542\pm0.0121}$ & $\mathbf{0.7853\pm0.0182}$ & $\mathbf{0.7626\pm0.0191}$ & $\mathbf{0.6593\pm0.0191}$ \\ \bottomrule
\end{tabular}
\end{table}

\begin{table}[t]
\centering
\caption{\textbf{Few-shot evaluation on the Cholec80 dataset for surgical phase recognition (1-shot).} The upper block reports results for self-supervised learning pretrained models, and the lower block reports results for existing pretrained models. Best-performing models for each metric are highlighted in bold. Values are reported as mean $\pm$ standard deviation over five independent runs.}
\label{tab:cholec80-fewshot-1}
\setlength{\tabcolsep}{1.5pt} 
\begin{tabular}{lcccccc}
\toprule
\multirow{2}{*}{Method} & \multirow{2}{*}{Backbone} & \multicolumn{2}{c}{Video-level} & \multicolumn{3}{c}{Phase-level} \\ \cmidrule(lr){3-4} \cmidrule(lr){5-7}
                        &                           & F1 score           & Accuracy           & Precision          & Recall             & Jaccard            \\ \midrule
MIS-MAE                 & ViT-B                     & $0.2087\pm0.0476$  & $0.5363\pm0.0852$  & $0.4899\pm0.1419$  & $0.2625\pm0.0379$  & $0.1605\pm0.0384$  \\
MIS-MoCoV3              & ViT-B                     & $0.2096\pm0.0461$  & $0.5233\pm0.0560$  & $0.4983\pm0.0778$  & $0.2613\pm0.0402$  & $0.1596\pm0.0361$  \\
MIS-MSN                 & ViT-B                     & $0.2055\pm0.0450$  & $0.5008\pm0.0755$  & $0.4989\pm0.1138$  & $0.2571\pm0.0423$  & $0.1554\pm0.0364$  \\
MIS-DINO                & ViT-B                     & $\mathbf{0.3168\pm0.1103}$ & $\mathbf{0.5698\pm0.0540}$ & $0.4477\pm0.0751$  & $\mathbf{0.3911\pm0.1110}$ & $\mathbf{0.2452\pm0.0927}$ \\
MIS-DINOv2              & ViT-B                     & $0.2307\pm0.0768$  & $0.5397\pm0.0740$  & $\mathbf{0.5256\pm0.0863}$ & $0.2885\pm0.0733$  & $0.1776\pm0.0627$  \\
MIS-DINOv2              & ViT-L                     & $0.2400\pm0.0456$  & $0.5553\pm0.0914$  & $0.4380\pm0.0985$  & $0.3065\pm0.0487$  & $0.1868\pm0.0414$  \\ \midrule
ImageNet                & ViT-B                     & $0.1871\pm0.0268$  & $0.5301\pm0.0525$  & $\mathbf{0.5124\pm0.1209}$ & $0.2416\pm0.0403$  & $0.1404\pm0.0183$  \\
EndoFM                  & ViT-B                     & $0.2227\pm0.0491$  & $0.5215\pm0.0562$  & $0.4467\pm0.0976$  & $0.2812\pm0.0446$  & $0.1704\pm0.0396$  \\
GSViT                   & EfficientViT              & $0.2139\pm0.0417$  & $0.5198\pm0.0622$  & $0.4913\pm0.1140$  & $0.2639\pm0.0387$  & $0.1625\pm0.0329$  \\
EndoViT                 & ViT-B                     & $0.2082\pm0.0543$  & $0.5162\pm0.0650$  & $0.4858\pm0.0975$  & $0.2683\pm0.0546$  & $0.1589\pm0.0440$  \\
PeskaVLP                & ResNet50                  & $0.2098\pm0.0604$  & $0.5147\pm0.0584$  & $0.4604\pm0.1246$  & $0.2681\pm0.0723$  & $0.1592\pm0.0471$  \\
SurgeNet                & CAFormer18                & $0.2193\pm0.0230$  & $0.5551\pm0.0460$  & $0.4690\pm0.0271$  & $0.2719\pm0.0387$  & $0.1698\pm0.0162$  \\
ZEN                     & ViT-B                     & $\mathbf{0.2269\pm0.0341}$ & $\mathbf{0.5586\pm0.0936}$ & $0.4672\pm0.0935$  & $\mathbf{0.2895\pm0.0446}$ & $\mathbf{0.1759\pm0.0328}$ \\ \bottomrule
\end{tabular}
\end{table}

\begin{table}[t]
\centering
\caption{\textbf{Few-shot evaluation on the Cholec80 dataset for surgical phase recognition (2-shot).} The upper block reports results for self-supervised learning pretrained models, and the lower block reports results for existing pretrained models. Best-performing models for each metric are highlighted in bold. Values are reported as mean $\pm$ standard deviation over five independent runs.}
\label{tab:cholec80-fewshot-2}
\setlength{\tabcolsep}{1.5pt} 
\begin{tabular}{lcccccc}
\toprule
\multirow{2}{*}{Method} & \multirow{2}{*}{Backbone} & \multicolumn{2}{c}{Video-level} & \multicolumn{3}{c}{Phase-level} \\ \cmidrule(lr){3-4} \cmidrule(lr){5-7}
                        &                           & F1 score           & Accuracy           & Precision          & Recall             & Jaccard            \\ \midrule
MIS-MAE                 & ViT-B                     & $0.3094\pm0.0914$  & $0.5685\pm0.0424$  & $0.4877\pm0.0882$  & $0.3738\pm0.0953$  & $0.2417\pm0.0765$  \\
MIS-MoCoV3              & ViT-B                     & $0.3540\pm0.0802$  & $0.5675\pm0.0545$  & $0.4608\pm0.0949$  & $0.4150\pm0.0871$  & $0.2749\pm0.0690$  \\
MIS-MSN                 & ViT-B                     & $0.2490\pm0.1093$  & $0.5170\pm0.0918$  & $0.4707\pm0.1148$  & $0.3114\pm0.1153$  & $0.1919\pm0.0897$  \\
MIS-DINO                & ViT-B                     & $\mathbf{0.4243\pm0.0512}$ & $0.6456\pm0.0361$  & $0.5152\pm0.0389$  & $\mathbf{0.4993\pm0.0593}$ & $0.3346\pm0.0394$  \\
MIS-DINOv2              & ViT-B                     & $0.3544\pm0.0476$  & $0.6230\pm0.0203$  & $0.4945\pm0.0543$  & $0.4209\pm0.0481$  & $0.2782\pm0.0404$  \\
MIS-DINOv2              & ViT-L                     & $0.4190\pm0.0989$  & $\mathbf{0.6579\pm0.0556}$ & $\mathbf{0.5207\pm0.0781}$ & $0.4975\pm0.0964$  & $\mathbf{0.3362\pm0.0869}$ \\ \midrule
ImageNet                & ViT-B                     & $0.2765\pm0.0649$  & $0.5588\pm0.0314$  & $0.4565\pm0.0667$  & $0.3300\pm0.0645$  & $0.2125\pm0.0532$  \\
EndoFM                  & ViT-B                     & $0.3135\pm0.0707$  & $0.5518\pm0.0609$  & $0.4167\pm0.1123$  & $0.3684\pm0.0632$  & $0.2396\pm0.0631$  \\
GSViT                   & EfficientViT              & $0.2097\pm0.0824$  & $0.5046\pm0.0852$  & $0.5044\pm0.0626$  & $0.2695\pm0.0806$  & $0.1607\pm0.0667$  \\
EndoViT                 & ViT-B                     & $0.2818\pm0.0916$  & $0.5274\pm0.0806$  & $0.4324\pm0.1339$  & $0.3449\pm0.0961$  & $0.2193\pm0.0754$  \\
PeskaVLP                & ResNet50                  & $0.2991\pm0.1185$  & $0.5521\pm0.0754$  & $0.4430\pm0.1412$  & $0.3630\pm0.1303$  & $0.2322\pm0.0981$  \\
SurgeNet                & CAFormer18                & $0.3508\pm0.1344$  & $\mathbf{0.6216\pm0.0224}$ & $\mathbf{0.5849\pm0.0708}$ & $0.4144\pm0.1523$  & $0.2743\pm0.1094$  \\
ZEN                     & ViT-B                     & $\mathbf{0.3635\pm0.1223}$ & $0.6164\pm0.0505$  & $0.4796\pm0.0846$  & $\mathbf{0.4399\pm0.1321}$ & $\mathbf{0.2881\pm0.1038}$ \\ \bottomrule
\end{tabular}
\end{table}

\begin{table}[t]
\centering
\caption{\textbf{Few-shot evaluation on the Cholec80 dataset for surgical phase recognition (3-shot).} The upper block reports results for self-supervised learning pretrained models, and the lower block reports results for existing pretrained models. Best-performing models for each metric are highlighted in bold. Values are reported as mean $\pm$ standard deviation over five independent runs.}
\label{tab:cholec80-fewshot-3}
\setlength{\tabcolsep}{1.5pt} 
\begin{tabular}{lcccccc}
\toprule
\multirow{2}{*}{Method} & \multirow{2}{*}{Backbone} & \multicolumn{2}{c}{Video-level} & \multicolumn{3}{c}{Phase-level} \\ \cmidrule(lr){3-4} \cmidrule(lr){5-7}
                        &                           & F1 score           & Accuracy           & Precision          & Recall             & Jaccard            \\ \midrule
MIS-MAE                 & ViT-B                     & $0.3337\pm0.0880$  & $0.6059\pm0.0450$  & $0.4978\pm0.0552$  & $0.3890\pm0.0976$  & $0.2643\pm0.0770$  \\
MIS-MoCoV3              & ViT-B                     & $0.3977\pm0.0351$  & $0.6040\pm0.0486$  & $0.5248\pm0.0529$  & $0.4382\pm0.0375$  & $0.3138\pm0.0318$  \\
MIS-MSN                 & ViT-B                     & $0.3070\pm0.1053$  & $0.5483\pm0.0804$  & $0.4590\pm0.1165$  & $0.3712\pm0.1132$  & $0.2384\pm0.0869$  \\
MIS-DINO                & ViT-B                     & $0.4742\pm0.0684$  & $0.6862\pm0.0124$  & $0.5390\pm0.0500$  & $0.5515\pm0.0828$  & $0.3852\pm0.0571$  \\
MIS-DINOv2              & ViT-B                     & $0.4586\pm0.0601$  & $0.6737\pm0.0231$  & $0.5753\pm0.0512$  & $0.5187\pm0.0585$  & $0.3687\pm0.0517$  \\
MIS-DINOv2              & ViT-L                     & $\mathbf{0.5220\pm0.0326}$ & $\mathbf{0.7331\pm0.0152}$ & $\mathbf{0.6417\pm0.0225}$ & $\mathbf{0.5834\pm0.0439}$ & $\mathbf{0.4298\pm0.0240}$ \\ \midrule
ImageNet                & ViT-B                     & $0.3647\pm0.1054$  & $0.6102\pm0.0334$  & $0.5365\pm0.0368$  & $0.4151\pm0.1081$  & $0.2859\pm0.0840$  \\
EndoFM                  & ViT-B                     & $0.3805\pm0.0518$  & $0.5968\pm0.0387$  & $0.4982\pm0.0505$  & $0.4348\pm0.0451$  & $0.2953\pm0.0506$  \\
GSViT                   & EfficientViT              & $0.2083\pm0.0779$  & $0.5566\pm0.0803$  & $0.5619\pm0.0846$  & $0.2568\pm0.0751$  & $0.1617\pm0.0647$  \\
EndoViT                 & ViT-B                     & $0.3299\pm0.0779$  & $0.5770\pm0.0429$  & $0.4757\pm0.0647$  & $0.3898\pm0.0845$  & $0.2588\pm0.0633$  \\
PeskaVLP                & ResNet50                  & $0.4159\pm0.0690$  & $0.6277\pm0.0464$  & $0.4908\pm0.0738$  & $0.4774\pm0.0763$  & $0.3343\pm0.0632$  \\
SurgeNet                & CAFormer18                & $0.4114\pm0.0980$  & $0.6544\pm0.0407$  & $0.5406\pm0.0487$  & $0.4767\pm0.1057$  & $0.3333\pm0.0831$  \\
ZEN                     & ViT-B                     & $\mathbf{0.5338\pm0.0387}$ & $\mathbf{0.7212\pm0.0477}$ & $\mathbf{0.5932\pm0.0661}$ & $\mathbf{0.6084\pm0.0246}$ & $\mathbf{0.4375\pm0.0394}$ \\ \bottomrule
\end{tabular}
\end{table}

\begin{table}[t]
\centering
\caption{\textbf{Few-shot evaluation on the Cholec80 dataset for surgical phase recognition (4-shot).} The upper block reports results for self-supervised learning pretrained models, and the lower block reports results for existing pretrained models. Best-performing models for each metric are highlighted in bold. Values are reported as mean $\pm$ standard deviation over five independent runs.}
\label{tab:cholec80-fewshot-4}
\setlength{\tabcolsep}{1.5pt}
\begin{tabular}{lcccccc}
\toprule
\multirow{2}{*}{Method} & \multirow{2}{*}{Backbone} & \multicolumn{2}{c}{Video-level} & \multicolumn{3}{c}{Phase-level} \\ \cmidrule(lr){3-4} \cmidrule(lr){5-7}
                        &                           & F1 score           & Accuracy           & Precision          & Recall             & Jaccard            \\ \midrule
MIS-MAE                 & ViT-B                     & $0.4290\pm0.0642$  & $0.6521\pm0.0276$  & $0.5828\pm0.0524$  & $0.4879\pm0.0662$  & $0.3458\pm0.0548$  \\
MIS-MoCoV3              & ViT-B                     & $0.4389\pm0.0348$  & $0.6461\pm0.0164$  & $0.5613\pm0.0173$  & $0.4875\pm0.0305$  & $0.3536\pm0.0295$  \\
MIS-MSN                 & ViT-B                     & $0.3747\pm0.0743$  & $0.6049\pm0.0578$  & $0.5664\pm0.0755$  & $0.4347\pm0.0565$  & $0.3003\pm0.0661$  \\
MIS-DINO                & ViT-B                     & $0.5086\pm0.0510$  & $0.7105\pm0.0134$  & $0.5687\pm0.0519$  & $0.5782\pm0.0621$  & $0.4139\pm0.0395$  \\
MIS-DINOv2              & ViT-B                     & $0.4570\pm0.0673$  & $0.6991\pm0.0143$  & $0.5912\pm0.0286$  & $0.5139\pm0.0736$  & $0.3775\pm0.0522$  \\
MIS-DINOv2              & ViT-L                     & $\mathbf{0.6044\pm0.0357}$ & $\mathbf{0.7710\pm0.0149}$ & $\mathbf{0.6715\pm0.0290}$ & $\mathbf{0.6787\pm0.0542}$ & $\mathbf{0.5072\pm0.0345}$ \\ \midrule
ImageNet                & ViT-B                     & $0.4033\pm0.0261$  & $0.6426\pm0.0110$  & $0.5252\pm0.0450$  & $0.4496\pm0.0274$  & $0.3173\pm0.0183$  \\
EndoFM                  & ViT-B                     & $0.3393\pm0.0811$  & $0.6264\pm0.0245$  & $0.5632\pm0.0209$  & $0.3892\pm0.0825$  & $0.2702\pm0.0664$  \\
GSViT                   & EfficientViT              & $0.2439\pm0.0463$  & $0.5940\pm0.0063$  & $0.6138\pm0.0537$  & $0.2857\pm0.0478$  & $0.1897\pm0.0350$  \\
EndoViT                 & ViT-B                     & $0.4129\pm0.0368$  & $0.6134\pm0.0345$  & $0.5020\pm0.0436$  & $0.4703\pm0.0245$  & $0.3300\pm0.0378$  \\
PeskaVLP                & ResNet50                  & $0.4682\pm0.0577$  & $0.6567\pm0.0211$  & $0.5680\pm0.0670$  & $0.5204\pm0.0590$  & $0.3800\pm0.0505$  \\
SurgeNet                & CAFormer18                & $0.5474\pm0.0317$  & $0.7030\pm0.0128$  & $0.6140\pm0.0149$  & $\mathbf{0.6229\pm0.0555}$ & $0.4471\pm0.0285$  \\
ZEN                     & ViT-B                     & $\mathbf{0.5489\pm0.0834}$ & $\mathbf{0.7483\pm0.0467}$ & $\mathbf{0.6199\pm0.0890}$ & $0.6170\pm0.0734$  & $\mathbf{0.4593\pm0.0721}$ \\ \bottomrule
\end{tabular}
\end{table}

\begin{table}[t]
\centering
\caption{\textbf{Few-shot evaluation on the Cholec80 dataset for surgical phase recognition (5-shot).} The upper block reports results for self-supervised learning pretrained models, and the lower block reports results for existing pretrained models. Best-performing models for each metric are highlighted in bold. Values are reported as mean $\pm$ standard deviation over five independent runs.}
\label{tab:cholec80-fewshot-5}
\setlength{\tabcolsep}{1.5pt} 
\begin{tabular}{lcccccc}
\toprule
\multirow{2}{*}{Method} & \multirow{2}{*}{Backbone} & \multicolumn{2}{c}{Video-level} & \multicolumn{3}{c}{Phase-level} \\ \cmidrule(lr){3-4} \cmidrule(lr){5-7}
                        &                           & F1 score           & Accuracy           & Precision          & Recall             & Jaccard            \\ \midrule
MIS-MAE                 & ViT-B                     & $0.4564\pm0.0624$  & $0.6589\pm0.0415$  & $0.5355\pm0.0613$  & $0.5170\pm0.0631$  & $0.3726\pm0.0564$  \\
MIS-MoCoV3              & ViT-B                     & $0.4671\pm0.0472$  & $0.6684\pm0.0118$  & $0.5550\pm0.0490$  & $0.5178\pm0.0518$  & $0.3831\pm0.0402$  \\
MIS-MSN                 & ViT-B                     & $0.3900\pm0.1066$  & $0.6216\pm0.0360$  & $0.5560\pm0.0746$  & $0.4439\pm0.1082$  & $0.3099\pm0.0881$  \\
MIS-SimDINO             & ViT-B                     & $0.5645\pm0.0730$  & $0.7463\pm0.0202$  & $0.6224\pm0.0494$  & $0.6280\pm0.0824$  & $0.4705\pm0.0645$  \\
MIS-DINOv2              & ViT-B                     & $0.5416\pm0.0095$  & $0.7277\pm0.0242$  & $0.6129\pm0.0228$  & $0.6028\pm0.0136$  & $0.4494\pm0.0121$  \\
MIS-DINOv2              & ViT-L                     & $\mathbf{0.6055\pm0.0649}$ & $\mathbf{0.7847\pm0.0313}$ & $\mathbf{0.6540\pm0.0329}$ & $\mathbf{0.6778\pm0.0833}$ & $\mathbf{0.5110\pm0.0559}$ \\ \midrule
ImageNet                & ViT-B                     & $0.4153\pm0.0711$  & $0.6398\pm0.0230$  & $0.5544\pm0.0543$  & $0.4618\pm0.0640$  & $0.3320\pm0.0579$  \\
EndoFM                  & ViT-B                     & $0.3748\pm0.0997$  & $0.6323\pm0.0269$  & $0.5438\pm0.0393$  & $0.4181\pm0.0999$  & $0.2982\pm0.0796$  \\
GSViT                   & EfficientViT              & $0.2637\pm0.0397$  & $0.5941\pm0.0150$  & $0.6092\pm0.0995$  & $0.3038\pm0.0403$  & $0.2051\pm0.0314$  \\
EndoViT                 & ViT-B                     & $0.3482\pm0.0891$  & $0.6394\pm0.0260$  & $0.5828\pm0.0796$  & $0.3951\pm0.0920$  & $0.2827\pm0.0778$  \\
PeskaVLP                & ResNet50                  & $0.4626\pm0.1362$  & $0.6776\pm0.0419$  & $0.6040\pm0.0170$  & $0.5198\pm0.1476$  & $0.3817\pm0.1161$  \\
SurgeNet                & CAFormer18                & $0.5626\pm0.0242$  & $0.7234\pm0.0213$  & $0.6302\pm0.0283$  & $0.6306\pm0.0192$  & $0.4613\pm0.0232$  \\
ZEN                     & ViT-B                     & $\mathbf{0.5837\pm0.0502}$ & $\mathbf{0.7523\pm0.0200}$ & $\mathbf{0.6320\pm0.0182}$ & $\mathbf{0.6628\pm0.0691}$ & $\mathbf{0.4858\pm0.0423}$ \\ \bottomrule
\end{tabular}
\end{table}

\begin{table}[t]
\centering
\caption{\textbf{Few-shot evaluation on the MultiByPass140 dataset for surgical phase recognition (1-shot).} The upper block reports results for self-supervised learning pretrained models, and the lower block reports results for existing pretrained models. Best-performing models for each metric are highlighted in bold. Values are reported as mean $\pm$ standard deviation over five independent runs.}
\label{tab:mbp-fewshot-1}
\setlength{\tabcolsep}{1.5pt} 
\begin{tabular}{lcccccc}
\toprule
\multirow{2}{*}{Method} & \multirow{2}{*}{Backbone} & \multicolumn{2}{c}{Video-level} & \multicolumn{3}{c}{Phase-level} \\ \cmidrule(lr){3-4} \cmidrule(lr){5-7}
                        &                           & F1 score           & Accuracy           & Precision          & Recall             & Jaccard            \\ \midrule
MIS-MAE                 & ViT-B                     & $0.1669\pm0.0211$  & $0.4056\pm0.0338$  & $0.3892\pm0.0571$  & $0.1777\pm0.0272$  & $0.0957\pm0.0215$  \\
MIS-MoCoV3              & ViT-B                     & $0.1889\pm0.0208$  & $0.4238\pm0.0191$  & $0.4308\pm0.0831$  & $0.1935\pm0.0242$  & $0.1091\pm0.0180$  \\
MIS-MSN                 & ViT-B                     & $0.1459\pm0.0369$  & $0.3708\pm0.0314$  & $0.3701\pm0.0780$  & $0.1636\pm0.0387$  & $0.0852\pm0.0290$  \\
MIS-DINO                & ViT-B                     & $0.2240\pm0.0424$  & $0.5061\pm0.0355$  & $0.4760\pm0.0756$  & $0.2259\pm0.0435$  & $0.1353\pm0.0293$  \\
MIS-DINOv2              & ViT-B                     & $0.2199\pm0.0423$  & $0.5018\pm0.0396$  & $\mathbf{0.5138\pm0.0962}$ & $0.2112\pm0.0391$  & $0.1299\pm0.0299$  \\
MIS-DINOv2              & ViT-L                     & $\mathbf{0.2282\pm0.0325}$ & $\mathbf{0.5358\pm0.0326}$ & $0.4221\pm0.0970$  & $\mathbf{0.2304\pm0.0252}$ & $\mathbf{0.1432\pm0.0170}$ \\ \midrule
ImageNet                & ViT-B                     & $0.1619\pm0.0089$  & $0.4081\pm0.0442$  & $0.4188\pm0.0996$  & $0.1714\pm0.0195$  & $0.0910\pm0.0068$  \\
EndoFM                  & ViT-B                     & $0.1841\pm0.0178$  & $0.4320\pm0.0320$  & $0.4324\pm0.0761$  & $0.1851\pm0.0296$  & $0.1029\pm0.0141$  \\
GSViT                   & EfficientViT              & $0.1401\pm0.0372$  & $0.3600\pm0.0417$  & $0.3859\pm0.0688$  & $0.1595\pm0.0359$  & $0.0818\pm0.0290$  \\
EndoViT                 & ViT-B                     & $0.1442\pm0.0371$  & $0.3592\pm0.0369$  & $0.3754\pm0.0738$  & $0.1607\pm0.0376$  & $0.0822\pm0.0279$  \\
PeskaVLP                & ResNet50                  & $0.1952\pm0.0483$  & $0.4526\pm0.0454$  & $0.3761\pm0.0833$  & $0.2084\pm0.0611$  & $0.1207\pm0.0408$  \\
SurgeNet                & CAFormer18                & $0.2252\pm0.0360$  & $0.5064\pm0.0178$  & $0.4417\pm0.0790$  & $0.2214\pm0.0371$  & $0.1350\pm0.0262$  \\
ZEN                     & ViT-B                     & $\mathbf{0.2359\pm0.0470}$ & $\mathbf{0.5348\pm0.0433}$ & $\mathbf{0.5018\pm0.0624}$ & $\mathbf{0.2307\pm0.0529}$ & $\mathbf{0.1440\pm0.0326}$ \\ \bottomrule
\end{tabular}
\end{table}

\begin{table}[t]
\centering
\caption{\textbf{Few-shot evaluation on the MultiByPass140 dataset for surgical phase recognition (2-shot).} The upper block reports results for self-supervised learning pretrained models, and the lower block reports results for existing pretrained models. Best-performing models for each metric are highlighted in bold. Values are reported as mean $\pm$ standard deviation over five independent runs.}
\label{tab:mbp-fewshot-2}
\setlength{\tabcolsep}{1.5pt} 
\begin{tabular}{lcccccc}
\toprule
\multirow{2}{*}{Method} & \multirow{2}{*}{Backbone} & \multicolumn{2}{c}{Video-level} & \multicolumn{3}{c}{Phase-level} \\ \cmidrule(lr){3-4} \cmidrule(lr){5-7}
                        &                           & F1 score           & Accuracy           & Precision          & Recall             & Jaccard            \\ \midrule
MIS-MAE                 & ViT-B                     & $0.2592\pm0.0266$  & $0.5032\pm0.0517$  & $0.4363\pm0.0813$  & $0.2501\pm0.0341$  & $0.1583\pm0.0214$  \\
MIS-MoCoV3              & ViT-B                     & $0.2616\pm0.0426$  & $0.4893\pm0.0721$  & $0.4597\pm0.1196$  & $0.2589\pm0.0528$  & $0.1643\pm0.0347$  \\
MIS-MSN                 & ViT-B                     & $0.2429\pm0.0297$  & $0.4602\pm0.0333$  & $0.4147\pm0.1343$  & $0.2362\pm0.0154$  & $0.1478\pm0.0206$  \\
MIS-DINO                & ViT-B                     & $\mathbf{0.3515\pm0.0747}$ & $0.6439\pm0.0616$  & $\mathbf{0.4614\pm0.0595}$ & $\mathbf{0.3499\pm0.0818}$ & $\mathbf{0.2458\pm0.0658}$ \\
MIS-DINOv2              & ViT-B                     & $0.3241\pm0.0309$  & $0.6032\pm0.0571$  & $0.3953\pm0.0405$  & $0.3140\pm0.0336$  & $0.2129\pm0.0235$  \\
MIS-DINOv2              & ViT-L                     & $0.3361\pm0.0704$  & $\mathbf{0.6456\pm0.0378}$ & $0.4384\pm0.0750$  & $0.3215\pm0.0742$  & $0.2261\pm0.0626$  \\ \midrule
ImageNet                & ViT-B                     & $0.2303\pm0.0455$  & $0.4964\pm0.0371$  & $\mathbf{0.4786\pm0.1190}$ & $0.2204\pm0.0501$  & $0.1364\pm0.0336$  \\
EndoFM                  & ViT-B                     & $0.2250\pm0.0256$  & $0.4854\pm0.0287$  & $0.4425\pm0.0934$  & $0.2369\pm0.0232$  & $0.1394\pm0.0136$  \\
GSViT                   & EfficientViT              & $0.2031\pm0.0428$  & $0.4242\pm0.0449$  & $0.4512\pm0.0795$  & $0.2029\pm0.0358$  & $0.1210\pm0.0306$  \\
EndoViT                 & ViT-B                     & $0.2067\pm0.0542$  & $0.4382\pm0.0551$  & $0.3969\pm0.0787$  & $0.2151\pm0.0424$  & $0.1266\pm0.0362$  \\
PeskaVLP                & ResNet50                  & $0.2759\pm0.0357$  & $0.5609\pm0.0409$  & $0.4183\pm0.0637$  & $0.2696\pm0.0445$  & $0.1719\pm0.0279$  \\
SurgeNet                & CAFormer18                & $0.3217\pm0.0733$  & $0.6053\pm0.0477$  & $0.4505\pm0.0467$  & $0.3058\pm0.0678$  & $0.2053\pm0.0554$  \\
ZEN                     & ViT-B                     & $\mathbf{0.3574\pm0.0484}$ & $\mathbf{0.6320\pm0.0647}$ & $0.4622\pm0.0859$  & $\mathbf{0.3395\pm0.0585}$ & $\mathbf{0.2390\pm0.0479}$ \\ \bottomrule
\end{tabular}
\end{table}

\begin{table}[t]
\centering
\caption{\textbf{Few-shot evaluation on the MultiByPass140 dataset for surgical phase recognition (3-shot).} The upper block reports results for self-supervised learning pretrained models, and the lower block reports results for existing pretrained models. Best-performing models for each metric are highlighted in bold. Values are reported as mean $\pm$ standard deviation over five independent runs.}
\label{tab:mbp-fewshot-3}
\setlength{\tabcolsep}{1.5pt} 
\begin{tabular}{lcccccc}
\toprule
\multirow{2}{*}{Method} & \multirow{2}{*}{Backbone} & \multicolumn{2}{c}{Video-level} & \multicolumn{3}{c}{Phase-level} \\ \cmidrule(lr){3-4} \cmidrule(lr){5-7}
                        &                           & F1 score           & Accuracy           & Precision          & Recall             & Jaccard            \\ \midrule
MIS-MAE                 & ViT-B                     & $0.2930\pm0.0238$  & $0.5341\pm0.0427$  & $0.4250\pm0.0779$  & $0.2770\pm0.0232$  & $0.1835\pm0.0236$  \\
MIS-MoCoV3              & ViT-B                     & $0.2819\pm0.0341$  & $0.5223\pm0.0564$  & $0.4136\pm0.0487$  & $0.2697\pm0.0322$  & $0.1789\pm0.0292$  \\
MIS-MSN                 & ViT-B                     & $0.2668\pm0.0369$  & $0.4950\pm0.0412$  & $0.4447\pm0.0918$  & $0.2550\pm0.0276$  & $0.1664\pm0.0274$  \\
MIS-DINO                & ViT-B                     & $\mathbf{0.3742\pm0.0330}$ & $0.6530\pm0.0537$  & $\mathbf{0.5093\pm0.1004}$ & $\mathbf{0.3577\pm0.0546}$ & $\mathbf{0.2548\pm0.0387}$ \\
MIS-DINOv2              & ViT-B                     & $0.3175\pm0.0192$  & $0.6202\pm0.0560$  & $0.4560\pm0.0881$  & $0.3182\pm0.0129$  & $0.2179\pm0.0100$  \\
MIS-DINOv2              & ViT-L                     & $0.3635\pm0.0509$  & $\mathbf{0.6617\pm0.0411}$ & $0.5072\pm0.0512$  & $0.3467\pm0.0513$  & $0.2438\pm0.0494$  \\ \midrule
ImageNet                & ViT-B                     & $0.2693\pm0.0154$  & $0.5222\pm0.0277$  & $\mathbf{0.5353\pm0.1212}$ & $0.2486\pm0.0139$  & $0.1647\pm0.0171$  \\
EndoFM                  & ViT-B                     & $0.2596\pm0.0137$  & $0.5024\pm0.0321$  & $0.4575\pm0.0870$  & $0.2469\pm0.0148$  & $0.1625\pm0.0107$  \\
GSViT                   & EfficientViT              & $0.2191\pm0.0204$  & $0.4335\pm0.0362$  & $0.4833\pm0.0661$  & $0.2167\pm0.0154$  & $0.1333\pm0.0192$  \\
EndoViT                 & ViT-B                     & $0.2222\pm0.0558$  & $0.4565\pm0.0640$  & $0.4340\pm0.1051$  & $0.2200\pm0.0443$  & $0.1359\pm0.0372$  \\
PeskaVLP                & ResNet50                  & $0.2858\pm0.0110$  & $0.5489\pm0.0627$  & $0.4682\pm0.1204$  & $0.2868\pm0.0201$  & $0.1923\pm0.0097$  \\
SurgeNet                & CAFormer18                & $0.3330\pm0.0393$  & $0.6285\pm0.0476$  & $0.4837\pm0.1330$  & $0.3251\pm0.0412$  & $0.2261\pm0.0385$  \\
ZEN                     & ViT-B                     & $\mathbf{0.3560\pm0.0432}$ & $\mathbf{0.6442\pm0.0585}$ & $0.4653\pm0.0824$  & $\mathbf{0.3595\pm0.0625}$ & $\mathbf{0.2538\pm0.0493}$ \\ \bottomrule
\end{tabular}
\end{table}

\begin{table}[t]
\centering
\caption{\textbf{Few-shot evaluation on the MultiByPass140 dataset for surgical phase recognition (4-shot).} The upper block reports results for self-supervised learning pretrained models, and the lower block reports results for existing pretrained models. Best-performing models for each metric are highlighted in bold. Values are reported as mean $\pm$ standard deviation over five independent runs.}
\label{tab:mbp-fewshot-4}
\setlength{\tabcolsep}{1.5pt} 
\begin{tabular}{lcccccc}
\toprule
\multirow{2}{*}{Method} & \multirow{2}{*}{Backbone} & \multicolumn{2}{c}{Video-level} & \multicolumn{3}{c}{Phase-level} \\ \cmidrule(lr){3-4} \cmidrule(lr){5-7}
                        &                           & F1 score           & Accuracy           & Precision          & Recall             & Jaccard            \\ \midrule
MIS-MAE                 & ViT-B                     & $0.3247\pm0.0337$  & $0.5797\pm0.0399$  & $0.4512\pm0.1184$  & $0.3129\pm0.0304$  & $0.2127\pm0.0252$  \\
MIS-MoCoV3              & ViT-B                     & $0.2867\pm0.0226$  & $0.5691\pm0.0314$  & $\mathbf{0.5146\pm0.1060}$ & $0.2775\pm0.0403$  & $0.1831\pm0.0319$  \\
MIS-MSN                 & ViT-B                     & $0.2949\pm0.0391$  & $0.5274\pm0.0317$  & $0.4106\pm0.0501$  & $0.2826\pm0.0273$  & $0.1856\pm0.0302$  \\
MIS-DINO                & ViT-B                     & $0.3762\pm0.0346$  & $0.6685\pm0.0197$  & $0.4945\pm0.0982$  & $\mathbf{0.3636\pm0.0464}$ & $0.2584\pm0.0352$  \\
MIS-DINOv2              & ViT-B                     & $0.3472\pm0.0331$  & $0.6702\pm0.0179$  & $0.4664\pm0.0620$  & $0.3239\pm0.0266$  & $0.2333\pm0.0221$  \\
MIS-DINOv2              & ViT-L                     & $\mathbf{0.3808\pm0.0321}$ & $\mathbf{0.6917\pm0.0154}$ & $0.4734\pm0.0731$  & $0.3612\pm0.0334$  & $\mathbf{0.2620\pm0.0337}$ \\ \midrule
ImageNet                & ViT-B                     & $0.2815\pm0.0173$  & $0.5422\pm0.0278$  & $0.4597\pm0.1174$  & $0.2683\pm0.0252$  & $0.1782\pm0.0118$  \\
EndoFM                  & ViT-B                     & $0.2729\pm0.0198$  & $0.5282\pm0.0236$  & $\mathbf{0.5308\pm0.1132}$ & $0.2602\pm0.0404$  & $0.1679\pm0.0273$  \\
GSViT                   & EfficientViT              & $0.2345\pm0.0122$  & $0.4536\pm0.0166$  & $0.5173\pm0.0600$  & $0.2233\pm0.0058$  & $0.1392\pm0.0101$  \\
EndoViT                 & ViT-B                     & $0.2453\pm0.0153$  & $0.4817\pm0.0196$  & $0.4781\pm0.0970$  & $0.2268\pm0.0104$  & $0.1418\pm0.0123$  \\
PeskaVLP                & ResNet50                  & $0.3309\pm0.0245$  & $0.6221\pm0.0197$  & $0.4456\pm0.0886$  & $0.3202\pm0.0469$  & $0.2209\pm0.0300$  \\
SurgeNet                & CAFormer18                & $0.3703\pm0.0410$  & $0.6664\pm0.0298$  & $0.4546\pm0.0932$  & $0.3688\pm0.0538$  & $0.2619\pm0.0413$  \\
ZEN                     & ViT-B                     & $\mathbf{0.3951\pm0.0405}$ & $\mathbf{0.6789\pm0.0265}$ & $0.4596\pm0.0558$  & $\mathbf{0.3879\pm0.0532}$ & $\mathbf{0.2758\pm0.0471}$ \\ \bottomrule
\end{tabular}
\end{table}

\begin{table}[t]
\centering
\caption{\textbf{Few-shot evaluation on the MultiByPass140 dataset for surgical phase recognition (5-shot).} The upper block reports results for self-supervised learning pretrained models, and the lower block reports results for existing pretrained models. Best-performing models for each metric are highlighted in bold. Values are reported as mean $\pm$ standard deviation over five independent runs.}
\label{tab:mbp-fewshot-5}
\setlength{\tabcolsep}{1.5pt} 
\begin{tabular}{lcccccc}
\toprule
\multirow{2}{*}{Method} & \multirow{2}{*}{Backbone} & \multicolumn{2}{c}{Video-level} & \multicolumn{3}{c}{Phase-level} \\ \cmidrule(lr){3-4} \cmidrule(lr){5-7}
                        &                           & F1 score           & Accuracy           & Precision          & Recall             & Jaccard            \\ \midrule
MIS-MAE                 & ViT-B                     & $0.3392\pm0.0115$  & $0.5824\pm0.0269$  & $0.3855\pm0.0402$  & $0.3298\pm0.0266$  & $0.2281\pm0.0161$  \\
MIS-MoCoV3              & ViT-B                     & $0.3214\pm0.0235$  & $0.5870\pm0.0351$  & $0.3984\pm0.0870$  & $0.3085\pm0.0310$  & $0.2101\pm0.0184$  \\
MIS-MSN                 & ViT-B                     & $0.3006\pm0.0302$  & $0.5487\pm0.0223$  & $0.3984\pm0.0561$  & $0.2884\pm0.0213$  & $0.1934\pm0.0200$  \\
MIS-DINO                & ViT-B                     & $\mathbf{0.3908\pm0.0434}$ & $0.6840\pm0.0241$  & $0.4573\pm0.0456$  & $\mathbf{0.3718\pm0.0504}$ & $\mathbf{0.2680\pm0.0410}$ \\
MIS-DINOv2              & ViT-B                     & $0.3617\pm0.0298$  & $0.6668\pm0.0224$  & $0.4557\pm0.0599$  & $0.3421\pm0.0317$  & $0.2486\pm0.0201$  \\
MIS-DINOv2              & ViT-L                     & $0.3815\pm0.0359$  & $\mathbf{0.6870\pm0.0171}$ & $\mathbf{0.5315\pm0.1548}$ & $0.3671\pm0.0551$  & $0.2673\pm0.0396$  \\ \midrule
ImageNet                & ViT-B                     & $0.2860\pm0.0197$  & $0.5470\pm0.0218$  & $0.4825\pm0.1036$  & $0.2706\pm0.0217$  & $0.1790\pm0.0192$  \\
EndoFM                  & ViT-B                     & $0.2821\pm0.0171$  & $0.5322\pm0.0274$  & $0.4771\pm0.1155$  & $0.2652\pm0.0292$  & $0.1745\pm0.0167$  \\
GSViT                   & EfficientViT              & $0.2389\pm0.0067$  & $0.4613\pm0.0148$  & $0.5000\pm0.0569$  & $0.2274\pm0.0077$  & $0.1445\pm0.0080$  \\
EndoViT                 & ViT-B                     & $0.2509\pm0.0110$  & $0.4877\pm0.0231$  & $0.3973\pm0.0826$  & $0.2430\pm0.0198$  & $0.1532\pm0.0106$  \\
PeskaVLP                & ResNet50                  & $0.3566\pm0.0205$  & $0.6298\pm0.0086$  & $0.4302\pm0.0242$  & $0.3394\pm0.0329$  & $0.2403\pm0.0203$  \\
SurgeNet                & CAFormer18                & $\mathbf{0.4084\pm0.0116}$ & $0.6876\pm0.0171$  & $0.4731\pm0.0688$  & $\mathbf{0.4039\pm0.0502}$ & $\mathbf{0.2930\pm0.0302}$ \\
ZEN                     & ViT-B                     & $0.3975\pm0.0320$  & $\mathbf{0.6883\pm0.0211}$ & $\mathbf{0.5060\pm0.0787}$ & $0.3757\pm0.0497$  & $0.2764\pm0.0407$  \\ \bottomrule
\end{tabular}
\end{table}

\begin{table}[t]
\centering
\caption{\textbf{Few-shot evaluation on the AutoLaparo dataset for surgical phase recognition (1-shot).} The upper block reports results for self-supervised learning pretrained models, and the lower block reports results for existing pretrained models. Best-performing models for each metric are highlighted in bold. Values are reported as mean $\pm$ standard deviation over five independent runs.}
\label{tab:fewshot-auto-1}
\setlength{\tabcolsep}{1.5pt} 
\begin{tabular}{lcccccc}
\toprule
\multirow{2}{*}{Method} & \multirow{2}{*}{Backbone} & \multicolumn{2}{c}{Video-level} & \multicolumn{3}{c}{Phase-level} \\ \cmidrule(lr){3-4} \cmidrule(lr){5-7}
                        &                           & F1 score           & Accuracy           & Precision          & Recall             & Jaccard            \\ \midrule
MIS-MAE                 & ViT-B                     & $0.3066\pm0.0520$  & $0.4390\pm0.0804$  & $0.4586\pm0.0313$  & $0.3449\pm0.0516$  & $0.2151\pm0.0393$  \\
MIS-MoCoV3              & ViT-B                     & $0.3374\pm0.0975$  & $0.4482\pm0.1191$  & $0.4881\pm0.1022$  & $0.3654\pm0.0893$  & $0.2399\pm0.0741$  \\
MIS-MSN                 & ViT-B                     & $0.2928\pm0.0875$  & $0.4208\pm0.1013$  & $0.4423\pm0.1284$  & $0.3357\pm0.0639$  & $0.2074\pm0.0621$  \\
MIS-DINO                & ViT-B                     & $0.4508\pm0.1098$  & $0.5827\pm0.1132$  & $0.6206\pm0.0921$  & $0.4813\pm0.0711$  & $0.3357\pm0.0934$  \\
MIS-DINOv2              & ViT-B                     & $0.4244\pm0.0809$  & $0.5536\pm0.0741$  & $0.5891\pm0.0627$  & $0.4528\pm0.0617$  & $0.3077\pm0.0625$  \\
MIS-DINOv2              & ViT-L                     & $\mathbf{0.4838\pm0.1104}$ & $\mathbf{0.6237\pm0.0889}$ & $\mathbf{0.6421\pm0.0985}$ & $\mathbf{0.5165\pm0.0762}$ & $\mathbf{0.3629\pm0.0883}$ \\ \midrule
ImageNet                & ViT-B                     & $0.2968\pm0.1213$  & $0.4373\pm0.0882$  & $0.4745\pm0.1188$  & $0.3455\pm0.1045$  & $0.2071\pm0.0896$  \\
EndoFM                  & ViT-B                     & $0.2751\pm0.0556$  & $0.4003\pm0.0409$  & $0.4969\pm0.0743$  & $0.3198\pm0.0408$  & $0.1905\pm0.0359$  \\
GSViT                   & EfficientViT              & $0.2863\pm0.0965$  & $0.4313\pm0.1015$  & $0.4467\pm0.1134$  & $0.3313\pm0.0761$  & $0.2025\pm0.0702$  \\
EndoViT                 & ViT-B                     & $0.2788\pm0.0807$  & $0.4105\pm0.0450$  & $0.4860\pm0.1202$  & $0.3313\pm0.0597$  & $0.1947\pm0.0568$  \\
PeskaVLP                & ResNet50                  & $0.3439\pm0.1467$  & $0.5149\pm0.1452$  & $0.5379\pm0.1546$  & $0.3783\pm0.1198$  & $0.2571\pm0.1139$  \\
SurgeNet                & CAFormer18                & $0.3445\pm0.1193$  & $0.4972\pm0.0942$  & $0.5602\pm0.1037$  & $0.3890\pm0.0992$  & $0.2517\pm0.0936$  \\
ZEN                     & ViT-B                     & $\mathbf{0.4656\pm0.1010}$ & $\mathbf{0.6078\pm0.1089}$ & $\mathbf{0.6321\pm0.0965}$ & $\mathbf{0.4902\pm0.0706}$ & $\mathbf{0.3565\pm0.0803}$ \\ \bottomrule
\end{tabular}
\end{table}

\begin{table}[t]
\centering
\caption{\textbf{Few-shot evaluation on the AutoLaparo dataset for surgical phase recognition (2-shot).} The upper block reports results for self-supervised learning pretrained models, and the lower block reports results for existing pretrained models. Best-performing models for each metric are highlighted in bold. Values are reported as mean $\pm$ standard deviation over five independent runs.}
\label{tab:fewshot-auto-2}
\setlength{\tabcolsep}{1.5pt} 
\begin{tabular}{lcccccc}
\toprule
\multirow{2}{*}{Method} & \multirow{2}{*}{Backbone} & \multicolumn{2}{c}{Video-level} & \multicolumn{3}{c}{Phase-level} \\ \cmidrule(lr){3-4} \cmidrule(lr){5-7}
                        &                           & F1 score           & Accuracy           & Precision          & Recall             & Jaccard            \\ \midrule
MIS-MAE                 & ViT-B                     & $0.3635\pm0.0496$  & $0.5118\pm0.0647$  & $0.5364\pm0.0803$  & $0.4019\pm0.0551$  & $0.2641\pm0.0415$  \\
MIS-MoCoV3              & ViT-B                     & $0.3787\pm0.0885$  & $0.5214\pm0.1031$  & $0.4629\pm0.1145$  & $0.4074\pm0.0792$  & $0.2783\pm0.0732$  \\
MIS-MSN                 & ViT-B                     & $0.3304\pm0.0427$  & $0.4791\pm0.0562$  & $0.4651\pm0.0436$  & $0.3632\pm0.0381$  & $0.2346\pm0.0315$  \\
MIS-DINO                & ViT-B                     & $\mathbf{0.5410\pm0.0595}$ & $0.6846\pm0.0670$  & $0.6763\pm0.0688$  & $0.5530\pm0.0528$  & $0.4235\pm0.0588$  \\
MIS-DINOv2              & ViT-B                     & $0.4851\pm0.0542$  & $0.6409\pm0.0706$  & $0.6409\pm0.0743$  & $0.4945\pm0.0463$  & $0.3678\pm0.0462$  \\
MIS-DINOv2              & ViT-L                     & $0.5375\pm0.0383$  & $\mathbf{0.7081\pm0.0458}$ & $\mathbf{0.6789\pm0.0672}$ & $\mathbf{0.5562\pm0.0315}$ & $\mathbf{0.4293\pm0.0345}$ \\ \midrule
ImageNet                & ViT-B                     & $0.3409\pm0.0621$  & $0.4978\pm0.0716$  & $0.4955\pm0.0812$  & $0.3795\pm0.0464$  & $0.2421\pm0.0437$  \\
EndoFM                  & ViT-B                     & $0.3017\pm0.1075$  & $0.4656\pm0.1106$  & $0.4548\pm0.0483$  & $0.3440\pm0.0896$  & $0.2211\pm0.0839$  \\
GSViT                   & EfficientViT              & $0.3030\pm0.0652$  & $0.4577\pm0.1011$  & $0.4628\pm0.0752$  & $0.3435\pm0.0567$  & $0.2158\pm0.0494$  \\
EndoViT                 & ViT-B                     & $0.3205\pm0.0564$  & $0.4897\pm0.0786$  & $0.4635\pm0.0747$  & $0.3529\pm0.0609$  & $0.2316\pm0.0446$  \\
PeskaVLP                & ResNet50                  & $0.4717\pm0.0509$  & $0.6200\pm0.0858$  & $0.5831\pm0.0828$  & $0.4803\pm0.0449$  & $0.3552\pm0.0434$  \\
SurgeNet                & CAFormer18                & $0.5030\pm0.0554$  & $0.6470\pm0.0773$  & $0.6148\pm0.0686$  & $0.5257\pm0.0419$  & $0.3952\pm0.0489$  \\
ZEN                     & ViT-B                     & $\mathbf{0.5330\pm0.0660}$ & $\mathbf{0.6886\pm0.0678}$ & $\mathbf{0.6517\pm0.0955}$ & $\mathbf{0.5462\pm0.0436}$ & $\mathbf{0.4166\pm0.0608}$ \\ \bottomrule
\end{tabular}
\end{table}

\begin{table}[t]
\centering
\caption{\textbf{Few-shot evaluation on the AutoLaparo dataset for surgical phase recognition (3-shot).} The upper block reports results for self-supervised learning pretrained models, and the lower block reports results for existing pretrained models. Best-performing models for each metric are highlighted in bold. Values are reported as mean $\pm$ standard deviation over five independent runs.}
\label{tab:fewshot-auto-3}
\setlength{\tabcolsep}{1.5pt} 
\begin{tabular}{lcccccc}
\toprule
\multirow{2}{*}{Method} & \multirow{2}{*}{Backbone} & \multicolumn{2}{c}{Video-level} & \multicolumn{3}{c}{Phase-level} \\ \cmidrule(lr){3-4} \cmidrule(lr){5-7}
                        &                           & F1 score           & Accuracy           & Precision          & Recall             & Jaccard            \\ \midrule
MIS-MAE                 & ViT-B                     & $0.4438\pm0.0509$  & $0.5776\pm0.0563$  & $0.6240\pm0.0327$  & $0.4595\pm0.0548$  & $0.3245\pm0.0454$  \\
MIS-MoCoV3              & ViT-B                     & $0.5150\pm0.0386$  & $0.6554\pm0.0263$  & $0.6230\pm0.0334$  & $0.5284\pm0.0402$  & $0.3984\pm0.0346$  \\
MIS-MSN                 & ViT-B                     & $0.3534\pm0.0663$  & $0.5115\pm0.0740$  & $0.5882\pm0.0856$  & $0.3761\pm0.0512$  & $0.2529\pm0.0482$  \\
MIS-DINO                & ViT-B                     & $0.5993\pm0.0807$  & $0.7509\pm0.0376$  & $0.6972\pm0.0580$  & $0.6134\pm0.0679$  & $0.4912\pm0.0719$  \\
MIS-DINOv2              & ViT-B                     & $0.6126\pm0.0877$  & $0.7218\pm0.0640$  & $\mathbf{0.7125\pm0.0654}$ & $0.6274\pm0.0867$  & $0.4965\pm0.0896$  \\
MIS-DINOv2              & ViT-L                     & $\mathbf{0.6296\pm0.0660}$ & $\mathbf{0.7699\pm0.0381}$ & $0.7098\pm0.0528$  & $\mathbf{0.6531\pm0.0605}$ & $\mathbf{0.5343\pm0.0630}$ \\ \midrule
ImageNet                & ViT-B                     & $0.3782\pm0.0877$  & $0.5082\pm0.0358$  & $0.4944\pm0.0509$  & $0.4196\pm0.0893$  & $0.2717\pm0.0708$  \\
EndoFM                  & ViT-B                     & $0.4120\pm0.0439$  & $0.5600\pm0.0459$  & $0.5733\pm0.0629$  & $0.4378\pm0.0467$  & $0.3003\pm0.0425$  \\
GSViT                   & EfficientViT              & $0.3160\pm0.0460$  & $0.4596\pm0.0644$  & $0.4662\pm0.0749$  & $0.3391\pm0.0622$  & $0.2208\pm0.0381$  \\
EndoViT                 & ViT-B                     & $0.4059\pm0.0646$  & $0.5454\pm0.0909$  & $0.5884\pm0.0647$  & $0.4232\pm0.0561$  & $0.2984\pm0.0456$  \\
PeskaVLP                & ResNet50                  & $0.5615\pm0.0440$  & $0.7072\pm0.0301$  & $0.6846\pm0.0494$  & $0.5679\pm0.0415$  & $0.4482\pm0.0404$  \\
SurgeNet                & CAFormer18                & $0.5810\pm0.0988$  & $0.7111\pm0.0966$  & $0.6620\pm0.0768$  & $0.6141\pm0.0951$  & $0.4740\pm0.1026$  \\
ZEN                     & ViT-B                     & $\mathbf{0.6087\pm0.0769}$ & $\mathbf{0.7542\pm0.0412}$ & $\mathbf{0.7251\pm0.0681}$ & $\mathbf{0.6237\pm0.0623}$ & $\mathbf{0.5024\pm0.0684}$ \\ \bottomrule
\end{tabular}
\end{table}

\begin{table}[t]
\centering
\caption{\textbf{Few-shot evaluation on the AutoLaparo dataset for surgical phase recognition (4-shot).} The upper block reports results for self-supervised learning pretrained models, and the lower block reports results for existing pretrained models. Best-performing models for each metric are highlighted in bold. Values are reported as mean $\pm$ standard deviation over five independent runs.}
\label{tab:fewshot-auto-4}
\setlength{\tabcolsep}{1.5pt} 
\begin{tabular}{lcccccc}
\toprule
\multirow{2}{*}{Method} & \multirow{2}{*}{Backbone} & \multicolumn{2}{c}{Video-level} & \multicolumn{3}{c}{Phase-level} \\ \cmidrule(lr){3-4} \cmidrule(lr){5-7}
                        &                           & F1 score           & Accuracy           & Precision          & Recall             & Jaccard            \\ \midrule
MIS-MAE                 & ViT-B                     & $0.4672\pm0.0590$  & $0.6102\pm0.0391$  & $0.6428\pm0.0322$  & $0.4886\pm0.0592$  & $0.3561\pm0.0552$  \\
MIS-MoCoV3              & ViT-B                     & $0.5731\pm0.0348$  & $0.7057\pm0.0272$  & $0.6652\pm0.0188$  & $0.5926\pm0.0505$  & $0.4481\pm0.0447$  \\
MIS-MSN                 & ViT-B                     & $0.3756\pm0.0666$  & $0.5258\pm0.0384$  & $0.5899\pm0.0767$  & $0.3997\pm0.0478$  & $0.2689\pm0.0420$  \\
MIS-DINO                & ViT-B                     & $0.6656\pm0.0572$  & $0.7995\pm0.0245$  & $0.7432\pm0.0330$  & $\mathbf{0.6832\pm0.0575}$ & $0.5699\pm0.0568$  \\
MIS-DINOv2              & ViT-B                     & $0.6138\pm0.0603$  & $0.7515\pm0.0356$  & $0.7171\pm0.0194$  & $0.6339\pm0.0558$  & $0.5065\pm0.0587$  \\
MIS-DINOv2              & ViT-L                     & $\mathbf{0.6677\pm0.0510}$ & $\mathbf{0.8040\pm0.0255}$ & $\mathbf{0.7719\pm0.0360}$ & $0.6759\pm0.0498$  & $\mathbf{0.5702\pm0.0555}$ \\ \midrule
ImageNet                & ViT-B                     & $0.4690\pm0.0411$  & $0.5966\pm0.0655$  & $0.6072\pm0.0334$  & $0.5004\pm0.0338$  & $0.3542\pm0.0333$  \\
EndoFM                  & ViT-B                     & $0.4321\pm0.0622$  & $0.6000\pm0.0298$  & $0.5772\pm0.0237$  & $0.4617\pm0.0415$  & $0.3233\pm0.0463$  \\
GSViT                   & EfficientViT              & $0.3766\pm0.0678$  & $0.5544\pm0.0762$  & $0.5937\pm0.0818$  & $0.3985\pm0.0686$  & $0.2756\pm0.0508$  \\
EndoViT                 & ViT-B                     & $0.4608\pm0.0357$  & $0.5920\pm0.0445$  & $0.6049\pm0.0639$  & $0.4778\pm0.0416$  & $0.3422\pm0.0306$  \\
PeskaVLP                & ResNet50                  & $0.6103\pm0.0325$  & $0.7550\pm0.0157$  & $0.7152\pm0.0308$  & $0.6174\pm0.0332$  & $0.4972\pm0.0380$  \\
SurgeNet                & CAFormer18                & $0.6291\pm0.0287$  & $0.7594\pm0.0104$  & $0.7361\pm0.0369$  & $0.6430\pm0.0318$  & $0.5118\pm0.0308$  \\
ZEN                     & ViT-B                     & $\mathbf{0.6695\pm0.0483}$ & $\mathbf{0.8028\pm0.0154}$ & $\mathbf{0.7817\pm0.0404}$ & $\mathbf{0.6699\pm0.0426}$ & $\mathbf{0.5645\pm0.0452}$ \\ \bottomrule
\end{tabular}
\end{table}

\begin{table}[t]
\centering
\caption{\textbf{Few-shot evaluation on the AutoLaparo dataset for surgical phase recognition (5-shot).} The upper block reports results for self-supervised learning pretrained models, and the lower block reports results for existing pretrained models. Best-performing models for each metric are highlighted in bold. Values are reported as mean $\pm$ standard deviation over five independent runs.}
\label{tab:fewshot-auto-5}
\setlength{\tabcolsep}{1.5pt} 
\begin{tabular}{lcccccc}
\toprule
\multirow{2}{*}{Method} & \multirow{2}{*}{Backbone} & \multicolumn{2}{c}{Video-level} & \multicolumn{3}{c}{Phase-level} \\ \cmidrule(lr){3-4} \cmidrule(lr){5-7}
                        &                           & F1 score           & Accuracy           & Precision          & Recall             & Jaccard            \\ \midrule
MIS-MAE                 & ViT-B                     & $0.5128\pm0.0828$  & $0.6455\pm0.0529$  & $0.6532\pm0.0795$  & $0.5272\pm0.0727$  & $0.3902\pm0.0738$  \\
MIS-MoCoV3              & ViT-B                     & $0.6037\pm0.0350$  & $0.7311\pm0.0210$  & $0.6914\pm0.0359$  & $0.6092\pm0.0317$  & $0.4740\pm0.0306$  \\
MIS-MSN                 & ViT-B                     & $0.4568\pm0.0889$  & $0.5888\pm0.0765$  & $0.6394\pm0.0600$  & $0.4531\pm0.0780$  & $0.3292\pm0.0677$  \\
MIS-DINO                & ViT-B                     & $0.6647\pm0.0432$  & $0.7927\pm0.0182$  & $\mathbf{0.7634\pm0.0585}$ & $0.6675\pm0.0326$  & $0.5520\pm0.0339$  \\
MIS-DINOv2              & ViT-B                     & $0.6503\pm0.0416$  & $0.7737\pm0.0182$  & $0.7294\pm0.0217$  & $0.6597\pm0.0462$  & $0.5386\pm0.0408$  \\
MIS-DINOv2              & ViT-L                     & $\mathbf{0.7024\pm0.0232}$ & $\mathbf{0.8113\pm0.0138}$ & $0.7525\pm0.0262$  & $\mathbf{0.7218\pm0.0235}$ & $\mathbf{0.6088\pm0.0174}$ \\ \midrule
ImageNet                & ViT-B                     & $0.5429\pm0.0215$  & $0.6620\pm0.0395$  & $0.6314\pm0.0429$  & $0.5628\pm0.0181$  & $0.4213\pm0.0211$  \\
EndoFM                  & ViT-B                     & $0.4312\pm0.1189$  & $0.5667\pm0.0719$  & $0.5798\pm0.0885$  & $0.4620\pm0.0980$  & $0.3158\pm0.0947$  \\
GSViT                   & EfficientViT              & $0.3987\pm0.0582$  & $0.5505\pm0.0324$  & $0.6109\pm0.0436$  & $0.4077\pm0.0376$  & $0.2865\pm0.0418$  \\
EndoViT                 & ViT-B                     & $0.4792\pm0.0909$  & $0.6048\pm0.0648$  & $0.5863\pm0.0542$  & $0.4972\pm0.0885$  & $0.3586\pm0.0797$  \\
PeskaVLP                & ResNet50                  & $0.6222\pm0.0224$  & $0.7520\pm0.0147$  & $0.7092\pm0.0494$  & $0.6153\pm0.0229$  & $0.4997\pm0.0251$  \\
SurgeNet                & CAFormer18                & $0.6521\pm0.0612$  & $0.7710\pm0.0323$  & $0.7651\pm0.0607$  & $0.6501\pm0.0550$  & $0.5303\pm0.0553$  \\
ZEN                     & ViT-B                     & $\mathbf{0.7069\pm0.0345}$ & $\mathbf{0.8112\pm0.0128}$ & $\mathbf{0.7987\pm0.0382}$ & $\mathbf{0.7122\pm0.0477}$ & $\mathbf{0.5994\pm0.0405}$ \\ \bottomrule
\end{tabular}
\end{table}

\begin{sidewaystable}[p]
\centering
\caption{\textbf{Triplet recognition performance on the CholecT50 dataset using fine-tuned backbones.} The upper block reports results for self-supervised learning pretrained models, and the lower block reports results for existing pretrained models. Best-performing models for each metric are highlighted in bold. Values are reported as mean $\pm$ standard deviation over five independent runs.}
\label{tab:cholecT50-finetune}
\setlength{\tabcolsep}{1.5pt} 
\begin{tabular}{lccccccc}
\toprule
Method      & Backbone     & AP\textsubscript{I} & AP\textsubscript{V} & AP\textsubscript{T} & AP\textsubscript{IV} & AP\textsubscript{IT} & AP\textsubscript{IVT} \\ \midrule
MIS-MAE     & ViT-B        & $0.7486\pm0.0245$  & $0.4897\pm0.0123$  & $0.3760\pm0.0142$  & $0.2771\pm0.0081$  & $0.3144\pm0.0202$  & $0.2453\pm0.0194$  \\
MIS-MoCoV3  & ViT-B        & $0.7920\pm0.0075$  & $0.5092\pm0.0042$  & $0.3697\pm0.0136$  & $0.2959\pm0.0041$  & $0.3223\pm0.0116$  & $0.2497\pm0.0087$  \\
MIS-MSN     & ViT-B        & $0.8354\pm0.0066$  & $0.5534\pm0.0067$  & $0.4000\pm0.0095$  & $0.3203\pm0.0046$  & $0.3556\pm0.0115$  & $0.2789\pm0.0104$  \\
MIS-DINO    & ViT-B        & $0.8361\pm0.0119$  & $0.5559\pm0.0085$  & $0.3970\pm0.0089$  & $0.3151\pm0.0042$  & $0.3398\pm0.0141$  & $0.2663\pm0.0078$  \\
MIS-DINOv2  & ViT-B        & $0.8280\pm0.0060$  & $0.5523\pm0.0114$  & $0.4243\pm0.0243$  & $0.3216\pm0.0089$  & $0.3635\pm0.0176$  & $0.2813\pm0.0125$  \\
MIS-DINOv2  & ViT-L        & $\mathbf{0.8540\pm0.0051}$ & $\mathbf{0.5697\pm0.0162}$ & $\mathbf{0.4326\pm0.0152}$ & $\mathbf{0.3352\pm0.0128}$ & $\mathbf{0.3800\pm0.0147}$ & $\mathbf{0.2937\pm0.0130}$ \\ \midrule
ImageNet    & ViT-B        & $0.8091\pm0.0131$  & $0.5284\pm0.0120$  & $0.3940\pm0.0152$  & $0.3026\pm0.0112$  & $0.3409\pm0.0134$  & $0.2657\pm0.0158$  \\
EndoFM      & ViT-B        & $0.7818\pm0.0178$  & $0.5106\pm0.0134$  & $0.3739\pm0.0099$  & $0.2897\pm0.0066$  & $0.3277\pm0.0065$  & $0.2520\pm0.0108$  \\
GSViT       & EfficientViT & $0.3227\pm0.0113$  & $0.2112\pm0.0059$  & $0.1591\pm0.0046$  & $0.1108\pm0.0051$  & $0.0843\pm0.0023$  & $0.0676\pm0.0058$  \\
EndoViT     & ViT-B        & $0.7324\pm0.0205$  & $0.4863\pm0.0094$  & $0.3658\pm0.0075$  & $0.2730\pm0.0069$  & $0.3119\pm0.0171$  & $0.2449\pm0.0112$  \\
PeskaVLP    & ResNet50     & $0.8124\pm0.0046$  & $0.5458\pm0.0092$  & $0.3858\pm0.0113$  & $0.3152\pm0.0071$  & $0.3419\pm0.0106$  & $0.2604\pm0.0123$  \\
SurgeNet    & CAFormer18   & $0.8322\pm0.0148$  & $0.5461\pm0.0197$  & $0.3980\pm0.0184$  & $0.3173\pm0.0179$  & $0.3383\pm0.0242$  & $0.2691\pm0.0158$  \\
ZEN         & ViT-B        & $\mathbf{0.8611\pm0.0043}$ & $\mathbf{0.5898\pm0.0079}$ & $\mathbf{0.4607\pm0.0170}$ & $\mathbf{0.3515\pm0.0033}$ & $\mathbf{0.4015\pm0.0112}$ & $\mathbf{0.3207\pm0.0135}$ \\ \bottomrule
\end{tabular}
\end{sidewaystable}

\begin{sidewaystable}[p]
\centering
\caption{\textbf{Triplet recognition performance on the LLS48 dataset using fine-tuned backbones.} The upper block reports results for self-supervised learning pretrained models, and the lower block reports results for existing pretrained models. Best-performing models for each metric are highlighted in bold. Values are reported as mean $\pm$ standard deviation over five independent runs.}
\label{tab:lls48-finetune}
\setlength{\tabcolsep}{1.5pt} 
\begin{tabular}{lccccccc}
\toprule
Method      & Backbone     & AP\textsubscript{I} & AP\textsubscript{V} & AP\textsubscript{T} & AP\textsubscript{IV} & AP\textsubscript{IT} & AP\textsubscript{IVT} \\ \midrule
MIS-MAE     & ViT-B        & $0.4309\pm0.0108$  & $0.2895\pm0.0166$  & $0.2780\pm0.0224$  & $0.2598\pm0.0079$  & $0.2337\pm0.0205$  & $0.2276\pm0.0142$  \\
MIS-MoCoV3  & ViT-B        & $0.4640\pm0.0139$  & $0.3108\pm0.0257$  & $0.2751\pm0.0132$  & $0.2701\pm0.0132$  & $0.2015\pm0.0157$  & $0.1677\pm0.0088$  \\
MIS-MSN     & ViT-B        & $0.5857\pm0.0144$  & $\mathbf{0.4828\pm0.0344}$  & $0.3760\pm0.0201$  & $0.3893\pm0.0112$  & $0.3362\pm0.0218$  & $ 0.3138\pm0.0085$  \\
MIS-DINO    & ViT-B        & $0.5930\pm0.0146$  & $0.4555\pm0.0084$  & $\mathbf{0.3999\pm0.0290}$  & $0.3923\pm0.0188$  & $\mathbf{0.3417\pm0.0239}$  & $\mathbf{0.3138\pm0.0198}$  \\
MIS-DINOv2  & ViT-B        & $0.5240\pm0.0206$  & $0.4025\pm0.0384$  & $0.3348\pm0.0298$  & $0.3518\pm0.0277$  & $0.2926\pm0.0204$  & $0.2727\pm0.0107$  \\
MIS-DINOv2  & ViT-L        & $\mathbf{0.5978\pm0.0305}$ & $0.4790\pm0.0549$  & $0.3866\pm0.0347$  & $\mathbf{0.4029\pm0.0328}$ & $0.3376\pm0.0290$  & $0.3114\pm0.0197$  \\ \midrule
ImageNet    & ViT-B        & $0.5422\pm0.0280$  & $0.4061\pm0.0284$  & $0.3609\pm0.0246$  & $0.3599\pm0.0138$  & $0.2976\pm0.0117$  & $0.2768\pm0.0071$  \\
EndoFM      & ViT-B        & $0.5202\pm0.0243$  & $0.3789\pm0.0476$  & $0.3290\pm0.0176$  & $0.3439\pm0.0225$  & $0.2870\pm0.0173$  & $0.2672\pm0.0112$  \\
GSViT       & EfficientViT & $0.2310\pm0.0314$  & $0.2001\pm0.0357$  & $0.1443\pm0.0228$  & $0.1470\pm0.0220$  & $0.0909\pm0.0171$  & $0.0969\pm0.0149$  \\
EndoViT     & ViT-B        & $0.4374\pm0.0131$  & $0.3095\pm0.0213$  & $0.2752\pm0.0133$  & $0.2772\pm0.0134$  & $0.2366\pm0.0161$  & $0.2203\pm0.0130$  \\
PeskaVLP    & ResNet50     & $0.5469\pm0.0160$  & $0.4481\pm0.0232$  & $0.3929\pm0.0218$  & $0.3838\pm0.0185$  & $0.3329\pm0.0192$  & $0.2956\pm0.0134$  \\
SurgeNet    & CAFormer18   & $0.5375\pm0.0217$  & $0.3932\pm0.0385$  & $0.3376\pm0.0216$  & $0.3495\pm0.0352$  & $0.3035\pm0.0162$  & $0.2731\pm0.0193$  \\
ZEN         & ViT-B        & $\mathbf{0.5979\pm0.0283}$ & $\mathbf{0.5086\pm0.0545}$ & $\mathbf{0.4243\pm0.0269}$ & $\mathbf{0.4388\pm0.0345}$ & $\mathbf{0.3701\pm0.0250}$ & $\mathbf{0.3419\pm0.0162}$ \\ \bottomrule
\end{tabular}
\end{sidewaystable}

\begin{sidewaystable}[p]
\centering
\caption{\textbf{Triplet recognition performance on the CholecT50 dataset using frozen backbones.} The upper block reports results for self-supervised learning pretrained models, and the lower block reports results for existing pretrained models. Best-performing models for each metric are highlighted in bold. Values are reported as mean $\pm$ standard deviation over five independent runs.}
\label{tab:cholecT50-freeze}
\setlength{\tabcolsep}{1.5pt} 
\begin{tabular}{lccccccc}
\toprule
Method      & Backbone     & AP\textsubscript{I} & AP\textsubscript{V} & AP\textsubscript{T} & AP\textsubscript{IV} & AP\textsubscript{IT} & AP\textsubscript{IVT} \\ \midrule
MIS-MAE     & ViT-B        & $0.3770\pm0.0027$  & $0.2645\pm0.0038$  & $0.2178\pm0.0061$  & $0.1354\pm0.0013$  & $0.1239\pm0.0051$  & $0.0927\pm0.0042$  \\
MIS-MoCoV3  & ViT-B        & $0.4629\pm0.0051$  & $0.2993\pm0.0037$  & $0.2357\pm0.0011$  & $0.1482\pm0.0015$  & $0.1219\pm0.0009$  & $0.0876\pm0.0006$  \\
MIS-MSN     & ViT-B        & $0.3869\pm0.0081$  & $0.2794\pm0.0048$  & $0.2546\pm0.0150$  & $0.1417\pm0.0046$  & $0.1341\pm0.0033$  & $0.1044\pm0.0084$  \\
MIS-DINO    & ViT-B        & $0.5999\pm0.0200$  & $0.4122\pm0.0089$  & $0.3590\pm0.0025$  & $0.2258\pm0.0080$  & $0.2489\pm0.0050$  & $0.1986\pm0.0059$  \\
MIS-DINOv2  & ViT-B        & $0.6980\pm0.0098$  & $0.4698\pm0.0043$  & $0.3554\pm0.0048$  & $0.2571\pm0.0030$  & $0.2777\pm0.0102$  & $0.2146\pm0.0076$  \\
MIS-DINOv2  & ViT-L        & $\mathbf{0.8053\pm0.0020}$ & $\mathbf{0.5348\pm0.0053}$ & $\mathbf{0.4090\pm0.0074}$ & $\mathbf{0.3072\pm0.0049}$ & $\mathbf{0.3348\pm0.0081}$ & $\mathbf{0.2634\pm0.0036}$ \\ \midrule
ImageNet    & ViT-B        & $0.5222\pm0.0033$  & $0.3520\pm0.0025$  & $0.2509\pm0.0051$  & $0.1851\pm0.0041$  & $0.1536\pm0.0049$  & $0.1112\pm0.0026$  \\
EndoFM      & ViT-B        & $0.4315\pm0.0066$  & $0.3095\pm0.0033$  & $0.2362\pm0.0025$  & $0.1549\pm0.0024$  & $0.1419\pm0.0018$  & $0.1099\pm0.0038$  \\
GSViT       & EfficientViT & $0.2624\pm0.0032$  & $0.1668\pm0.0022$  & $0.1209\pm0.0019$  & $0.0838\pm0.0012$  & $0.0581\pm0.0010$  & $0.0459\pm0.0015$  \\
EndoViT     & ViT-B        & $0.3817\pm0.0095$  & $0.2712\pm0.0037$  & $0.2033\pm0.0025$  & $0.1381\pm0.0022$  & $0.1197\pm0.0034$  & $0.0915\pm0.0028$  \\
PeskaVLP    & ResNet50     & $0.5624\pm0.0067$  & $0.3992\pm0.0048$  & $0.2646\pm0.0073$  & $0.2045\pm0.0032$  & $0.1676\pm0.0033$  & $0.1321\pm0.0034$  \\
SurgeNet    & CAFormer18   & $0.6110\pm0.0112$  & $0.4363\pm0.0073$  & $0.3472\pm0.0094$  & $0.2284\pm0.0038$  & $0.2420\pm0.0145$  & $0.1897\pm0.0060$  \\
ZEN         & ViT-B        & $\mathbf{0.8048\pm0.0036}$ & $\mathbf{0.5253\pm0.0037}$ & $\mathbf{0.4170\pm0.0068}$ & $\mathbf{0.3012\pm0.0048}$ & $\mathbf{0.3401\pm0.0063}$ & $\mathbf{0.2654\pm0.0124}$ \\ \bottomrule
\end{tabular}
\end{sidewaystable}

\begin{sidewaystable}[p]
\centering
\caption{\textbf{Triplet recognition performance on the LLS48 dataset using frozen backbones.} The upper block reports results for self-supervised learning pretrained models, and the lower block reports results for existing pretrained models. Best-performing models for each metric are highlighted in bold. Values are reported as mean $\pm$ standard deviation over five independent runs.}
\label{tab:lls48-freeze}
\setlength{\tabcolsep}{1.5pt} 
\begin{tabular}{lccccccc}
\toprule
Method      & Backbone     & AP\textsubscript{I} & AP\textsubscript{V} & AP\textsubscript{T} & AP\textsubscript{IV} & AP\textsubscript{IT} & AP\textsubscript{IVT} \\ \midrule
MIS-MAE     & ViT-B        & $0.3274\pm0.0239$  & $0.1933\pm0.0192$  & $0.1913\pm0.0167$  & $0.1886\pm0.0160$  & $0.1301\pm0.0113$  & $0.1308\pm0.0119$  \\
MIS-MoCoV3  & ViT-B        & $0.3759\pm0.0063$  & $0.2176\pm0.0081$  & $0.1981\pm0.0056$  & $0.1874\pm0.0062$  & $0.1238\pm0.0049$  & $0.1130\pm0.0036$  \\
MIS-MSN     & ViT-B        & $0.4419\pm0.0503$  & $0.3238\pm0.0441$  & $0.2404\pm0.0121$  & $0.2623\pm0.0295$  & $0.2206\pm0.0096$  & $0.1742\pm0.0033$  \\
MIS-DINO    & ViT-B        & $0.5764\pm0.0151$  & $0.4519\pm0.0304$  & $0.3621\pm0.0184$  & $0.3588\pm0.0122$  & $0.3045\pm0.0169$  & $0.2834\pm0.0109$  \\
MIS-DINOv2  & ViT-B        & $0.6014\pm0.0209$  & $0.4922\pm0.0685$  & $0.3876\pm0.0352$  & $0.3948\pm0.0266$  & $0.3090\pm0.0153$  & $0.2829\pm0.0084$  \\
MIS-DINOv2  & ViT-L        & $\mathbf{0.6138\pm0.0085}$ & $\mathbf{0.5678\pm0.0350}$ & $\mathbf{0.4799\pm0.0082}$ & $\mathbf{0.4653\pm0.0088}$ & $\mathbf{0.3716\pm0.0113}$ & $\mathbf{0.3476\pm0.0065}$ \\ \midrule
ImageNet    & ViT-B        & $0.3815\pm0.0146$  & $0.3114\pm0.0194$  & $0.2036\pm0.0065$  & $0.2410\pm0.0101$  & $0.1636\pm0.0027$  & $0.1415\pm0.0032$  \\
EndoFM      & ViT-B        & $0.3685\pm0.0069$  & $0.2596\pm0.0155$  & $0.1892\pm0.0082$  & $0.2196\pm0.0069$  & $0.1423\pm0.0024$  & $0.1319\pm0.0036$  \\
GSViT       & EfficientViT & $0.1642\pm0.0102$  & $0.1315\pm0.0121$  & $0.1098\pm0.0130$  & $0.0895\pm0.0054$  & $0.0712\pm0.0059$  & $0.0717\pm0.0073$  \\
EndoViT     & ViT-B        & $0.2985\pm0.0226$  & $0.1896\pm0.0226$  & $0.1794\pm0.0176$  & $0.1709\pm0.0080$  & $0.1213\pm0.0063$  & $0.1089\pm0.0035$  \\
PeskaVLP    & ResNet50     & $0.3775\pm0.0048$  & $0.3247\pm0.0359$  & $0.2328\pm0.0133$  & $0.2677\pm0.0190$  & $0.1689\pm0.0070$  & $0.1570\pm0.0015$  \\
SurgeNet    & CAFormer18   & $0.4892\pm0.0348$  & $0.3689\pm0.0452$  & $0.2915\pm0.0060$  & $0.3070\pm0.0238$  & $0.2494\pm0.0170$  & $0.2383\pm0.0094$  \\
ZEN         & ViT-B        & $\mathbf{0.5710\pm0.0162}$ & $\mathbf{0.5163\pm0.0141}$ & $\mathbf{0.3769\pm0.0192}$ & $\mathbf{0.4323\pm0.0154}$ & $\mathbf{0.3161\pm0.0147}$ & $\mathbf{0.2924\pm0.0088}$ \\ \bottomrule
\end{tabular}
\end{sidewaystable}

\begin{table}[t]
\centering
\caption{\textbf{Skill assessment performance on the Cholec80-CVS dataset with frozen and fine-tuned backbones.} The upper block reports results for self-supervised learning pretrained models, and the lower block reports results for existing pretrained models. Performance is evaluated using mAP and macro F1 score for both frozen and fine-tuned backbone settings. Best-performing models for each metric are highlighted in bold. Values are reported as mean $\pm$ standard deviation over five independent runs.}
\label{tab:cholec80-cvs}
\setlength{\tabcolsep}{1.5pt} 
\begin{tabular}{lccccc}
\toprule
\multirow{2}{*}{Method} & \multirow{2}{*}{Backbone} & \multicolumn{2}{c}{Frozen backbone} & \multicolumn{2}{c}{Fine-tuned backbone} \\ \cmidrule(lr){3-4} \cmidrule(lr){5-6}
                        &                           & mAP                & F1 score           & mAP                & F1 score           \\ \midrule
MIS-MAE                 & ViT-B                     & $0.3251\pm0.0183$  & $0.1761\pm0.0510$  & $0.3676\pm0.0129$  & $0.2453\pm0.0837$  \\
MIS-MoCoV3              & ViT-B                     & $0.3559\pm0.0292$  & $0.2280\pm0.0106$  & $0.3665\pm0.0178$  & $0.3065\pm0.0382$  \\
MIS-MSN                 & ViT-B                     & $0.3111\pm0.0138$  & $0.2021\pm0.0498$  & $0.3509\pm0.0135$  & $\mathbf{0.3390\pm0.0617}$ \\
MIS-DINO                & ViT-B                     & $\mathbf{0.3606\pm0.0361}$ & $\mathbf{0.3389\pm0.0258}$ & $0.3628\pm0.0307$  & $0.3346\pm0.0220$  \\
MIS-DINOv2              & ViT-B                     & $0.3526\pm0.0088$  & $0.2679\pm0.0223$  & $\mathbf{0.3714\pm0.0262}$ & $0.3062\pm0.0221$  \\
MIS-DINOv2              & ViT-L                     & $0.3557\pm0.0101$  & $0.3167\pm0.0294$  & $0.3695\pm0.0234$  & $0.3038\pm0.0477$  \\ \midrule
ImageNet                & ViT-B                     & $0.3684\pm0.0209$  & $0.2821\pm0.0190$  & $0.3722\pm0.0303$  & $0.2976\pm0.0309$  \\
EndoFM                  & ViT-B                     & $0.3278\pm0.0219$  & $0.1721\pm0.0366$  & $0.3595\pm0.0172$  & $0.2952\pm0.0323$  \\
GSViT                   & EfficientViT              & $0.3126\pm0.0203$  & $0.2648\pm0.0535$  & $0.3037\pm0.0185$  & $0.2088\pm0.0487$  \\
EndoViT                 & ViT-B                     & $0.3062\pm0.0213$  & $0.1118\pm0.1154$  & $0.3568\pm0.0134$  & $0.2662\pm0.0202$  \\
PeskaVLP                & ResNet50                  & $0.3692\pm0.0306$  & $0.2331\pm0.0220$  & $0.3815\pm0.0094$  & $0.2748\pm0.0172$  \\
SurgeNet                & CAFormer18                & $0.3305\pm0.0312$  & $0.3265\pm0.0354$  & $0.3468\pm0.0136$  & $0.2872\pm0.0250$  \\
ZEN                     & ViT-B                     & $\mathbf{0.3820\pm0.0242}$ & $\mathbf{0.3698\pm0.0682}$ & $\mathbf{0.3856\pm0.0230}$ & $\mathbf{0.3322\pm0.0446}$ \\ \bottomrule
\end{tabular}
\end{table}

\begin{sidewaystable}[p]
\centering
\caption{\textbf{Semantic segmentation performance with fine-tuned backbones.} Results are reported on CholecSeg8k, DSAD, and GraSP datasets using Dice similarity coefficient (DSC, higher is better) and 95th percentile Hausdorff distance (HD95, lower is better). The upper block reports results for self-supervised learning pretrained models, and the lower block reports results for existing pretrained models. Best-performing models for each metric are highlighted in bold. Values are reported as mean $\pm$ standard deviation over five independent runs.}
\label{tab:semseg-finetune}
\setlength{\tabcolsep}{1.5pt} 
\begin{tabular}{lccccccc}
\toprule
\multirow{2}{*}{Method} & \multirow{2}{*}{Backbone} & \multicolumn{2}{c}{CholecSeg8k} & \multicolumn{2}{c}{DSAD} & \multicolumn{2}{c}{GraSP} \\ \cmidrule(lr){3-4} \cmidrule(lr){5-6} \cmidrule(lr){7-8}
                        &                           & DSC $\uparrow$ & HD95 $\downarrow$ & DSC $\uparrow$ & HD95 $\downarrow$ & DSC $\uparrow$ & HD95 $\downarrow$ \\ \midrule
MIS-MAE                 & ViT-B                     & $0.6916\pm0.0280$  & $34.8959\pm6.9566$ & $0.2458\pm0.0277$  & $62.1893\pm5.7180$ & $0.6744\pm0.0393$  & $20.2501\pm5.0924$ \\
MIS-MoCoV3              & ViT-B                     & $0.7828\pm0.0145$  & $26.8623\pm1.6089$ & $0.3217\pm0.0206$  & $45.5967\pm2.9430$ & $0.7280\pm0.0052$  & $16.9154\pm0.7323$ \\
MIS-MSN                 & ViT-B                     & $0.7990\pm0.0059$  & $23.7916\pm1.6266$ & $0.4027\pm0.0526$  & $44.9000\pm3.5351$ & $0.7380\pm0.0072$  & $15.9382\pm0.7840$ \\
MIS-DINO                & ViT-B                     & $\mathbf{0.8020\pm0.0095}$ & $22.3890\pm2.2561$ & $\mathbf{0.4087\pm0.0437}$ & $\mathbf{40.0732\pm3.1323}$ & $0.7538\pm0.0052$  & $14.2806\pm1.0306$ \\
MIS-DINO v2             & ViT-B                     & $0.8018\pm0.0122$  & $\mathbf{21.2345\pm1.0944}$ & $0.3621\pm0.0411$  & $40.9867\pm3.2428$ & $\mathbf{0.7743\pm0.0057}$ & $\mathbf{12.0440\pm0.4085}$ \\
MIS-DINO v2             & ViT-L                     & $0.7192\pm0.0564$  & $34.5400\pm8.2258$ & $0.1982\pm0.0422$  & $61.4032\pm8.9463$ & $0.6367\pm0.1327$  & $24.0193\pm11.5118$ \\ \midrule
ImageNet                & ViT-B                     & $0.7554\pm0.0201$  & $27.7075\pm1.9578$ & $0.2739\pm0.0276$  & $48.8271\pm3.3750$ & $0.7044\pm0.0087$  & $17.5523\pm1.2971$ \\
EndoFM                  & ViT-B                     & $0.7190\pm0.0259$  & $35.1930\pm3.1623$ & $0.2542\pm0.0269$  & $58.0185\pm3.5420$ & $0.6766\pm0.0122$  & $22.4027\pm1.9139$ \\
GSViT                   & EfficientViT              & $0.5961\pm0.0170$  & $46.4264\pm3.5436$ & $0.2167\pm0.0259$  & $60.5168\pm7.3890$ & $0.4996\pm0.0168$  & $37.0787\pm2.7134$ \\
EndoViT                 & ViT-B                     & $0.7018\pm0.0133$  & $30.8534\pm2.4691$ & $0.2546\pm0.0330$  & $60.6693\pm3.4926$ & $0.6925\pm0.0093$  & $19.1756\pm1.4426$ \\
PeskaVLP                & ResNet50                  & $0.7530\pm0.0192$  & $29.6100\pm1.3554$ & $0.3225\pm0.0353$  & $50.9648\pm2.6621$ & $0.7454\pm0.0110$  & $17.6964\pm1.3970$ \\
SurgeNet                & CAFormer18                & $0.6585\pm0.0290$  & $43.4978\pm5.5861$ & $0.2324\pm0.0279$  & $67.4836\pm7.4230$ & $0.5758\pm0.0255$  & $33.1859\pm2.4024$ \\
ZEN                     & ViT-B                     & $\mathbf{0.8187\pm0.0089}$ & $\mathbf{20.4475\pm1.4726}$ & $\mathbf{0.3754\pm0.0205}$ & $\mathbf{36.3318\pm1.7441}$ & $\mathbf{0.7812\pm0.0038}$ & $\mathbf{11.5334\pm0.7093}$ \\ \bottomrule
\end{tabular}
\end{sidewaystable}

\begin{sidewaystable}[p]
\centering
\caption{\textbf{Semantic segmentation performance with frozen backbones.} Results are reported on CholecSeg8k, DSAD, and GraSP datasets using Dice similarity coefficient (DSC, higher is better) and 95th percentile Hausdorff distance (HD95, lower is better). The upper block reports results for self-supervised learning pretrained models, and the lower block reports results for existing pretrained models. Best-performing models for each metric are highlighted in bold. Values are reported as mean $\pm$ standard deviation over five independent runs.}
\label{tab:semseg-frozen}
\setlength{\tabcolsep}{1.5pt} 
\begin{tabular}{lccccccc}
\toprule
\multirow{2}{*}{Method} & \multirow{2}{*}{Backbone} & \multicolumn{2}{c}{CholecSeg8k} & \multicolumn{2}{c}{DSAD} & \multicolumn{2}{c}{GraSP} \\ \cmidrule(lr){3-4} \cmidrule(lr){5-6} \cmidrule(lr){7-8}
                        &                           & DSC $\uparrow$ & HD95 $\downarrow$ & DSC $\uparrow$ & HD95 $\downarrow$ & DSC $\uparrow$ & HD95 $\downarrow$ \\ \midrule
MIS-MAE                 & ViT-B                     & $0.6861\pm0.0255$  & $38.8152\pm2.5011$ & $0.2277\pm0.0124$  & $66.1680\pm3.1127$ & $0.5938\pm0.0212$  & $29.2246\pm1.7466$ \\
MIS-MoCoV3              & ViT-B                     & $0.7348\pm0.0239$  & $32.6056\pm1.9152$ & $0.2477\pm0.0207$  & $54.5426\pm4.4232$ & $0.6254\pm0.0139$  & $25.2477\pm1.4299$ \\
MIS-MSN                 & ViT-B                     & $0.7388\pm0.0091$  & $32.6919\pm2.6873$ & $0.3109\pm0.0420$  & $53.0506\pm3.1128$ & $0.6418\pm0.0046$  & $25.4631\pm0.7746$ \\
MIS-DINO                & ViT-B                     & $0.7701\pm0.0066$  & $26.7690\pm1.6323$ & $0.3090\pm0.0239$  & $47.0333\pm5.1523$ & $0.6833\pm0.0108$  & $19.8921\pm0.9109$ \\
MIS-DINOv2              & ViT-B                     & $0.7924\pm0.0127$  & $26.9685\pm3.6271$ & $0.3571\pm0.0428$  & $44.3142\pm3.1016$ & $0.7201\pm0.0085$  & $18.2404\pm1.7884$ \\
MIS-DINOv2              & ViT-L                     & $\mathbf{0.7995\pm0.0106}$ & $\mathbf{21.4576\pm1.7789}$ & $\mathbf{0.4033\pm0.0449}$ & $\mathbf{41.6426\pm2.7067}$ & $\mathbf{0.7639\pm0.0033}$ & $\mathbf{13.9271\pm0.5646}$ \\ \midrule
ImageNet                & ViT-B                     & $0.7166\pm0.0120$  & $40.8932\pm1.1910$ & $0.2506\pm0.0133$  & $56.3714\pm1.3337$ & $0.5745\pm0.0267$  & $32.1888\pm2.0011$ \\
EndoFM                  & ViT-B                     & $0.6679\pm0.0318$  & $43.6707\pm2.4823$ & $0.2330\pm0.0203$  & $64.9675\pm10.4080$ & $0.5656\pm0.0110$  & $39.3725\pm1.6396$ \\
GSViT                   & EfficientViT              & $0.4996\pm0.0172$  & $56.9302\pm5.2608$ & $0.1592\pm0.0071$  & $68.7672\pm2.9062$ & $0.3348\pm0.0232$  & $54.9375\pm2.7877$ \\
EndoViT                 & ViT-B                     & $0.6564\pm0.0269$  & $34.8515\pm1.6115$ & $0.2200\pm0.0097$  & $68.0728\pm2.8587$ & $0.6314\pm0.0174$  & $27.3860\pm1.9312$ \\
PeskaVLP                & ResNet50                  & $0.7331\pm0.0146$  & $40.1011\pm1.7930$ & $0.2822\pm0.0319$  & $59.5684\pm2.1167$ & $0.6550\pm0.0058$  & $31.0452\pm1.8067$ \\
SurgeNet                & CAFormer18                & $0.7199\pm0.0210$  & $36.6993\pm4.8076$ & $0.2917\pm0.0215$  & $50.7239\pm3.7081$ & $0.7193\pm0.0057$  & $22.4122\pm1.4450$ \\
ZEN                     & ViT-B                     & $\mathbf{0.7876\pm0.0109}$ & $\mathbf{21.3993\pm2.5680}$ & $\mathbf{0.3472\pm0.0272}$ & $\mathbf{40.7567\pm1.9581}$ & $\mathbf{0.7529\pm0.0057}$ & $\mathbf{15.1157\pm0.5242}$ \\ \bottomrule
\end{tabular}
\end{sidewaystable}

\begin{sidewaystable}[p]
\centering
\caption{\textbf{Few-shot evaluation across three datasets for semantic segmentation (1-shot).}
Results are reported using the Dice similarity coefficient (DSC; higher is better) and the 95th percentile Hausdorff distance (HD95; lower is better). The upper block reports results for self-supervised learning pretrained models, and the lower block reports results for existing pretrained models. Best-performing models for each metric are highlighted in bold. Values are reported as mean $\pm$ standard deviation over five independent runs.}
\label{tab:seg_1shot}
\setlength{\tabcolsep}{1.5pt} 
\begin{tabular}{llcccccc}
\toprule
\multirow{2}{*}{Method} &
\multirow{2}{*}{Backbone} &
\multicolumn{2}{c}{CholecSeg8k} &
\multicolumn{2}{c}{DSAD} &
\multicolumn{2}{c}{GraSP} \\
\cmidrule(lr){3-4} \cmidrule(lr){5-6} \cmidrule(lr){7-8}
 &  & DSC$\uparrow$ & HD95$\downarrow$ & DSC$\uparrow$ & HD95$\downarrow$ & DSC$\uparrow$ & HD95$\downarrow$ \\
\midrule
MIS-MAE        & ViT-B &
0.3792$\pm$0.0791 & 76.5120$\pm$15.2150 &
0.0955$\pm$0.0429 & 114.7797$\pm$24.3907 &
0.2316$\pm$0.0360 & 67.7123$\pm$9.5330 \\
MIS-MoCov3     & ViT-B &
0.3959$\pm$0.0968 & 73.9893$\pm$20.7903 &
0.1270$\pm$0.0504 & 112.6716$\pm$26.6163 &
0.2874$\pm$0.0417 & 52.4048$\pm$15.4301 \\
MSN            & ViT-B &
0.3913$\pm$0.1132 & 78.7525$\pm$22.3983 &
0.1235$\pm$0.0518 & 107.4917$\pm$25.3223 &
0.2936$\pm$0.0297 & 49.9007$\pm$4.3906 \\
MIS-DINO       & ViT-B &
0.4297$\pm$0.1257 & 73.8078$\pm$24.3163 &
0.1357$\pm$0.0555 & 102.2998$\pm$33.5338 &
0.3483$\pm$0.0264 & 38.8752$\pm$6.2540 \\
MIS-DINOv2     & ViT-B &
0.4384$\pm$0.1211 & 65.0300$\pm$18.9150 &
0.1568$\pm$0.0413 & \textbf{85.0391$\pm$17.1377} &
0.3760$\pm$0.0459 & 34.9997$\pm$5.9416 \\
MIS-DINOv2     & ViT-L &
\textbf{0.4591$\pm$0.1409} & \textbf{64.7460$\pm$20.5177} &
\textbf{0.1585$\pm$0.0534} & 95.6877$\pm$23.6925 &
\textbf{0.3908$\pm$0.0431} & \textbf{29.7430$\pm$2.4641} \\

\midrule
Supervised     & ViT-B        & 0.3994$\pm$0.1185 & 77.1273$\pm$21.9425 & 0.0990$\pm$0.0406 & 137.6828$\pm$32.7957 & 0.2538$\pm$0.0259 & 67.6870$\pm$5.2851 \\
EndoFM         & ViT-B        & 0.3644$\pm$0.0765 & 83.1372$\pm$14.0706 & 0.1031$\pm$0.0316 & 130.1707$\pm$7.9361  & 0.2082$\pm$0.0598 & 79.4412$\pm$22.9155 \\
GSViT          & EfficientViT & 0.2285$\pm$0.0615 & 96.7734$\pm$20.9503 & 0.0810$\pm$0.0194 & 116.2613$\pm$23.4674 & 0.1698$\pm$0.0295 & 78.6745$\pm$10.3655 \\
EndoViT        & ViT-B        & 0.3765$\pm$0.1021 & 79.1075$\pm$21.9769 & 0.0994$\pm$0.0314 & 126.4545$\pm$19.8909 & 0.2369$\pm$0.0238 & 65.7764$\pm$14.1126 \\
PeskaVLP       & ResNet50     & 0.4063$\pm$0.0797 & 77.7464$\pm$11.2262 & 0.0937$\pm$0.0552 & 131.3237$\pm$54.3932 & 0.2984$\pm$0.0517 & 68.9799$\pm$8.2798 \\
SurgeNet       & CAFormer18   & 0.3532$\pm$0.1166 & 84.5378$\pm$20.5921 & 0.0797$\pm$0.0471 & 133.2218$\pm$55.9615 & 0.2752$\pm$0.1302 & 46.1387$\pm$3.6071 \\
ZEN            & ViT-B        & \textbf{0.4541$\pm$0.1370} & \textbf{62.9728$\pm$19.2575} & \textbf{0.1373$\pm$0.0414} & \textbf{106.0836$\pm$34.0479} & \textbf{0.3851$\pm$0.0528} & \textbf{35.5378$\pm$6.8311} \\
\bottomrule
\end{tabular}
\end{sidewaystable}

\begin{sidewaystable}[p]
\centering
\caption{\textbf{Few-shot evaluation across three datasets for semantic segmentation (2-shot).}
Results are reported using the Dice similarity coefficient (DSC; higher is better) and the 95th percentile Hausdorff distance (HD95; lower is better). The upper block reports results for self-supervised learning pretrained models, and the lower block reports results for existing pretrained models. Best-performing models for each metric are highlighted in bold. Values are reported as mean $\pm$ standard deviation over five independent runs.}
\label{tab:seg_2shot}
\setlength{\tabcolsep}{1.5pt} 
\begin{tabular}{llcccccc}
\toprule
\multirow{2}{*}{Method} &
\multirow{2}{*}{Backbone} &
\multicolumn{2}{c}{CholecSeg8k} &
\multicolumn{2}{c}{DSAD} &
\multicolumn{2}{c}{GraSP} \\
\cmidrule(lr){3-4} \cmidrule(lr){5-6} \cmidrule(lr){7-8}
 &  & DSC$\uparrow$ & HD95$\downarrow$ & DSC$\uparrow$ & HD95$\downarrow$ & DSC$\uparrow$ & HD95$\downarrow$ \\
\midrule
MIS-MAE        & ViT-B &
0.4866$\pm$0.0683 & 68.9524$\pm$12.0818 &
0.1017$\pm$0.0554 & 117.0918$\pm$50.6241 &
0.3081$\pm$0.0316 & 54.2737$\pm$4.1249 \\
MIS-MoCov3     & ViT-B &
0.5585$\pm$0.0596 & 59.5376$\pm$12.6458 &
0.1401$\pm$0.0392 & 85.1487$\pm$14.8940 &
0.3919$\pm$0.0333 & 41.6219$\pm$2.0643 \\
MSN            & ViT-B &
0.5662$\pm$0.0533 & 60.3766$\pm$7.4621  &
0.1357$\pm$0.0454 & 91.6858$\pm$21.5929 &
0.3657$\pm$0.0301 & 41.3406$\pm$3.3496 \\
MIS-DINO       & ViT-B &
0.6178$\pm$0.0418 & 47.5757$\pm$5.9662  &
0.1536$\pm$0.0441 & 88.5657$\pm$22.8012 &
0.4268$\pm$0.0491 & 32.5568$\pm$3.0483 \\
MIS-DINOv2     & ViT-B &
0.6400$\pm$0.0687 & 45.6413$\pm$10.6015 &
\textbf{0.1836$\pm$0.0512} & 81.2450$\pm$21.5940 &
0.4789$\pm$0.0363 & 27.6889$\pm$2.2124 \\
MIS-DINOv2     & ViT-L &
\textbf{0.6649$\pm$0.0449} & \textbf{44.5351$\pm$11.2084} &
0.1828$\pm$0.0522 & \textbf{80.3261$\pm$20.3708} &
\textbf{0.5241$\pm$0.0409} & \textbf{21.2048$\pm$1.6209} \\
\midrule
ImageNet       & ViT-B        &
0.5389$\pm$0.0839 & 60.4903$\pm$9.5964  &
0.1434$\pm$0.0418 & 106.2664$\pm$21.1179 &
0.3655$\pm$0.0393 & 54.3416$\pm$2.3514 \\
EndoFM         & ViT-B        &
0.5119$\pm$0.0895 & 67.7710$\pm$15.2661 &
0.1245$\pm$0.0389 & 114.2099$\pm$17.5028 &
0.3063$\pm$0.0060 & 62.3997$\pm$9.0828 \\
GSViT          & EfficientViT &
0.4048$\pm$0.0410 & 75.8909$\pm$4.2465  &
0.0752$\pm$0.0493 & 106.5556$\pm$20.3083 &
0.2474$\pm$0.0160 & 66.5282$\pm$9.9651 \\
EndoViT        & ViT-B        &
0.4977$\pm$0.0492 & 67.4297$\pm$9.6738  &
0.0898$\pm$0.0559 & 113.5676$\pm$20.5245 &
0.3012$\pm$0.0419 & 51.4144$\pm$4.9574 \\
PeskaVLP       & ResNet50     &
0.5579$\pm$0.0640 & 63.7299$\pm$9.7248  &
0.1279$\pm$0.0482 & \textbf{82.7714$\pm$11.8551} &
0.4296$\pm$0.0343 & 57.4062$\pm$3.4523 \\
SurgeNet       & CAFormer18   &
0.5643$\pm$0.0525 & 58.0862$\pm$11.8622 &
0.1190$\pm$0.0755 & 118.9908$\pm$68.4151 &
0.4557$\pm$0.0211 & 35.7468$\pm$2.8892 \\
ZEN            & ViT-B        &
\textbf{0.6314$\pm$0.0512} & \textbf{45.2514$\pm$9.8517} &
\textbf{0.1658$\pm$0.0441} & 83.1591$\pm$20.1159 &
\textbf{0.4815$\pm$0.0273} & \textbf{25.7157$\pm$1.6724} \\
\bottomrule
\end{tabular}
\end{sidewaystable}

\begin{sidewaystable}[p]
\centering
\caption{\textbf{Few-shot evaluation across three datasets for semantic segmentation (3-shot).}
Results are reported using the Dice similarity coefficient (DSC; higher is better) and the 95th percentile Hausdorff distance (HD95; lower is better). The upper block reports results for self-supervised learning pretrained models, and the lower block reports results for existing pretrained models. Best-performing models for each metric are highlighted in bold. Values are reported as mean $\pm$ standard deviation over five independent runs.}
\label{tab:seg_3shot}
\setlength{\tabcolsep}{1.5pt} 
\begin{tabular}{llcccccc}
\toprule
\multirow{2}{*}{Method} &
\multirow{2}{*}{Backbone} &
\multicolumn{2}{c}{CholecSeg8k} &
\multicolumn{2}{c}{DSAD} &
\multicolumn{2}{c}{GraSP} \\
\cmidrule(lr){3-4} \cmidrule(lr){5-6} \cmidrule(lr){7-8}
 &  & DSC$\uparrow$ & HD95$\downarrow$ & DSC$\uparrow$ & HD95$\downarrow$ & DSC$\uparrow$ & HD95$\downarrow$ \\
\midrule
MIS-MAE        & ViT-B &
0.5786$\pm$0.0363 & 57.7067$\pm$5.6508 &
0.1351$\pm$0.0349 & 82.7762$\pm$18.7524 &
0.4141$\pm$0.0607 & 42.7034$\pm$5.4854 \\
MIS-MoCov3     & ViT-B &
0.6392$\pm$0.0309 & 45.8050$\pm$8.7912 &
0.1679$\pm$0.0265 & 75.6709$\pm$5.5137  &
0.4610$\pm$0.0305 & 36.5765$\pm$1.2037 \\
MSN            & ViT-B &
0.6489$\pm$0.0431 & 48.9809$\pm$3.7012 &
0.1711$\pm$0.0195 & 71.4510$\pm$9.4944  &
0.4687$\pm$0.0468 & 35.7465$\pm$1.6530 \\
MIS-DINO       & ViT-B &
0.6844$\pm$0.0378 & 39.1455$\pm$1.2117 &
0.1764$\pm$0.0179 & 66.1154$\pm$5.7820  &
0.5108$\pm$0.0569 & 28.2094$\pm$1.2300 \\
MIS-DINOv2     & ViT-B &
0.6956$\pm$0.0312 & 38.6578$\pm$10.2172 &
\textbf{0.2060$\pm$0.0242} & \textbf{62.4711$\pm$6.7668}  &
0.5889$\pm$0.0458 & 23.4928$\pm$0.8177 \\
MIS-DINOv2     & ViT-L &
\textbf{0.7278$\pm$0.0334} & \textbf{35.9219$\pm$7.3329} &
0.1995$\pm$0.0295 & 69.5943$\pm$12.3056 &
\textbf{0.6434$\pm$0.0710} & \textbf{18.4461$\pm$0.9156} \\
\midrule
ImageNet       & ViT-B        & 0.6206$\pm$0.0459 & 53.5868$\pm$7.4735 & 0.1663$\pm$0.0227 & 97.7866$\pm$12.2744 & 0.4378$\pm$0.0280 & 49.5688$\pm$3.7807 \\
EndoFM         & ViT-B        & 0.5889$\pm$0.0282 & 57.4098$\pm$8.2401 & 0.1478$\pm$0.0228 & 97.4356$\pm$13.6616 & 0.3695$\pm$0.0429 & 56.9927$\pm$4.8450 \\
GSViT          & EfficientViT & 0.4360$\pm$0.0420 & 71.0993$\pm$2.2085 & 0.1079$\pm$0.0296 & 80.6636$\pm$14.3993 & 0.3067$\pm$0.0292 & 58.2571$\pm$8.8046 \\
EndoViT        & ViT-B        & 0.5690$\pm$0.0370 & 58.0000$\pm$6.3007 & 0.1401$\pm$0.0349 & 93.8831$\pm$6.3572  & 0.4177$\pm$0.0490 & 39.7205$\pm$2.6880 \\
PeskaVLP       & ResNet50     & 0.6229$\pm$0.0262 & 58.2219$\pm$7.5270 & 0.1513$\pm$0.0202 & 87.2248$\pm$12.4500 & 0.5003$\pm$0.0412 & 48.0382$\pm$4.7204 \\
SurgeNet       & CAFormer18   & 0.6610$\pm$0.0261 & 49.2176$\pm$4.6224 & 0.1526$\pm$0.0579 & 72.2667$\pm$9.7763  & 0.5578$\pm$0.0536 & 31.4377$\pm$1.1079 \\
ZEN            & ViT-B        & \textbf{0.7154$\pm$0.0331} & \textbf{32.1924$\pm$7.5280} & \textbf{0.2007$\pm$0.0245} & \textbf{65.2202$\pm$10.1382} & \textbf{0.6047$\pm$0.0932} & \textbf{20.9223$\pm$1.6400} \\
\bottomrule
\end{tabular}
\end{sidewaystable}

\begin{sidewaystable}[p]
\centering
\caption{\textbf{Few-shot evaluation across three datasets for semantic segmentation (4-shot).}
Results are reported using the Dice similarity coefficient (DSC; higher is better) and the 95th percentile Hausdorff distance (HD95; lower is better). The upper block reports results for self-supervised learning pretrained models, and the lower block reports results for existing pretrained models. Best-performing models for each metric are highlighted in bold. Values are reported as mean $\pm$ standard deviation over five independent runs.}
\label{tab:seg_4shot}
\setlength{\tabcolsep}{1.5pt} 
\begin{tabular}{llcccccc}
\toprule
\multirow{2}{*}{Method} &
\multirow{2}{*}{Backbone} &
\multicolumn{2}{c}{CholecSeg8k} &
\multicolumn{2}{c}{DSAD} &
\multicolumn{2}{c}{GraSP} \\
\cmidrule(lr){3-4} \cmidrule(lr){5-6} \cmidrule(lr){7-8}
 &  & DSC$\uparrow$ & HD95$\downarrow$ & DSC$\uparrow$ & HD95$\downarrow$ & DSC$\uparrow$ & HD95$\downarrow$ \\
\midrule
MIS-MAE        & ViT-B &
0.5929$\pm$0.0288 & 53.3101$\pm$5.8728 &
0.1475$\pm$0.0217 & 81.8138$\pm$13.1047 &
0.4349$\pm$0.0452 & 38.7078$\pm$2.7976 \\
MIS-MoCov3     & ViT-B &
0.6677$\pm$0.0253 & 45.0314$\pm$1.5798 &
0.1811$\pm$0.0117 & 72.0254$\pm$4.5301  &
0.5192$\pm$0.0315 & 33.5958$\pm$3.8141 \\
MSN            & ViT-B &
0.6734$\pm$0.0381 & 45.0450$\pm$4.7207 &
0.1767$\pm$0.0119 & 68.7937$\pm$9.3974  &
0.5185$\pm$0.0270 & 31.3206$\pm$0.5738 \\
MIS-DINO       & ViT-B &
0.7066$\pm$0.0425 & 37.2462$\pm$4.8594 &
0.1889$\pm$0.0161 & 64.1794$\pm$7.9686  &
0.5799$\pm$0.0177 & 26.1727$\pm$1.3549 \\
MIS-DINOv2     & ViT-B &
0.7278$\pm$0.0498 & 34.8388$\pm$3.9918 &
0.2156$\pm$0.0292 & 61.8369$\pm$8.4555  &
0.6512$\pm$0.0199 & 21.7428$\pm$2.0638 \\
MIS-DINOv2     & ViT-L &
\textbf{0.7424$\pm$0.0326} & \textbf{30.4469$\pm$3.5235} &
\textbf{0.2226$\pm$0.0225} & \textbf{60.7209$\pm$4.3555} &
\textbf{0.6929$\pm$0.0326} & \textbf{17.6548$\pm$0.6112} \\
\midrule
Supervised     & ViT-B        & 0.6333$\pm$0.0454 & 51.6669$\pm$5.9001 & 0.1759$\pm$0.0182 & 92.2557$\pm$15.0001 & 0.4734$\pm$0.0120 & 43.4316$\pm$3.5349 \\
EndoFM         & ViT-B        & 0.6026$\pm$0.0173 & 55.2072$\pm$3.7543 & 0.1562$\pm$0.0144 & 90.6198$\pm$16.5918 & 0.4302$\pm$0.0113 & 50.9416$\pm$3.5520 \\
GSViT          & EfficientViT & 0.4778$\pm$0.0346 & 65.6504$\pm$4.8604 & 0.1117$\pm$0.0367 & 97.8121$\pm$36.7426 & 0.3406$\pm$0.0228 & 53.0146$\pm$4.4394 \\
EndoViT        & ViT-B        & 0.5958$\pm$0.0299 & 52.1638$\pm$6.3667 & 0.1530$\pm$0.0161 & 78.6007$\pm$6.8572  & 0.4668$\pm$0.0266 & 35.9429$\pm$1.7238 \\
PeskaVLP       & ResNet50     & 0.6582$\pm$0.0249 & 52.7978$\pm$4.5837 & 0.1764$\pm$0.0056 & 83.3994$\pm$3.4919  & 0.5339$\pm$0.0107 & 43.6190$\pm$3.8169 \\
SurgeNet       & CAFormer18   & 0.6607$\pm$0.0252 & 45.7105$\pm$5.4998 & 0.2057$\pm$0.0271 & 63.4717$\pm$3.7046  & 0.6248$\pm$0.0199 & 27.5561$\pm$2.0834 \\
ZEN            & ViT-B        & \textbf{0.7288$\pm$0.0365} & \textbf{30.3896$\pm$3.2979} & \textbf{0.2129$\pm$0.0186} & \textbf{59.2688$\pm$5.3527} & \textbf{0.6836$\pm$0.0245} & \textbf{18.3150$\pm$0.4946} \\
\bottomrule
\end{tabular}
\end{sidewaystable}

\begin{sidewaystable}[p]
\centering
\caption{\textbf{Few-shot evaluation across three datasets for semantic segmentation (5-shot).}
Results are reported using the Dice similarity coefficient (DSC; higher is better) and the 95th percentile Hausdorff distance (HD95; lower is better). The upper block reports results for self-supervised learning pretrained models, and the lower block reports results for existing pretrained models. Best-performing models for each metric are highlighted in bold. Values are reported as mean $\pm$ standard deviation over five independent runs.}
\label{tab:seg_5shot}
\setlength{\tabcolsep}{1.5pt} 
\begin{tabular}{llcccccc}
\toprule
\multirow{2}{*}{Method} & 
\multirow{2}{*}{Backbone} &
\multicolumn{2}{c}{CholecSeg8k} &
\multicolumn{2}{c}{DSAD} &
\multicolumn{2}{c}{GraSP} \\
\cmidrule(lr){3-4} \cmidrule(lr){5-6} \cmidrule(lr){7-8}
 &  & DSC$\uparrow$ & HD95$\downarrow$ & DSC$\uparrow$ & HD95$\downarrow$ & DSC$\uparrow$ & HD95$\downarrow$ \\
\midrule
MIS-MAE        & ViT-B &
0.6052$\pm$0.0248 & 54.7292$\pm$5.0447 &
0.1569$\pm$0.0089 & 74.8170$\pm$7.4783 &
0.4976$\pm$0.0167 & 35.7042$\pm$0.9075 \\
MIS-MoCov3     & ViT-B &
0.7025$\pm$0.0186 & 41.8514$\pm$9.7523 &
0.1762$\pm$0.0061 & 69.8475$\pm$8.5838 &
0.5611$\pm$0.0149 & 30.7770$\pm$1.7680 \\
MSN            & ViT-B &
0.6874$\pm$0.0262 & 42.6891$\pm$5.1891 &
0.1886$\pm$0.0089 & 66.0214$\pm$5.9567 &
0.5659$\pm$0.0226 & 29.8681$\pm$0.8912 \\
MIS-DINO       & ViT-B &
0.7333$\pm$0.0124 & 35.1733$\pm$4.2421 &
0.1991$\pm$0.0182 & 64.1841$\pm$8.9671 &
0.6215$\pm$0.0250 & 24.6398$\pm$1.2782 \\
MIS-DINOv2     & ViT-B &
0.7561$\pm$0.0128 & 30.4994$\pm$3.9548 &
0.2233$\pm$0.0191 & 57.6411$\pm$6.6478 &
0.6721$\pm$0.0175 & 20.7291$\pm$0.8776 \\
MIS-DINOv2     & ViT-L &
\textbf{0.7760$\pm$0.0176} & \textbf{28.0174$\pm$3.0523} &
\textbf{0.2406$\pm$0.0164} & \textbf{51.6475$\pm$3.3627} &
\textbf{0.7286$\pm$0.0105} & \textbf{16.2606$\pm$0.8184} \\
\midrule
Supervised     & ViT-B        & 0.6596$\pm$0.0344 & 48.8830$\pm$7.5070 & 0.1811$\pm$0.0135 & 76.4114$\pm$13.2860 & 0.4953$\pm$0.0231 & 40.1513$\pm$2.5687 \\
EndoFM         & ViT-B        & 0.6453$\pm$0.0246 & 52.0217$\pm$7.6721 & 0.1651$\pm$0.0126 & 84.5214$\pm$15.0283 & 0.4685$\pm$0.0275 & 46.5001$\pm$3.8578 \\
GSViT          & EfficientViT & 0.5297$\pm$0.0080 & 61.3753$\pm$3.3816 & 0.1263$\pm$0.0162 & 76.0734$\pm$14.3654 & 0.3921$\pm$0.0252 & 51.4884$\pm$2.9585 \\
EndoViT        & ViT-B        & 0.6464$\pm$0.0320 & 47.2788$\pm$4.1942 & 0.1531$\pm$0.0132 & 81.5460$\pm$9.4837 & 0.5124$\pm$0.0335 & 36.2277$\pm$2.2468 \\
PeskaVLP       & ResNet50     & 0.6621$\pm$0.0375 & 52.4071$\pm$5.3815 & 0.1795$\pm$0.0183 & 76.4342$\pm$9.7092 & 0.5915$\pm$0.0205 & 37.9177$\pm$1.6946 \\
SurgeNet       & CAFormer18   & 0.6872$\pm$0.0134 & 41.8466$\pm$4.7426 & 0.2137$\pm$0.0182 & 67.4553$\pm$3.9403 & 0.6594$\pm$0.0218 & 26.2174$\pm$1.1220 \\
ZEN            & ViT-B        & \textbf{0.7624$\pm$0.0154} & \textbf{27.7687$\pm$4.2435} & \textbf{0.2138$\pm$0.0227} & \textbf{59.6110$\pm$3.2988} & \textbf{0.7229$\pm$0.0106} & \textbf{17.3003$\pm$0.7612} \\
\bottomrule
\end{tabular}
\end{sidewaystable}

\begin{table}[t]
\caption{\textbf{Instance segmentation performance on the GraSP dataset.} Results are reported in terms of bounding box (BBox) and mask mAP@[0.5:0.95] under frozen and fine-tuned backbone settings. The upper block reports results for self-supervised learning pretrained models, and the lower block reports results for existing pretrained models. Best-performing models for each metric are highlighted in bold. Values are reported as mean $\pm$ standard deviation over five independent runs.}
\label{tab:grasp-instance}
\centering
\setlength{\tabcolsep}{3pt}
\begin{tabular}{lccccc}
\toprule
\multirow{2}{*}{Method} & \multirow{2}{*}{Backbone} & \multicolumn{2}{c}{Frozen backbone} & \multicolumn{2}{c}{Fine-tuned backbone} \\
\cmidrule(lr){3-4}\cmidrule(lr){5-6}
& & BBox & Mask & BBox & Mask \\
\midrule
MIS-MAE    & ViT-B        & $0.5008\pm0.0165$ & $0.4613\pm0.0119$ & $0.5753\pm0.0097$ & $0.5204\pm0.0067$ \\
MIS-MoCoV3 & ViT-B        & $0.4843\pm0.0074$ & $0.4486\pm0.0093$ & $0.5438\pm0.0058$ & $0.5003\pm0.0090$ \\
MIS-MSN    & ViT-B        & $0.5276\pm0.0096$ & $0.4717\pm0.0076$ & $0.5582\pm0.0053$ & $0.5167\pm0.0035$ \\
MIS-DINO   & ViT-B        & $0.5478\pm0.0120$ & $0.4952\pm0.0088$ & $0.5700\pm0.0103$ & $0.5231\pm0.0075$ \\
MIS-DINOv2 & ViT-B        & $0.5504\pm0.0076$ & $0.5001\pm0.0106$ & $0.5936\pm0.0069$ & $0.5282\pm0.0062$ \\
MIS-DINOv2 & ViT-L        & $\mathbf{0.5746\pm0.0071}$ & $\mathbf{0.5263\pm0.0077}$ & $\mathbf{0.6088\pm0.0048}$ & $\mathbf{0.5432\pm0.0053}$ \\
\midrule
ImageNet   & ViT-B        & $0.4862\pm0.0037$ & $0.4561\pm0.0031$ & $0.5365\pm0.0100$ & $0.4886\pm0.0084$ \\
EndoFM     & ViT-B        & $0.4523\pm0.0146$ & $0.4323\pm0.0072$ & $0.5409\pm0.0061$ & $0.4984\pm0.0054$ \\
GSViT      & EfficientViT & $0.3513\pm0.0258$ & $0.3436\pm0.0268$ & $0.3938\pm0.0234$ & $0.3778\pm0.0197$ \\
EndoViT    & ViT-B        & $0.5120\pm0.0116$ & $0.4663\pm0.0123$ & $0.5723\pm0.0045$ & $0.5192\pm0.0055$ \\
PeskaVLP   & ResNet50     & $0.4857\pm0.0270$ & $0.4731\pm0.0227$ & $0.5789\pm0.0058$ & $0.5241\pm0.0034$ \\
SurgeNet   & CAFormer18   & $0.5400\pm0.0101$ & $0.5301\pm0.0068$ & $0.5998\pm0.0092$ & $0.5596\pm0.0111$ \\
ZEN        & ViT-B        & $\mathbf{0.5841\pm0.0071}$ & $\mathbf{0.5317\pm0.0037}$ & $\mathbf{0.6250\pm0.0054}$ & $\mathbf{0.5597\pm0.0074}$ \\
\bottomrule
\end{tabular}
\end{table}

\begin{table}[t]
\caption{\textbf{Depth estimation performance on the SCARED dataset using fine-tuned backbones.} Performance is evaluated using Abs Rel, Sq Rel, RMSE, RMSE log (lower is better), and $\delta$ (higher is better). The upper block reports results for self-supervised learning pretrained models, and the lower block reports results for existing pretrained models. Best-performing models for each metric are highlighted in bold. Values are reported as mean $\pm$ standard deviation over five independent runs.}
\label{tab:depth-scared-finetune}
\centering
\setlength{\tabcolsep}{1.0pt}
\begin{tabular}{lcccccc}
\toprule
Method & Backbone & Abs Rel $\downarrow$ & Sq Rel $\downarrow$ & RMSE $\downarrow$ & RMSE log $\downarrow$ & $\delta \uparrow$ \\
\midrule
MIS-MAE    & ViT-B & $0.1416\pm0.0117$ & $1.9033\pm0.3600$ & $8.8798\pm0.7592$ & $0.1544\pm0.0125$ & $0.8163\pm0.0370$ \\
MIS-MoCoV3 & ViT-B & $\mathbf{0.1326\pm0.0120}$ & $\mathbf{1.6049\pm0.2371}$ & $\mathbf{8.2549\pm0.4240}$ & $\mathbf{0.1440\pm0.0078}$ & $0.8325\pm0.0162$ \\
MIS-MSN    & ViT-B & $0.1333\pm0.0088$ & $1.6449\pm0.2098$ & $8.3485\pm0.5328$ & $0.1451\pm0.0095$ & $\mathbf{0.8407\pm0.0231}$ \\
MIS-DINO   & ViT-B & $0.1361\pm0.0071$ & $1.6881\pm0.1999$ & $8.5661\pm0.2402$ & $0.1492\pm0.0039$ & $0.8271\pm0.0095$ \\
MIS-DINOv2 & ViT-B & $0.1385\pm0.0046$ & $1.7175\pm0.0622$ & $8.4867\pm0.3943$ & $0.1491\pm0.0073$ & $0.8193\pm0.0123$ \\
MIS-DINOv2 & ViT-L & $0.1417\pm0.0040$ & $1.7527\pm0.1226$ & $8.7620\pm0.2505$ & $0.1529\pm0.0040$ & $0.8214\pm0.0143$ \\
\midrule
ImageNet & ViT-B & $0.1364\pm0.0155$ & $1.7377\pm0.3488$ & $8.5183\pm0.6496$ & $0.1480\pm0.0122$ & $0.8299\pm0.0317$ \\
EndoFM & ViT-B & $0.1365\pm0.0145$ & $1.7387\pm0.3745$ & $8.5228\pm0.5233$ & $0.1491\pm0.0090$ & $0.8278\pm0.0309$ \\
GSViT & EfficientViT & $0.2246\pm0.0228$ & $4.3211\pm0.7802$ & $13.3113\pm0.8691$ & $0.2337\pm0.0154$ & $0.6281\pm0.0362$ \\
EndoViT & ViT-B & $0.1477\pm0.0146$ & $2.1513\pm0.5365$ & $9.0645\pm0.6363$ & $0.1574\pm0.0097$ & $0.8142\pm0.0244$ \\
PeskaVLP & ResNet50 & $0.1545\pm0.0125$ & $2.1260\pm0.3175$ & $9.9993\pm0.7286$ & $0.1748\pm0.0133$ & $0.7835\pm0.0423$ \\
SurgeNet & CAFormer18 & $0.1329\pm0.0111$ & $\mathbf{1.5100\pm0.2089}$ & $8.1794\pm0.3829$ & $0.1438\pm0.0074$ & $0.8473\pm0.0322$ \\
ZEN & ViT-B & $\mathbf{0.1306\pm0.0128}$ & $1.5402\pm0.2908$ & $\mathbf{7.9178\pm0.4724}$ & $\mathbf{0.1386\pm0.0097}$ & $\mathbf{0.8592\pm0.0232}$ \\
\bottomrule
\end{tabular}
\end{table}

\begin{table}[t]
\caption{\textbf{Depth estimation performance on the Hamlyn dataset using fine-tuned backbones.} Performance is evaluated using Abs Rel, Sq Rel, RMSE, RMSE log (lower is better), and $\delta$ (higher is better). The upper block reports results for self-supervised learning pretrained models, and the lower block reports results for existing pretrained models. Best-performing models for each metric are highlighted in bold. Values are reported as mean $\pm$ standard deviation over five independent runs.}
\label{tab:depth-hamlyn-finetune}
\centering
\setlength{\tabcolsep}{0.5pt}
\begin{tabular}{lcccccc}
\toprule
Method & Backbone & Abs Rel $\downarrow$ & Sq Rel $\downarrow$ & RMSE $\downarrow$ & RMSE log $\downarrow$ & $\delta \uparrow$ \\
\midrule
MIS-MAE    & ViT-B        & $0.1908\pm0.0116$ & $3.9669\pm0.3956$ & $14.6177\pm1.1492$ & $0.2206\pm0.0177$ & $0.6661\pm0.0595$ \\
MIS-MoCoV3 & ViT-B        & $0.1745\pm0.0048$ & $\mathbf{3.4721\pm0.2579}$ & $\mathbf{14.1841\pm0.5051}$ & $\mathbf{0.2127\pm0.0074}$ & $\mathbf{0.6921\pm0.0141}$ \\
MIS-MSN    & ViT-B        & $0.1727\pm0.0070$ & $3.6208\pm0.2635$ & $14.4188\pm0.4736$ & $0.2151\pm0.0079$ & $0.6836\pm0.0193$ \\
MIS-DINO   & ViT-B        & $0.1802\pm0.0090$ & $3.8866\pm0.4094$ & $14.9283\pm0.6281$ & $0.2257\pm0.0111$ & $0.6449\pm0.0366$ \\
MIS-DINOv2 & ViT-B        & $0.1714\pm0.0133$ & $3.5428\pm0.5843$ & $14.2567\pm1.0107$ & $0.2134\pm0.0172$ & $0.6858\pm0.0423$ \\
MIS-DINOv2 & ViT-L        & $\mathbf{0.1614\pm0.0073}$ & $3.1527\pm0.2854$ & $13.1128\pm0.6479$ & $0.1953\pm0.0103$ & $0.7304\pm0.0264$ \\
\midrule
ImageNet   & ViT-B        & $0.1864\pm0.0036$ & $4.1933\pm0.2394$ & $15.0118\pm0.4344$ & $0.2236\pm0.0065$ & $0.6848\pm0.0099$ \\
EndoFM     & ViT-B        & $0.1817\pm0.0041$ & $4.0425\pm0.0905$ & $14.8925\pm0.1973$ & $0.2213\pm0.0031$ & $0.6884\pm0.0113$ \\
GSViT      & EfficientViT & $0.2407\pm0.0287$ & $6.0054\pm1.2287$ & $17.0957\pm1.3735$ & $0.2585\pm0.0192$ & $0.5751\pm0.0378$ \\
EndoViT    & ViT-B        & $0.1906\pm0.0031$ & $4.2328\pm0.2870$ & $15.4385\pm0.5967$ & $0.2323\pm0.0100$ & $0.6519\pm0.0306$ \\
PeskaVLP   & ResNet50     & $0.1812\pm0.0069$ & $3.9584\pm0.3029$ & $14.8482\pm0.5933$ & $0.2257\pm0.0098$ & $0.6583\pm0.0345$ \\
SurgeNet   & CAFormer18   & $0.1715\pm0.0067$ & $3.6143\pm0.2836$ & $14.1965\pm0.5329$ & $0.2125\pm0.0087$ & $0.6836\pm0.0374$ \\
ZEN        & ViT-B        & $\mathbf{0.1554\pm0.0098}$ & $\mathbf{2.8806\pm0.3651}$ & $\mathbf{12.5496\pm0.7705}$ & $\mathbf{0.1870\pm0.0122}$ & $\mathbf{0.7351\pm0.0496}$ \\
\bottomrule
\end{tabular}
\end{table}

\begin{table}[t]
\caption{\textbf{Depth estimation performance on the SCARED dataset using frozen backbones.} Performance is evaluated using Abs Rel, Sq Rel, RMSE, RMSE log (lower is better), and $\delta$ (higher is better). The upper block reports results for self-supervised learning pretrained models, and the lower block reports results for existing pretrained models. Best-performing models for each metric are highlighted in bold. Values are reported as mean $\pm$ standard deviation over five independent runs.}
\label{tab:depth-scared-frozen}
\centering
\setlength{\tabcolsep}{1.0pt} 
\begin{tabular}{lcccccc}
\toprule
Method & Backbone & Abs Rel $\downarrow$ & Sq Rel $\downarrow$ & RMSE $\downarrow$ & RMSE log $\downarrow$ & $\delta \uparrow$ \\
\midrule
MIS-MAE & ViT-B & $0.2117\pm0.0201$ & $3.9499\pm0.5276$ & $12.3165\pm0.7439$ & $0.2140\pm0.0155$ & $0.6661\pm0.0354$ \\
MIS-MoCoV3 & ViT-B & $0.1504\pm0.0117$ & $2.0385\pm0.2570$ & $9.2167\pm0.4940$ & $0.1629\pm0.0088$ & $0.7803\pm0.0234$ \\
MIS-MSN & ViT-B & $0.1504\pm0.0066$ & $2.0154\pm0.0918$ & $9.4050\pm0.2375$ & $0.1675\pm0.0067$ & $0.7812\pm0.0245$ \\
MIS-DINO & ViT-B & $0.1557\pm0.0068$ & $2.2076\pm0.2312$ & $9.5348\pm0.3725$ & $0.1675\pm0.0059$ & $0.7798\pm0.0178$ \\
MIS-DINOv2 & ViT-B & $0.1472\pm0.0068$ & $2.0371\pm0.1705$ & $8.7419\pm0.4077$ & $0.1567\pm0.0073$ & $0.7814\pm0.0264$ \\
MIS-DINOv2 & ViT-L & $\mathbf{0.1340\pm0.0081}$ & $\mathbf{1.6110\pm0.2032}$ & $\mathbf{8.2607\pm0.3470}$ & $\mathbf{0.1458\pm0.0059}$ & $\mathbf{0.8091\pm0.0165}$ \\
\midrule
ImageNet & ViT-B & $0.1595\pm0.0135$ & $2.3831\pm0.2941$ & $9.8408\pm0.6187$ & $0.1712\pm0.0107$ & $0.7671\pm0.0276$ \\
EndoFM & ViT-B & $0.1541\pm0.0134$ & $2.1728\pm0.2640$ & $9.5215\pm0.7185$ & $0.1712\pm0.0146$ & $0.7631\pm0.0479$ \\
GSViT & EfficientViT & $0.2646\pm0.0169$ & $5.5267\pm0.5563$ & $15.3095\pm0.7114$ & $0.2674\pm0.0118$ & $0.5511\pm0.0291$ \\
EndoViT & ViT-B & $0.2034\pm0.0111$ & $3.6276\pm0.2977$ & $12.4025\pm0.4216$ & $0.2160\pm0.0096$ & $0.6468\pm0.0300$ \\
PeskaVLP & ResNet50 & $0.1984\pm0.0097$ & $3.2513\pm0.2617$ & $12.4412\pm0.3488$ & $0.2188\pm0.0061$ & $0.6728\pm0.0207$ \\
SurgeNet & CAFormer18 & $0.1701\pm0.0084$ & $2.4920\pm0.2443$ & $10.7763\pm0.2705$ & $0.1884\pm0.0045$ & $0.7268\pm0.0138$ \\
ZEN & ViT-B & $\mathbf{0.1309\pm0.0069}$ & $\mathbf{1.6220\pm0.2034}$ & $\mathbf{8.1781\pm0.3220}$ & $\mathbf{0.1431\pm0.0055}$ & $\mathbf{0.8242\pm0.0096}$ \\
\bottomrule
\end{tabular}
\end{table}

\begin{table}[t]
\caption{\textbf{Depth estimation performance on the Hamlyn dataset using frozen backbones.} Performance is evaluated using Abs Rel, Sq Rel, RMSE, RMSE log (lower is better), and $\delta$ (higher is better). The upper block reports results for self-supervised learning pretrained models, and the lower block reports results for existing pretrained models. Best-performing models for each metric are highlighted in bold. Values are reported as mean $\pm$ standard deviation over five independent runs.}
\label{tab:depth-hamlyn-frozen}
\centering
\setlength{\tabcolsep}{1.0pt}
\begin{tabular}{lcccccc}
\toprule
Method & Backbone & Abs Rel $\downarrow$ & Sq Rel $\downarrow$ & RMSE $\downarrow$ & RMSE log $\downarrow$ & $\delta \uparrow$ \\
\midrule
MIS-MAE    & ViT-B        & $0.1897\pm0.0133$ & $4.2759\pm0.6031$ & $14.7664\pm0.6432$ & $0.2213\pm0.0099$ & $0.6902\pm0.0208$ \\
MIS-MoCoV3 & ViT-B        & $0.1753\pm0.0073$ & $3.4550\pm0.2221$ & $14.1648\pm0.4104$ & $0.2131\pm0.0069$ & $0.6884\pm0.0285$ \\
MIS-MSN    & ViT-B        & $0.1689\pm0.0062$ & $3.2287\pm0.3055$ & $13.7272\pm0.6784$ & $0.2060\pm0.0093$ & $0.7086\pm0.0229$ \\
MIS-DINO   & ViT-B        & $0.1691\pm0.0087$ & $3.2971\pm0.4387$ & $13.9077\pm0.8205$ & $0.2096\pm0.0122$ & $0.7003\pm0.0248$ \\
MIS-DINOv2 & ViT-B        & $0.1680\pm0.0131$ & $3.3986\pm0.5651$ & $13.8990\pm1.2661$ & $0.2075\pm0.0200$ & $0.7102\pm0.0476$ \\
MIS-DINOv2 & ViT-L        & $\mathbf{0.1579\pm0.0128}$ & $\mathbf{3.0895\pm0.4933}$ & $\mathbf{12.9146\pm1.0656}$ & $\mathbf{0.1923\pm0.0164}$ & $\mathbf{0.7373\pm0.0327}$ \\
\midrule
ImageNet   & ViT-B        & $0.1831\pm0.0060$ & $4.1421\pm0.0952$ & $14.9893\pm0.2588$ & $0.2243\pm0.0065$ & $0.6684\pm0.0232$ \\
EndoFM     & ViT-B        & $0.1830\pm0.0023$ & $3.9253\pm0.0524$ & $14.3977\pm0.1409$ & $0.2153\pm0.0022$ & $0.6968\pm0.0120$ \\
GSViT      & EfficientViT & $0.2470\pm0.0170$ & $6.2404\pm0.7032$ & $17.8172\pm0.7431$ & $0.2670\pm0.0091$ & $0.5720\pm0.0221$ \\
EndoViT    & ViT-B        & $0.2177\pm0.0070$ & $5.2286\pm0.2621$ & $17.2679\pm0.5594$ & $0.2695\pm0.0108$ & $0.5444\pm0.0263$ \\
PeskaVLP   & ResNet50     & $0.1886\pm0.0078$ & $4.2342\pm0.3281$ & $14.9181\pm0.5209$ & $0.2221\pm0.0076$ & $0.6901\pm0.0125$ \\
SurgeNet   & CAFormer18   & $0.1929\pm0.0055$ & $4.1303\pm0.2559$ & $15.5793\pm0.4399$ & $0.2381\pm0.0076$ & $0.6144\pm0.0164$ \\
ZEN        & ViT-B        & $\mathbf{0.1621\pm0.0100}$ & $\mathbf{3.2286\pm0.3645}$ & $\mathbf{13.2018\pm0.7469}$ & $\mathbf{0.1971\pm0.0125}$ & $\mathbf{0.7273\pm0.0345}$ \\
\bottomrule
\end{tabular}
\end{table}

\begin{sidewaystable}[p]
\centering
\caption{\textbf{Closed-ended VQA performance on PitVQA.} F-score, balanced accuracy, and recall are reported for frozen and fine-tuned backbones. The upper block shows self-supervised learning pretrained models and the lower block shows existing pretrained models. Best-performing models are highlighted in bold. Values are reported as mean $\pm$ standard deviation over five independent runs.}
\label{tab:vqa-pitvqa-classification}
\setlength{\tabcolsep}{4pt} 
\begin{tabular}{lccccccc}
\toprule
\multirow{2}{*}{Method} & \multirow{2}{*}{Backbone} & \multicolumn{3}{c}{Frozen backbone} & \multicolumn{3}{c}{Fine-tuning backbone} \\
\cmidrule(lr){3-5}\cmidrule(lr){6-8}
& & F1 score $\uparrow$ & B.\,Accuracy $\uparrow$ & Recall $\uparrow$ & F1 score $\uparrow$ & B.\,Accuracy $\uparrow$ & Recall $\uparrow$ \\
\midrule
MIS-MAE    & ViT-B        & $0.4291\pm0.0120$ & $0.4619\pm0.0137$ & $0.4852\pm0.0138$ & $0.5465\pm0.0197$ & $0.5549\pm0.0221$ & $0.5684\pm0.0218$ \\
MIS-MoCoV3 & ViT-B        & $0.4783\pm0.0237$ & $0.4923\pm0.0042$ & $0.5127\pm0.0052$ & $0.5381\pm0.0121$ & $0.5425\pm0.0192$ & $0.5563\pm0.0227$ \\
MIS-MSN    & ViT-B        & $0.2708\pm0.0190$ & $0.2848\pm0.0194$ & $0.2943\pm0.0273$ & $0.4047\pm0.0175$ & $0.4221\pm0.0041$ & $0.4357\pm0.0130$ \\
MIS-DINO   & ViT-B        & $0.5281\pm0.0094$ & $0.5290\pm0.0158$ & $0.5447\pm0.0196$ & $0.5523\pm0.0055$ & $0.5568\pm0.0184$ & $0.5715\pm0.0230$ \\
MIS-DINOv2 & ViT-B        & $0.5447\pm0.0118$ & $0.5446\pm0.0085$ & $0.5585\pm0.0103$ & $0.5658\pm0.0128$ & $0.5668\pm0.0107$ & $0.5785\pm0.0129$ \\
MIS-DINOv2 & ViT-L        & $\mathbf{0.5610\pm0.0174}$ & $\mathbf{0.5680\pm0.0240}$ & $\mathbf{0.5851\pm0.0257}$ & $\mathbf{0.5920\pm0.0126}$ & $\mathbf{0.5928\pm0.0127}$ & $\mathbf{0.6038\pm0.0133}$ \\
\midrule
ImageNet   & ViT-B        & $0.4770\pm0.0220$ & $0.4952\pm0.0081$ & $0.5121\pm0.0111$ & $0.5523\pm0.0138$ & $0.5617\pm0.0117$ & $0.5749\pm0.0158$ \\
EndoFM     & ViT-B        & $0.4759\pm0.0162$ & $0.4820\pm0.0178$ & $0.4995\pm0.0163$ & $0.5209\pm0.0221$ & $0.5306\pm0.0083$ & $0.5464\pm0.0088$ \\
GSViT      & EfficientViT & $0.2574\pm0.0338$ & $0.2737\pm0.0273$ & $0.2957\pm0.0284$ & $0.2490\pm0.0112$ & $0.2681\pm0.0094$ & $0.2903\pm0.0138$ \\
EndoViT    & ViT-B        & $0.4163\pm0.0212$ & $0.4426\pm0.0099$ & $0.4650\pm0.0097$ & $0.5078\pm0.0187$ & $0.5221\pm0.0228$ & $0.5413\pm0.0204$ \\
PeskaVLP   & ResNet50     & $0.4438\pm0.0182$ & $0.4555\pm0.0078$ & $0.4738\pm0.0100$ & $0.5364\pm0.0127$ & $0.5452\pm0.0180$ & $0.5590\pm0.0191$ \\
SurgeNet   & CAFormer18   & $0.4708\pm0.0161$ & $0.4846\pm0.0178$ & $0.5018\pm0.0200$ & $0.5917\pm0.0235$ & $0.5984\pm0.0175$ & $0.6094\pm0.0126$ \\
ZEN        & ViT-B        & $\mathbf{0.5510\pm0.0125}$ & $\mathbf{0.5537\pm0.0048}$ & $\mathbf{0.5673\pm0.0085}$ & $\mathbf{0.6043\pm0.0140}$ & $\mathbf{0.6071\pm0.0105}$ & $\mathbf{0.6190\pm0.0119}$ \\
\bottomrule
\end{tabular}
\end{sidewaystable}

\begin{sidewaystable}[p]
\centering
\caption{\textbf{Closed-ended VQA performance on LLS48-VQA.} F-score, balanced accuracy, and recall are reported for frozen and fine-tuned backbones. The upper block shows self-supervised learning pretrained models and the lower block shows existing pretrained models. Best-performing models are highlighted in bold. Values are reported as mean $\pm$ standard deviation over five independent runs.}

\label{tab:vqa-lls48-classification}
\setlength{\tabcolsep}{4pt}
\begin{tabular}{lccccccc}
\toprule
\multirow{2}{*}{Method} & \multirow{2}{*}{Backbone} & \multicolumn{3}{c}{Frozen backbone} & \multicolumn{3}{c}{Fine-tuning backbone} \\
\cmidrule(lr){3-5}\cmidrule(lr){6-8}
& & F1 score $\uparrow$ & B.\,Accuracy $\uparrow$ & Recall $\uparrow$ & F1 score $\uparrow$ & B.\,Accuracy $\uparrow$ & Recall $\uparrow$ \\
\midrule
MIS-MAE    & ViT-B        & $0.1457\pm0.0063$ & $0.1742\pm0.0107$ & $0.2957\pm0.0240$ & $0.1934\pm0.0067$ & $0.2286\pm0.0107$ & $0.3447\pm0.0191$ \\
MIS-MoCoV3 & ViT-B        & $0.1730\pm0.0165$ & $0.1944\pm0.0228$ & $0.2584\pm0.0424$ & $0.1718\pm0.0140$ & $0.1981\pm0.0226$ & $0.2701\pm0.0353$ \\
MIS-MSN    & ViT-B        & $0.0399\pm0.0037$ & $0.0606\pm0.0047$ & $0.0606\pm0.0047$ & $0.1240\pm0.0108$ & $0.1442\pm0.0129$ & $0.1570\pm0.0198$ \\
MIS-DINO   & ViT-B        & $0.1883\pm0.0182$ & $0.2261\pm0.0236$ & $0.3640\pm0.0349$ & $0.1944\pm0.0171$ & $0.2305\pm0.0205$ & $0.3484\pm0.0252$ \\
MIS-DINOv2 & ViT-B        & $0.1923\pm0.0096$ & $0.2322\pm0.0165$ & $0.3699\pm0.0308$ & $0.2041\pm0.0127$ & $0.2459\pm0.0201$ & $0.3791\pm0.0245$ \\
MIS-DINOv2 & ViT-L        & $\mathbf{0.2130\pm0.0138}$ & $\mathbf{0.2600\pm0.0232}$ & $\mathbf{0.4027\pm0.0280}$ & $\mathbf{0.2174\pm0.0063}$ & $\mathbf{0.2664\pm0.0161}$ & $\mathbf{0.4020\pm0.0285}$ \\
\midrule
ImageNet   & ViT-B        & $0.1483\pm0.0107$ & $0.1780\pm0.0199$ & $0.3160\pm0.0407$ & $0.1595\pm0.0124$ & $0.1898\pm0.0172$ & $0.3139\pm0.0299$ \\
EndoFM     & ViT-B        & $0.1375\pm0.0100$ & $0.1680\pm0.0146$ & $0.3130\pm0.0246$ & $0.1431\pm0.0123$ & $0.1725\pm0.0182$ & $0.3098\pm0.0316$ \\
GSViT      & EfficientViT & $0.0526\pm0.0064$ & $0.0683\pm0.0060$ & $0.0738\pm0.0076$ & $0.0568\pm0.0072$ & $0.0734\pm0.0080$ & $0.0780\pm0.0082$ \\
EndoViT    & ViT-B        & $0.1366\pm0.0114$ & $0.1536\pm0.0121$ & $0.2539\pm0.0167$ & $0.1645\pm0.0104$ & $0.1904\pm0.0176$ & $0.2951\pm0.0322$ \\
PeskaVLP   & ResNet50     & $0.1528\pm0.0133$ & $0.1796\pm0.0172$ & $0.3002\pm0.0335$ & $0.1764\pm0.0138$ & $0.2085\pm0.0148$ & $0.3273\pm0.0244$ \\
SurgeNet   & CAFormer18   & $0.1646\pm0.0171$ & $0.2013\pm0.0130$ & $0.3358\pm0.0247$ & $0.2048\pm0.0199$ & $0.2396\pm0.0225$ & $0.3425\pm0.0225$ \\
ZEN        & ViT-B        & $\mathbf{0.2076\pm0.0134}$ & $\mathbf{0.2532\pm0.0209}$ & $\mathbf{0.3950\pm0.0302}$ & $\mathbf{0.2335\pm0.0190}$ & $\mathbf{0.2806\pm0.0240}$ & $\mathbf{0.4006\pm0.0346}$ \\
\bottomrule
\end{tabular}
\end{sidewaystable}

\begin{sidewaystable}[p]
\centering
\caption{\textbf{Open-ended visual question answering sentence generation performance on the LLS48-VQA dataset with fine-tuned backbones.} Performance is evaluated using BLEU-1 to BLEU-4, ROUGE-L, and METEOR (higher is better). The upper block reports results for self-supervised learning pretrained models, and the lower block reports results for existing pretrained models. Best-performing models for each metric are highlighted in bold. Values are reported as mean $\pm$ standard deviation over five independent runs.}

\label{tab:vqa-lls48-generation-finetune}
\setlength{\tabcolsep}{4pt}
\begin{tabular}{lccccccc}
\toprule
Method & Backbone & BLEU-1 $\uparrow$ & BLEU-2 $\uparrow$ & BLEU-3 $\uparrow$ & BLEU-4 $\uparrow$ & ROUGE-L $\uparrow$ & METEOR $\uparrow$ \\
\midrule
MIS-MAE    & ViT-B        & $0.5131\pm0.0108$ & $0.4212\pm0.0119$ & $0.3601\pm0.0122$ & $0.3149\pm0.0121$ & $0.4391\pm0.0123$ & $0.2616\pm0.0054$ \\
MIS-MoCoV3 & ViT-B        & $0.4980\pm0.0107$ & $0.4060\pm0.0117$ & $0.3455\pm0.0118$ & $0.3010\pm0.0117$ & $0.4268\pm0.0129$ & $0.2558\pm0.0064$ \\
MIS-MSN    & ViT-B        & $0.4926\pm0.0129$ & $0.4019\pm0.0143$ & $0.3423\pm0.0158$ & $0.2983\pm0.0167$ & $0.4197\pm0.0124$ & $0.2515\pm0.0057$ \\
MIS-DINO   & ViT-B        & $0.5116\pm0.0080$ & $0.4197\pm0.0083$ & $0.3582\pm0.0082$ & $0.3126\pm0.0080$ & $0.4384\pm0.0099$ & $0.2614\pm0.0049$ \\
MIS-DINOv2 & ViT-B        & $0.5173\pm0.0116$ & $0.4250\pm0.0123$ & $0.3629\pm0.0126$ & $0.3167\pm0.0127$ & $0.4432\pm0.0093$ & $0.2657\pm0.0038$ \\
MIS-DINOv2 & ViT-L        & $\mathbf{0.5245\pm0.0082}$ & $\mathbf{0.4336\pm0.0094}$ & $\mathbf{0.3725\pm0.0094}$ & $\mathbf{0.3270\pm0.0091}$ & $\mathbf{0.4529\pm0.0104}$ & $\mathbf{0.2727\pm0.0047}$ \\
\midrule
ImageNet   & ViT-B        & $0.5006\pm0.0037$ & $0.4066\pm0.0050$ & $0.3443\pm0.0052$ & $0.2985\pm0.0053$ & $0.4237\pm0.0088$ & $0.2505\pm0.0052$ \\
EndoFM     & ViT-B        & $0.4917\pm0.0115$ & $0.3958\pm0.0119$ & $0.3328\pm0.0122$ & $0.2869\pm0.0122$ & $0.4152\pm0.0091$ & $0.2448\pm0.0041$ \\
GSViT      & EfficientViT & $0.4167\pm0.0118$ & $0.3102\pm0.0148$ & $0.2468\pm0.0156$ & $0.2022\pm0.0155$ & $0.3363\pm0.0170$ & $0.1926\pm0.0092$ \\
EndoViT    & ViT-B        & $0.5009\pm0.0133$ & $0.4086\pm0.0136$ & $0.3476\pm0.0141$ & $0.3028\pm0.0144$ & $0.4275\pm0.0109$ & $0.2537\pm0.0053$ \\
PeskaVLP   & ResNet50     & $0.5000\pm0.0094$ & $0.4085\pm0.0108$ & $0.3479\pm0.0113$ & $0.3034\pm0.0115$ & $0.4300\pm0.0133$ & $0.2558\pm0.0066$ \\
SurgeNet   & CAFormer18   & $0.5186\pm0.0085$ & $0.4284\pm0.0089$ & $0.3678\pm0.0087$ & $0.3228\pm0.0083$ & $0.4445\pm0.0081$ & $0.2672\pm0.0034$ \\
ZEN        & ViT-B        & $\mathbf{0.5244\pm0.0126}$ & $\mathbf{0.4340\pm0.0131}$ & $\mathbf{0.3734\pm0.0129}$ & $\mathbf{0.3282\pm0.0122}$ & $\mathbf{0.4519\pm0.0119}$ & $\mathbf{0.2726\pm0.0054}$ \\
\bottomrule
\end{tabular}
\end{sidewaystable}

\begin{sidewaystable}[p]
\centering
\caption{\textbf{Open-ended visual question answering sentence generation performance on the LLS48-VQA dataset with frozen backbones.} Performance is evaluated using BLEU-1 to BLEU-4, ROUGE-L, and METEOR (higher is better). The upper block reports results for self-supervised learning pretrained models, and the lower block reports results for existing pretrained models. Best-performing models for each metric are highlighted in bold. Values are reported as mean $\pm$ standard deviation over five independent runs.}
\label{tab:vqa-lls48-generation-freeze}
\setlength{\tabcolsep}{4pt}
\begin{tabular}{lccccccc}
\toprule
Method & Backbone & BLEU-1 $\uparrow$ & BLEU-2 $\uparrow$ & BLEU-3 $\uparrow$ & BLEU-4 $\uparrow$ & ROUGE-L $\uparrow$ & METEOR $\uparrow$ \\
\midrule
MIS-MAE    & ViT-B        & $0.4873\pm0.0053$ & $0.3937\pm0.0062$ & $0.3323\pm0.0068$ & $0.2873\pm0.0072$ & $0.4125\pm0.0073$ & $0.2443\pm0.0040$ \\
MIS-MoCoV3 & ViT-B        & $0.5002\pm0.0155$ & $0.4079\pm0.0163$ & $0.3473\pm0.0167$ & $0.3027\pm0.0167$ & $0.4244\pm0.0137$ & $0.2527\pm0.0063$ \\
MIS-MSN    & ViT-B        & $0.3992\pm0.0233$ & $0.2924\pm0.0230$ & $0.2288\pm0.0236$ & $0.1842\pm0.0234$ & $0.3272\pm0.0143$ & $0.1933\pm0.0114$ \\
MIS-DINO   & ViT-B        & $0.5062\pm0.0080$ & $0.4138\pm0.0080$ & $0.3524\pm0.0080$ & $0.3071\pm0.0080$ & $0.4345\pm0.0075$ & $0.2594\pm0.0041$ \\
MIS-DINOv2 & ViT-B        & $0.5110\pm0.0103$ & $0.4167\pm0.0105$ & $0.3542\pm0.0103$ & $0.3081\pm0.0102$ & $0.4347\pm0.0099$ & $0.2596\pm0.0041$ \\
MIS-DINOv2 & ViT-L        & $\mathbf{0.5175\pm0.0100}$ & $\mathbf{0.4238\pm0.0112}$ & $\mathbf{0.3618\pm0.0114}$ & $\mathbf{0.3159\pm0.0112}$ & $\mathbf{0.4419\pm0.0127}$ & $\mathbf{0.2639\pm0.0055}$ \\
\midrule
ImageNet   & ViT-B        & $0.4846\pm0.0074$ & $0.3882\pm0.0077$ & $0.3248\pm0.0083$ & $0.2787\pm0.0088$ & $0.4066\pm0.0076$ & $0.2401\pm0.0041$ \\
EndoFM     & ViT-B        & $0.4820\pm0.0055$ & $0.3883\pm0.0078$ & $0.3267\pm0.0086$ & $0.2819\pm0.0090$ & $0.4079\pm0.0095$ & $0.2399\pm0.0086$ \\
GSViT      & EfficientViT & $0.3838\pm0.0382$ & $0.2862\pm0.0313$ & $0.2261\pm0.0269$ & $0.1843\pm0.0238$ & $0.3292\pm0.0135$ & $0.1861\pm0.0172$ \\
EndoViT    & ViT-B        & $0.4824\pm0.0103$ & $0.3857\pm0.0114$ & $0.3226\pm0.0121$ & $0.2770\pm0.0123$ & $0.4077\pm0.0092$ & $0.2403\pm0.0041$ \\
PeskaVLP   & ResNet50     & $0.4817\pm0.0142$ & $0.3869\pm0.0154$ & $0.3256\pm0.0160$ & $0.2808\pm0.0160$ & $0.4068\pm0.0130$ & $0.2417\pm0.0070$ \\
SurgeNet   & CAFormer18   & $0.5030\pm0.0129$ & $0.4097\pm0.0140$ & $0.3479\pm0.0144$ & $0.3023\pm0.0144$ & $0.4285\pm0.0111$ & $0.2550\pm0.0048$ \\
ZEN        & ViT-B        & $\mathbf{0.5148\pm0.0057}$ & $\mathbf{0.4227\pm0.0062}$ & $\mathbf{0.3615\pm0.0061}$ & $\mathbf{0.3162\pm0.0058}$ & $\mathbf{0.4402\pm0.0094}$ & $\mathbf{0.2648\pm0.0062}$ \\
\bottomrule
\end{tabular}
\end{sidewaystable}

\begin{table}[t]
\centering
\caption{\textbf{Cross-modal retrieval performance on the LLS48-VQA dataset.} Image-to-text and text-to-image retrieval performance is reported using Recall@1, Recall@5, Recall@10, and mean recall. Values indicate mean $\pm$ standard deviation computed across videos.}
\label{tab:retrieval_results}
\begin{tabular}{lcccc}
\hline
\textbf{Method} & \textbf{Recall@1} & \textbf{Recall@5} & \textbf{Recall@10} & \textbf{Mean Recall} \\
\hline
\multicolumn{5}{l}{\textbf{Image-to-Text}} \\
PeskaVLP & $\mathbf{0.0814\pm 0.0666}$ & $\mathbf{0.3116\pm 0.1666}$ & $\mathbf{0.5176\pm 0.1928}$ & $\mathbf{0.3035\pm 0.1328}$ \\
ZEN      & $0.0770 \pm 0.0638$          & $0.3053 \pm 0.1676$          & $0.5169 \pm 0.1962$          & $0.2997 \pm 0.1323$ \\
\hline
\multicolumn{5}{l}{\textbf{Text-to-Image}} \\
PeskaVLP & $0.0675 \pm 0.0559$          & $0.2760 \pm 0.1498$          & $0.4577 \pm 0.1700$          & $0.2671 \pm 0.1165$ \\
ZEN      & $\mathbf{0.0703\pm 0.0560}$ & $\mathbf{0.2793\pm 0.1521}$ & $\mathbf{0.4723\pm 0.1774}$ & $\mathbf{0.2740\pm 0.1193}$ \\
\hline
\end{tabular}
\end{table}

\begin{table}[t]
\centering
\caption{\textbf{Zero-shot surgical phase recognition performance.} Models are evaluated without task-specific training on Cholec80, MultiBypass140, and AutoLaparo datasets. Performance is reported in terms of video-level F1 score and accuracy. Values indicate mean $\pm$ standard deviation across test videos in each dataset.}
\label{tab:zero-shot-phase}
\setlength{\tabcolsep}{1.5pt} 
\begin{tabular}{lcccccc}
\toprule
\multirow{2}{*}{Method} 
& \multicolumn{2}{c}{Cholec80} 
& \multicolumn{2}{c}{MultiBypass140} 
& \multicolumn{2}{c}{AutoLaparo} \\
\cmidrule(lr){2-3}\cmidrule(lr){4-5}\cmidrule(lr){6-7}
& F1 score & Accuracy 
& F1 score & Accuracy 
& F1 score & Accuracy \\
\midrule
PeskaVLP 
& $0.3056\pm0.0567$ & $0.4272\pm0.0938$
& $0.1748\pm0.0490$ & $0.2963\pm0.0868$
& $0.2433\pm0.0608$ & $\mathbf{0.3310\pm0.0873}$ \\
ZEN 
& $\mathbf{0.3135\pm0.0546}$ & $\mathbf{0.4427\pm0.0908}$
& $\mathbf{0.1822\pm0.0517}$ & $\mathbf{0.3099\pm0.0907}$
& $\mathbf{0.2445\pm0.0628}$ & $0.3299\pm0.0911$ \\
\bottomrule
\end{tabular}
\end{table}

\begin{table}[t]
\caption{\textbf{Summary of upstream datasets used for pretraining.} Overview of surgical videos and frames across different anatomical regions and procedures.}
\label{tab:upstream-datasets}
\centering
\setlength{\tabcolsep}{6pt}
\renewcommand{\arraystretch}{1.15}
\begin{tabular}{lllll}
\toprule
Dataset & Anatomical region & Surgical procedure & Videos & Frames \\
\midrule
\multicolumn{5}{l}{\textbf{Upstream datasets}} \\
SMC & Hepatobiliary & Hepatectomy & 100 & 507{,}568 \\
hSBD-instrument &  & Cholecystectomy & 24 & 18{,}064 \\
HeiChole &  & Cholecystectomy & 30 & 54{,}199 \\
Cholec80 &  & Cholecystectomy & 40 & 84{,}629 \\
Endoscapes &  & Cholecystectomy & 201 & 55{,}809 \\
HeiCo & Colon \& Rectum & Proctocolectomy & 30 & 91{,}760 \\
       &                & Rectal resection &  & 100{,}995 \\
       &                & Sigmoidectomy &  & 71{,}457 \\
ART-Net & Pelvis & Hysterectomy & 29 & 1{,}538 \\
LapGyn4 &  & Gynecologic procedure & 500 & 30{,}682 \\
SurgicalAction160 &  & Gynecologic procedure & 59 & 759 \\
GLENDA &  & Gynecologic procedure & 400 & 25{,}682 \\
ESAD &  & Prostatectomy & 4 & 51{,}552 \\
MultiByPass140 & Stomach & Gastric bypass & 80 & 417{,}518 \\
Sisvse &  & Gastrectomy & 30 & 4{,}510 \\
SurgToolLoc2022 & Others & Porcine procedure & N/A & 745{,}196 \\
Surgical YouTube & Hepatobiliary & Cholecystectomy & 3{,}253 & 315{,}666 \\
                  &               & Hepatectomy &  & 16{,}467 \\
                  & Spleen & Splenectomy &  & 20{,}865 \\
                  & Abdominal wall & Hernia repair &  & 133{,}360 \\
                  & Esophageal & Heller myotomy &  & 81{,}226 \\
                  &             & Esophagectomy &  & 48{,}641 \\
                  & Appendix & Appendectomy &  & 54{,}403 \\
                  & Stomach & Gastrectomy &  & 856 \\
                  &          & Gastrojejunostomy &  & 10{,}077 \\
                  &          & Fundoplication &  & 11{,}358 \\
                  & Colon & Colectomy &  & 290{,}423 \\
                  &        & Hemicolectomy &  & 8{,}466 \\
                  &        & Sigmoidectomy &  & 11{,}741 \\
                  & Rectum & Rectopexy &  & 15{,}092 \\
                  & Colon \& Rectum & Colorectal and rectal cancer surgery &  & 131{,}127 \\
                  & Small bowel & Ladd’s procedure &  & 7{,}058 \\
                  & Others & Unsorted procedures &  & 897{,}666 \\
\midrule
\multicolumn{3}{r}{\textbf{Total}} & \textbf{4{,}780+} & \textbf{4{,}316{,}410} \\
\bottomrule
\end{tabular}
\end{table}

\begin{table}[t]
\centering
\caption{\textbf{Summary of baseline models.}}
\label{tab:surg_foundation_summary}
\setlength{\tabcolsep}{4.5pt}
\renewcommand{\arraystretch}{1.18}
\begin{tabular}{lccl}
\toprule
\textbf{Model} & \textbf{Backbone} & \textbf{Pretraining data} & \textbf{Training strategy} \\
\midrule
ViT        & ViT-B        & ImageNet                  & Supervised \\
GSViT      & EfficientViT & 70M frames                & Next frame prediction \\
EndoFM     & ViT-B        & 5M frames                 & Teacher--student spatiotemporal matching \\
EndoViT    & ViT-B        & 700k+ frames              & MAE \\
PeskaVLP   & ResNet50     & 26k clip--narration pairs & Video--text contrastive learning \\
SurgeNet   & CAFormer18   & 4.7M+ frames                & DINO \\
\bottomrule
\end{tabular}
\end{table}

\begin{table}[t]
\caption{\textbf{Summary of downstream datasets and tasks.} Internal and external datasets used for evaluating surgical understanding across multiple tasks.}
\label{tab:downstream-dataset}
\centering
\setlength{\tabcolsep}{6pt}
\renewcommand{\arraystretch}{1.2}
\begin{tabular}{lll}
\toprule
Dataset & Surgical procedure & Task \\
\midrule
\multicolumn{3}{l}{\textbf{Internal datasets}} \\
Cholec80 & Cholecystectomy & Surgical phase recognition \\
         &                  & Zero-shot phase recognition \\
CholecT50 & Cholecystectomy & Surgical action triplet recognition \\
CholecSeg8k & Cholecystectomy & Semantic segmentation \\
MultiByPass140 & Gastric bypass surgery & Surgical phase recognition \\
               &                         & Zero-shot phase recognition \\
LLS48 & Hepatectomy & Surgical action triplet recognition \\
LLS48-VQA & Hepatectomy & VQA (Closed- and open-ended) \\
          &                         & Cross-modal retrieval \\
\midrule
\multicolumn{3}{l}{\textbf{External datasets}} \\
AutoLaparo & Hysterectomy & Surgical phase recognition \\
           &              & Zero-shot phase recognition \\
DSAD & Rectal surgery & Semantic segmentation \\
GraSP & Radical prostatectomy & Semantic segmentation \\
      &                        & Instance segmentation \\
PitVQA & Pituitary surgery & VQA (Closed-ended) \\
SCARED & Porcine procedure & Depth estimation \\
Hamlyn & In vivo endoscopic video & Depth estimation \\
\bottomrule
\end{tabular}
\end{table}

\begin{table}[t]
\centering
\caption{\textbf{Public surgical datasets used in this study.}}
\label{tab:public_datasets}
\begin{tabular}{ll}
\hline
\textbf{Dataset} & \textbf{Link} \\
\hline
Cholec80 & \url{https://camma.unistra.fr/datasets/} \\
CholecT50 & \url{https://github.com/CAMMA-public/cholect50} \\
CholecSeg8k & \url{https://www.kaggle.com/datasets/newslab/cholecseg8k} \\
Cholec80-CVS & \url{https://github.com/ManuelRios18/CHOLEC80-CVS-PUBLIC} \\
hSBD-instrument & \url{https://hsdb-instrument.github.io/} \\
Sisvse & \url{https://sisvse.github.io/} \\
HeiChole & \url{https://www.synapse.org/Synapse:syn18824884/wiki/591922} \\
Endoscapes & \url{https://github.com/CAMMA-public/Endoscapes} \\
ART-Net & \url{https://github.com/kamruleee51/ART-Net} \\
ESAD & \url{https://saras-esad.grand-challenge.org/} \\
AutoLaparo & \url{https://autolaparo.github.io/} \\
MultiByPass140 & \url{https://github.com/CAMMA-public/MultiBypass140} \\
LapGyn4 & \url{https://ftp.itec.aau.at/datasets/LapGyn4} \\
SurgicalAction160 & \url{https://ftp.itec.aau.at/datasets/SurgicalActions160} \\
GLENDA & \url{https://ftp.itec.aau.at/datasets/GLENDA} \\
HeiCo & \url{https://www.synapse.org/Synapse:syn21903917/wiki/601992} \\
SurgToolLoc2022 & \url{https://surgtoolloc23.grand-challenge.org/surgtoolloc-2022-resources/} \\
DSAD & \url{https://www.kaggle.com/datasets/anindyamajumder/the-dresden-surgical-anatomy-dataset} \\
GraSP & \url{https://github.com/BCV-Uniandes/GraSP} \\
SCARED & \url{https://endovissub2019-scared.grand-challenge.org/} \\
Hamlyn & \url{https://hamlyn.doc.ic.ac.uk/vision/} \\
PitVQA & \url{https://github.com/mobarakol/PitVQA} \\
Surgical Youtube & \url{https://github.com/TimJaspers0801/SurgeNet} \\
\hline
\end{tabular}
\end{table}

\end{document}